\documentclass[12pt,twoside]{article}
\usepackage{times}
\usepackage{amssymb}

\mathchardef\Gammaoffont="7000
\mathchardef\Gamma="0100
\mathchardef\Deltaoffont="7001
\mathchardef\Delta="0101
\mathchardef\Thetaoffont="7002
\mathchardef\Theta="0102
\mathchardef\Lambdaoffont="7003
\mathchardef\Lambda="0103
\mathchardef\Xioffont="7004
\mathchardef\Xi="0104
\mathchardef\Pioffont="7005
\mathchardef\Pi="0105
\mathchardef\Sigmaoffont="7006
\mathchardef\Sigma="0106
\mathchardef\Upsilonoffont="7007
\mathchardef\Upsilon="0107
\mathchardef\Phioffont="7008
\mathchardef\Phi="0108
\mathchardef\Psioffont="7009
\mathchardef\Psi="0109
\mathchardef\Omegaoffont="700A
\mathchardef\Omega="010A
\mathchardef\itype="017B

\catcode`\@=11

\gdef\allowhyphens{\penalty\@M \hskip\z@skip}

\gdef\set@low@box#1{\setbox\tw@\hbox{,}\setbox\z@\hbox{#1}\dimen\z@\ht\z@
     \advance\dimen\z@ -\ht\tw@
     \setbox\z@\hbox{\lower\dimen\z@ \box\z@}\ht\z@\ht\tw@ \dp\z@\dp\tw@ }
\gdef\set@low@boxsingle#1{\setbox\tw@\hbox{\rm,}\setbox\z@\hbox{#1}\dimen\z@\ht\z@
     \advance\dimen\z@ -\ht\tw@
     \setbox\z@\hbox{\lower\dimen\z@ \box\z@}\ht\z@\ht\tw@ \dp\z@\dp\tw@ }

\gdef\@glqq{%
\ifhmode\edef\@SF{\spacefactor\the\spacefactor}%
\else\let\@SF\empty
\fi
\CheckFamily\font\fraknomath\ifSameFamily ``\relax
\else\CheckFamily\font\swab\ifSameFamily ``\relax
\else\leavevmode\set@low@box{''}\box\z@\kern-.04em\allowhyphens\@SF\relax
\fi\fi}
\gdef\glqq{\protect\@glqq\kern+.07em}
\gdef\@grqq{%
\ifhmode\edef\@SF{\spacefactor\the\spacefactor}%
\else\let\@SF\empty 
\fi 
\CheckFamily\font\fraknomath\ifSameFamily ''\relax
\else\CheckFamily\font\swab\ifSameFamily ''\relax
\else\kern+.07em``\kern.07em\@SF\relax
\fi\fi}
\gdef\grqq{\protect\@grqq}
\gdef\@glq{{\ifhmode \edef\@SF{\spacefactor\the\spacefactor}\else
     \let\@SF\empty \fi \leavevmode
     \set@low@boxsingle{'\/}\box\z@\kern-.04em\allowhyphens\@SF\relax}}
\gdef\glq{\protect\@glq\kern+.07em}
\gdef\@grq{\ifhmode \edef\@SF{\spacefactor\the\spacefactor}\else
     \let\@SF\empty \fi \kern-.0125em`\kern.07em\@SF\relax}
\gdef\grq{\protect\@grq}

\catcode`\@=12

\makeatletter
\if \@ptsize 0
   \newfont{\scriptscriptscriptgoth}{ygoth scaled 760}
   \newfont{\scriptscriptgoth}{ygoth scaled 833}
   \newfont{\scriptgoth}{ygoth scaled 912}
   \newfont{\gothnomath}{ygoth}
   \newfont{\Goth}{ygoth scaled \magstephalf}
   \newfont{\GOth}{ygoth scaled \magstep1}
   \newfont{\GOTh}{ygoth scaled \magstep2}
   \newfont{\GOTH}{ygoth scaled \magstep3}

   \newfont{\scriptscriptscriptswab}{yswab scaled 760}
   \newfont{\scriptscriptswab}{yswab scaled 833}
   \newfont{\scriptswab}{yswab scaled 912}
   \newfont{\swab}{yswab}
   \newfont{\Swab}{yswab scaled \magstephalf}
   \newfont{\SWab}{yswab scaled \magstep1}
   \newfont{\SWAb}{yswab scaled \magstep2}
   \newfont{\SWAB}{yswab scaled \magstep3}

   \newfont{\scriptscriptscriptfrak}{yfrak scaled 760}
   \newfont{\scriptscriptfrak}{yfrak scaled 833}
   \newfont{\scriptfrak}{yfrak scaled 912}
   \newfont{\fraknomath}{yfrak}
   \newfont{\Frak}{yfrak scaled \magstephalf}
   \newfont{\FRak}{yfrak scaled \magstep1}
   \newfont{\FRAk}{yfrak scaled \magstep2}
   \newfont{\FRAK}{yfrak scaled \magstep3}

   \newfont{\init}{yinit}
   \newfont{\Init}{yinit scaled \magstephalf}
   \newfont{\INit}{yinit scaled \magstep1}
   \newfont{\INIt}{yinit scaled \magstep2}
   \newfont{\INIT}{yinit scaled \magstep3}
\fi
\if \@ptsize 1
   \newfont{\scriptscriptscriptgoth}{ygoth scaled 833}
   \newfont{\scriptscriptgoth}{ygoth scaled 912}
   \newfont{\scriptgoth}{ygoth}
   \newfont{\gothnomath}{ygoth scaled \magstephalf}
   \newfont{\Goth}{ygoth scaled \magstep1}
   \newfont{\GOth}{ygoth scaled \magstep2}
   \newfont{\GOTh}{ygoth scaled \magstep3}
   \newfont{\GOTH}{ygoth scaled \magstep4}

   \newfont{\scriptscriptscriptswab}{yswab scaled 833}
   \newfont{\scriptscriptswab}{yswab scaled 912}
   \newfont{\scriptswab}{yswab}
   \newfont{\swab}{yswab scaled \magstephalf}
   \newfont{\Swab}{yswab scaled \magstep1}
   \newfont{\SWab}{yswab scaled \magstep2}
   \newfont{\SWAb}{yswab scaled \magstep3}
   \newfont{\SWAB}{yswab scaled \magstep4}

   \newfont{\scriptscriptscriptfrak}{yfrak scaled 833}
   \newfont{\scriptscriptfrak}{yfrak scaled 912}
   \newfont{\scriptfrak}{yfrak}
   \newfont{\fraknomath}{yfrak scaled \magstephalf}
   \newfont{\Frak}{yfrak scaled \magstep1}
   \newfont{\FRak}{yfrak scaled \magstep2}
   \newfont{\FRAk}{yfrak scaled \magstep3}
   \newfont{\FRAK}{yfrak scaled \magstep4}

   \newfont{\init}{yinit scaled \magstephalf}
   \newfont{\Init}{yinit scaled \magstep1}
   \newfont{\INit}{yinit scaled \magstep2}
   \newfont{\INIt}{yinit scaled \magstep3}
   \newfont{\INIT}{yinit scaled \magstep4}
\fi
\if \@ptsize 2
   \newfont{\scriptscriptscriptgoth}{ygoth scaled 912}
   \newfont{\scriptscriptgoth}{ygoth}
   \newfont{\scriptgoth}{ygoth scaled \magstephalf}
   \newfont{\gothnomath}{ygoth scaled \magstep1}
   \newfont{\Goth}{ygoth scaled \magstep2}
   \newfont{\GOth}{ygoth scaled \magstep3}
   \newfont{\GOTh}{ygoth scaled \magstep4}
   \newfont{\GOTH}{ygoth scaled \magstep5}

   \newfont{\scriptscriptscriptswab}{yswab scaled 912}
   \newfont{\scriptscriptswab}{yswab}
   \newfont{\scriptswab}{yswab scaled \magstephalf}
   \newfont{\swab}{yswab scaled \magstep1}
   \newfont{\Swab}{yswab scaled \magstep2}
   \newfont{\SWab}{yswab scaled \magstep3}
   \newfont{\SWAb}{yswab scaled \magstep4}
   \newfont{\SWAB}{yswab scaled \magstep5}

   \newfont{\scriptscriptscriptfrak}{yfrak scaled 912}
   \newfont{\scriptscriptfrak}{yfrak}
   \newfont{\scriptfrak}{yfrak scaled \magstephalf}
   \newfont{\fraknomath}{yfrak scaled \magstep1}
   \newfont{\Frak}{yfrak scaled \magstep2}
   \newfont{\FRak}{yfrak scaled \magstep3}
   \newfont{\FRAk}{yfrak scaled \magstep4}
   \newfont{\FRAK}{yfrak scaled \magstep5}

   \newfont{\init}{yinit scaled \magstep1}
   \newfont{\Init}{yinit scaled \magstep2}
   \newfont{\INit}{yinit scaled \magstep3}
   \newfont{\INIt}{yinit scaled \magstep4}
   \newfont{\INIT}{yinit scaled \magstep5}
\fi

\makeatother

\newif\ifSameFamily
\def\CheckFamily#1#2{\GetFamilyName{#1}\ArgOne
        \GetFamilyName{#2}\ArgTwo
        \ifx\ArgOne\ArgTwo\SameFamilytrue\else\SameFamilyfalse\fi}
\def\GetFamilyName#1{\edef\Tempa{#1}\def\Tempb{#1}\ifx\Tempa\Tempb
        \edef\Tempa{\fontname#1}\fi
        \edef\Tempa{\Tempa\space}%
        \expandafter\iGetFamilyName\Tempa\\}
\def\iGetFamilyName#1 #2\\#3{\def#3{#1}}
\def\DefFontName#1#2{{\escapechar-1\expandafter\expandafter\expandafter
        \iDefFontName\expandafter{\csname#2\endcsname}%
        \xdef#1{\expandafter\string\Tempa}}}
\def\iDefFontName{\def\Tempa}

\newcommand\unprotectedae
{\CheckFamily\font\fraknomath\ifSameFamily *a\else
 \CheckFamily\font\swab\ifSameFamily\char'212\else\"a\fi\fi}
\newcommand\unprotectedoe
{\CheckFamily\font\fraknomath\ifSameFamily 
*o\else\CheckFamily\font\swab\ifSameFamily\char'232\else\"o\fi\fi}
\newcommand\unprotectedue
{\CheckFamily\font\fraknomath\ifSameFamily 
*u\else\CheckFamily\font\swab\ifSameFamily\char'237\else\"u\fi\fi}

\DefFontName\eccclarge{eccc1200}
\DefFontName\eccc{eccc1000}
\DefFontName\ecccsmall{eccc0900}
\DefFontName\ecccfootnotesize{eccc0800}

\newcommand\unprotectedes
{\CheckFamily\font\fraknomath\ifSameFamily\char'215\else
\CheckFamily\font\swab\ifSameFamily\char'215\else  
s\fi\fi}

\newcommand\unprotectedmyparagraphsymbol
{\CheckFamily\font\fraknomath\ifSameFamily 
\char'244\else\CheckFamily\font\swab\ifSameFamily
\char'244\else\S\fi\fi}

\newcommand\unprotectedewithtrema
{\-\CheckFamily\font\fraknomath\ifSameFamily 
 "e\else\CheckFamily\font\swab\ifSameFamily "e\else\"e\fi\fi}

\renewcommand\ae{\protect\unprotectedae}
\renewcommand\oe{\protect\unprotectedoe}
\newcommand\ue  {\protect\unprotectedue}

\newcommand\es  {\protect\unprotectedes}

\newcommand\myparagraphsymbol{\protect\unprotectedmyparagraphsymbol}

\newcommand\ewithtrema{\protect\unprotectedewithtrema}

\newcommand\namefont{}

\newcommand\footroom{\raisebox{-1.5ex}{\rule{0ex}{.5ex}}}

\newcommand\smallfootroom{\raisebox{-1.0ex}{\rule{0ex}{3ex}}}

\newcommand\majorheadroom{\rule{0ex}{3.2ex}}
\newcommand\headroom{\rule{0ex}{2.8ex}}
\newcommand\mediumheadroom{\rule{0ex}{2.4ex}}

\newcommand\andreas {An\-dre\-a\es}

\newcommand\bernhard{Bern\-hard}

\newcommand\christian{Christian}

\newcommand\claus   {Clau\es}

\newcommand\donald  {Do\-nald}

\newcommand\gerhard {Ger\-hard}

\newcommand\hans    {Han\es}

\newcommand\henk    {Henk}

\newcommand\joerg   {J\oe rg}

\newcommand\juergen {J\ue r\-gen}

\newcommand\klaus   {Klau\es}
\newcommand\kurt    {Kurt}

\newcommand\maxchristianname{Max} 

\newcommand\matthias{Matthia\es}
\newcommand\michael {Mi\-cha\-\ewithtrema l}

\newcommand\peter   {Peter}
\newcommand\pierre  {Pierre}
\newcommand\raymond {Raymond}

\newcommand\roland  {Ro\-land}

\newcommand\wolfgang{Wolf\-gang}
\newcommand\ulrich  {Ul\-rich}

\newcommand\baaz            {Baaz}
\newcommand\baazname        {\matthias\ \baaz}

\newcommand\backhouse       {Backhouse}
\newcommand\backhousename   {\roland\ \backhouse}

\newcommand\baumgartner     {Baum\-gart\-ner}
\newcommand\baumgartnername {\peter\ \baumgartner}
\newcommand\becker          {Becker}
\newcommand\beckername      {\klaus\ \becker}
\newcommand\beckert         {Beckert}
\newcommand\beckertname     {\bernhard\ \beckert}

\newcommand\bibel           {Bi\-bel}
\newcommand\bibelname       {\wolfgang\ \bibel}

\newcommand\bourbaki        {Bourbaki}
\newcommand\bourbakiname    {Nicola\es\ \bourbaki}

\newcommand\buerckert       {B\ue rckert}
\newcommand\buerckertname   {\hans-\juergen\ \buerckert}

\newcommand\caferra         {Caferra}
\newcommand\caferraname     {Ricardo \caferra}

\newcommand\doornbos        {Doornbo\es}
\newcommand\doornbosname    {\henk\ \doornbos}

\newcommand\fermat          {\mbox{\namefont Fermat}}
\newcommand\fermatname      {{\namefont \pierre\ \fermat}}

\newcommand\fermueller      {Fer\-m\ue ller}
\newcommand\fermuellername  {\christian\ G. \fermueller}

\newcommand\fitting         {Fitting}
\newcommand\fittingname     
 {Melvin       \fitting}

\newcommand\furbach         {Furbach}
\newcommand\furbachname     {\ulrich\ \furbach}

\newcommand\gentzen         {Gentzen}

\newcommand\gentzenname     {\gerhard\ \gentzen}

\newcommand\goedel          {{\namefont G\oe del}}
\newcommand\goedelname      {{\namefont \kurt\ \goedel}}

\newcommand\gramlich        {Gram\-lich}
\newcommand\gramlichname    {\bernhard\ \gramlich}

\newcommand\haehnle         {H\ae hnle}
\newcommand\haehnlename     {Reiner \haehnle}

\newcommand\hilbert         {{\namefont Hil\-bert}}

\newcommand\kanger          {Kan\-ger}
\newcommand\kangername      {Stig \kanger}

\newcommand\knuth           {Knuth}
\newcommand\knuthname       {\donald\ E. \knuth}

\newcommand\kohlhase        {Kohl\-hase}
\newcommand\kohlhasename    {\michael\ \kohlhase}

\newcommand\miller          {Mil\-ler}
\newcommand\millername      {Dale A. \miller} 

\newcommand\nonnengart      {Nonnengart}
\newcommand\nonnengartname  {\andreas\ \nonnengart}

\newcommand\padawitz        {Pada\-witz}
\newcommand\padawitzname    {\peter\ \padawitz}

\newcommand\paterson        {Paterson}
\newcommand\patersonname    {\michael\ S. \paterson}

\newcommand\planck          {Planck}
\newcommand\planckname      {\maxchristianname\ \planck}

\newcommand\prawitz         {Pra\-witz}
\newcommand\prawitzname     {Dag \prawitz}

\newcommand\salzer          {Salzer}
\newcommand\salzername      {Gernot \salzer}

\newcommand\schmitt         {Schmitt}
\newcommand\schmittname     {\peter\ H. \schmitt}

\newcommand\siekmann        {{\namefont Siek\-mann}}
\newcommand\siekmannname    {{\namefont \joerg\ \siekmann}}

\newcommand\smullyan        {Smullyan}
\newcommand\smullyanname    {\raymond\ M. \smullyan}

\newcommand\stolzenburg     {Stol\-zen\-burg}
\newcommand\stolzenburgname {Frie\-der \stolzenburg}

\newcommand\wegman          {Wegman} 
\newcommand\wegmanname      {Mark N. \wegman}

\newcommand\wirth           {{\namefont Wirth}}
\newcommand\wirthname       {{\namefont\claus-\peter\ \wirth}}

\newcommand\woudename       {Jaap van der Woude}
\newcommand\wrightson       {Wright\-son}
\newcommand\wrightsonname   {Graham \wrightson}

\newcommand\defi {Definition} 

\newcommand\Int  {Int.}

\newcommand\Inst {Inst.}

\newcommand\Nov  {Nov.}

\newcommand\qedhelp[1]{Q.e.d.~({#1})}
\newcommand\getittotheright[1]  
{\hfill\mbox{}\penalty 100\mbox{\ \,}\nolinebreak
\hfill\hfill\hfill\nolinebreak\mbox{#1}\ignorespaces}
\newcommand\Qed      [1]{\underline{\qedhelp{#1}}}

\newcommand\Qedbf    [1]{\mbox{\bf\qedhelp{#1}}}
\newcommand\QED      [1]{\getittotheright{\Qed      {#1}}}

\newcommand\QEDbf    [1]{\getittotheright{\Qedbf    {#1}}}

\newcommand\Sep  {Sept.}

\newcommand\theo {Theorem}
\newcommand\uni  {Uni\-ver\-si\-t\ae t}

\newcommand\Vol  {Vol.}

\newcommand\www  {\url{http://www.ags.uni-sb.de/~cp}}

\newcommand\AI   {Artificial Intelligence}

\newcommand\Cf   {Cf.}
\newcommand\cf   {cf.}

\newcommand\ConstructorBased{Con\-struc\-tor-Based}
\newcommand\CS   {Computer \Sci}

\newcommand\Dissertation{Dissertation (\PhDthesis)}

\newcommand\ed   {ed.}

\newcommand\eds  {eds.}
\newcommand\eg   {e.g.}
\newcommand\Eg   {E.g.}

\newcommand\etalabbrev{\&al.}
\newcommand\etc  {\&c.}

\newcommand\extd {extd.}

\newcommand\signatureenlarge     {enrich}

\newcommand\ie   {i.e.}
\newcommand\Ie   {I.e.}

\newcommand\uiff {\ iff\ }
\newcommand\udiff{\ if\ }

\newcommand\ITP  {Inductive Theorem Proving}

\newcommand\p    {p.}
\newcommand\pp   {pp.}
\newcommand\PP[2]{\pp\,\ignorespaces#1--\ignorespaces#2}

\newcommand\PhD  {Ph.D.}
\newcommand\PhDthesis{\PhD\ thesis}

\newcommand\Proc {Proc.}

\newcommand\PNC  {Posi\-tive/Ne\-ga\-tive-Condi\-tional}

\newcommand\Publ {Publ.}

\newcommand\resp {resp.}
\newcommand\rev  {rev.}

\newcommand\sect {\myparagraphsymbol} 

\newcommand\Sci  {Sci.}

\newcommand\wrog {w.l.o.g.} 
\newcommand\wrt  {w.r.t.}
\newcommand\Wrt  {W.r.t.}

\newcommand\littheoref[1]{\theo\,#1}
\newcommand\litsectref[1]{\sect\,#1} 

\newcommand\litdefiref[1]{\defi\,#1}
\newcommand\Examplename{Ex\-am\-ple}
\newcommand\litexamref[1]{\Examplename\,#1}

\newcommand\litlemmref[1]{Lem\-ma\,#1}

\newcommand\lemmref[1]{\litlemmref{\ref{#1}}}

\newcommand\examref[1]{\litexamref{\ref{#1}}}

\newcommand\defiref[1]{\litdefiref{\ref{#1}}}
\newcommand\theoref[1]{\littheoref{\ref{#1}}}

\newcommand\sectref[1]{\litsectref{\ref{#1}}}

\newcommand\nthpositioner[2]
{#1\raisebox{0.52ex}{\scriptsize\hspace{0.07em}#2}}
\newcommand\nth[1]{\nthtinypositioner{#1}{\nthstring{#1}}}
\newcommand\nthtinypositioner[2]{#1\raisebox{0.52ex}{\tiny\hspace{0.07em}#2}}

\newcommand\mthpositioner[2]
{\math{#1}\raisebox{0.52ex}{\scriptsize\hspace{0.07em}#2}}
\newcommand\modulointocountzero[2]
{\count1=#1
 \count2=#2
 \count0=\count1
 \divide  \count0 by \count2
 \multiply\count0 by-\count2
 \advance \count0 by \count1}
\newcommand\absolutevalueintocountzero[1]
{\count0=#1
 \ifnum\count0<0\multiply\count0 by -1\fi}
\newcommand\nthstring[1]
{\def\myargone{#1}\ifcat a\myargone th\else\nthstringnochar{#1}\fi}
\newcommand\nthstringnochar[1]
{\absolutevalueintocountzero{#1}%
 \modulointocountzero{\count0}{100} 
 \ifnum\count0>9\ifnum\count0<20 th\else\stupidnthstring\fi
                                   \else\stupidnthstring\fi}
\newcommand\stupidnthstring
{\modulointocountzero{\count0}{10}
 \ifnum\count0=1 \hskip-0.2em st\else
 \ifnum\count0=2 nd\else
 \ifnum\count0=3 rd\else 
                 th\fi\fi\fi}

\newcommand\writeascents
[1]{\count4=#1
\ifnum\count4<0 
-\multiply\count4 by -1\fi
\modulointocountzero{\count4}{10}
\divide\count4 by 10
\count3=\the\count0
\modulointocountzero{\count4}{10}
\divide\count4 by 10
\the\count4
.\the\count0
\the\count3
}

\newcommand\frenchnthstring[1]
{\def\myargone{#1}\ifcat a\myargone th\else\frenchnthstringnochar{#1}\fi}
\newcommand\frenchnthstringnochar[1]
{\absolutevalueintocountzero{#1}%
 \modulointocountzero{\count0}{100} 
 \ifnum\count0>9\ifnum\count0<20 th\else\frenchstupidnthstring\fi
                                   \else\frenchstupidnthstring\fi}
\newcommand\frenchstupidnthstring
{\modulointocountzero{\count0}{10}
 \ifnum\count0=1 \hskip-0.2em re\else
 \ifnum\count0=2 me\else
 \ifnum\count0=3 rd\else 
                 th\fi\fi\fi}

\newcommand\CLAM      {{\rm CL\kern-.36em\raise.39ex\hbox{\sc a}\kern-.15emM}}

\newcommand\TEXMACS   {{\sc T\kern-.1667em\lower.5ex\hbox{E}\kern-.125emX\kern-.1em\lower.5ex\hbox{\textsc{m\kern-.05ema\kern-.125emc\kern-.05ems}}}}

\newcommand\MPII{\planckname\ \Inst\ f\ue r Informatik}

\newcommand\DO             {Dort\-mund}

\newcommand\SB             {Saar\-br\ue cken}

\newcommand\uniDO{\uni\ \DO}

\def       \email        {{\tt wirth@logic.at}}

\newcommand\LNCSvol[1]
{\mbox{LNCS}\,\ignorespaces#1}

\newcommand\LNAIvol[1]
{\mbox{LNAI}\,\ignorespaces#1}

\newcommand\academicpress{Academic Press (\elsevier)}

\newcommand\addisonwesley{Addi\-son-Wes\-ley}

\newcommand\elsevier{Elsevier}

\newcommand\kluwer{Kluwer (\springerverlag)
}

\newcommand\springerverlag{Sprin\-ger}

\newcommand\mpireportnoaddress[2]{MPI--I--#1--2--#2}
\newcommand\mpireport[2]{\mpireportnoaddress{#1}{#2}, \MPII, \SB}

\newcommand\lncsconf[6]{\nth{#2}\,#1\,#3, #4, \PP{#5}{#6}, \springerverlag}

\newcommand\CADEshort{CADE}

\newcommand\twelvethCADEninetyfour               
{\lncsconf\CADEshort{12}{1994}{\LNAIvol{ 814}}}

\newcommand\CTRSshort{CTRS}

\newcommand\fourthCTRSninetyfour   {\lncsconf\CTRSshort{4}{1994}{\LNCSvol{968}}}

\newcommand\TABLEAUshort{TAB\-LEAUX}

\newcommand\fourthTABLEAUninetyfive
{\lncsconf\TABLEAUshort{4} {1995}{\LNAIvol{ 918}}}

\newcommand\eighthTABLEAUninetynine
{\lncsconf\TABLEAUshort{8} {1999}{\LNAIvol{1617}}}

\newcommand\newspaperreference[5]
{\def\nameofjournalpress{#2}#1, #4 #5, #3\if?\nameofjournalpress
 \else, #2\fi}

\newcommand\dateinjournal[1]{}

\newcommand\journalreference[6]
{\def\nameofjournalpress{#2}#1\nolinebreak\hskip.2em%
 \dateinjournal{(#3) }{\mbox{\bf #4}}, \PP{#5}{#6}\if?\nameofjournalpress
 \else, #2\fi}

\newcommand\journalreferenceprintyear[6]
{\def\nameofjournalpress{#2}#1 
 {(#3) }{\bf #4}, \PP{#5}{#6}\if?\nameofjournalpress
 \else, #2\fi}

\newcommand\journalreferenceprintyearaspartofnumber[6]
{\def\nameofjournalpress{#2}#1 
 {#4/#3}, \PP{#5}{#6}\if?\nameofjournalpress
 \else, #2\fi}

\newcommand\artificialintelligence
{\journalreference{\AI}?}

\newcommand\jarname
{J. Automated Reasoning}
\newcommand\jar
{\journalreference\jarname\kluwer}

\newcommand\jcssname{J. Computer and System \Sci}
\newcommand\jcss
{\journalreference{\jcssname}\academicpress}

\newcommand\jscname
{J. Symbolic Computation}
\newcommand\jsc
{\journalreference{\jscname}\academicpress}

\newcommand\jscprintyear
{\journalreferenceprintyear{\jscname}\academicpress}

\newcommand\mathematischezeitschrift
{\journalreference{Mathematische Zeitschrift}?}

\newcommand\tcsname{Theoretical \CS}
\newcommand\tcsjournal
{\journalreference\tcsname\elsevier}
\newcommand\tcsjournalprintyear
{\journalreferenceprintyear\tcsname\elsevier}

\newcommand\SEKIedition
{Searchable Online Edition}
\catcode`\@=11
\input dina4.sty
\input mssymb.sty
\catcode`\@=12
\newlength{\sectionintocsep} 
\newlength{\contentsandreferencesheadroom} 
\newlength{\contentsandreferencesfootroom} 
\catcode`\@=11

\def\tableofcontents{\ignorespaces
\section*{\contentsname\@mkboth
{\uppercase{\contentsname}}{\uppercase{\contentsname}}}%
\vspace*\contentsandreferencesfootroom
\@starttoc{toc}}
\def\contentsname{Contents} 

\catcode`\@=12

\input headerhot

\newcommand\Proofof{Proof of}
\renewenvironment{proof}[1]{\yestop\begin{sloppypar}\noindent
{\bf\Proofof\ {#1}}\\}{\end{sloppypar}}
\renewenvironment{proofqed}[1]
{\begin{sloppypar}\def\fooqed{#1}\noindent{\bf\Proofof\ \fooqed}}
{\QEDbf\fooqed\end{sloppypar}}
\renewenvironment{proofparsep}[1]{\parindent=0pt\begin
{sloppypar}\noindent{\bf\Proofof\ {#1}}\nopagebreak\par}
{\end{sloppypar}}
\renewenvironment{proofparsepqed}[1]{\parindent=0pt\begin
{sloppypar}\def\fooqed{#1}\noindent{\bf\Proofof\ \fooqed}\nopagebreak\par}
{\nopagebreak\QEDbf\fooqed\end{sloppypar}}
\renewenvironment{proofshort}[1]{\begin{sloppypar}\noindent
{\bf\Proofof\ {#1}:}}{\end{sloppypar}}

{\end{enumerate}\renewcommand{\theenumi}{\arabic{enumi}}}

\mathcommand\myfootnotemark[1]{^{#1}}

\newcommand\repname{{\rm set}}
\mathapplycommand\rep\repname
\mathcommand\repr[1]{{\repname[{#1}]}}
\mathcommand\msa{\langle}
\mathcommand\mse{\rangle}
\mathcommand\msu{\,\sqcup\,}
\mathcommand\msin{{\rm\;in\;}}
\mathcommand\mssetminus{\setminus\!\!\setminus}
\mathcommand\tightmssubseteq{\sqsubseteq}
\mathcommand\mssubseteq{\ \tightmssubseteq\ }
\mathcommand\approxapprox{\approx\:\!\!\approx}
\mathcommand\quasilquasil{\,\lesssim\!\lesssim\,}
\mathcommand\quasibquasib{\,\gtrsim\!\gtrsim\,}
\mathcommand\fmul[1]{{\rm FMul}(#1)}
\mathcommand\smul[1]{{\rm SMul}(#1)}
\mathcommand\multisetwith [2]{\msa\ {#1}\ |\ {#2}\ \mse}
\mathcommand\multisetwithq[3]{\msa\ {#2}\ |_{#1}\ {#3}\ \mse}
\newcommand\superseteq     {\supseteq}

\mathcommand\quasilhd
{\mathop{\mbox{\raisebox{0.31ex}{\math\lhd}\hspace{-0.75em}\raisebox
{-0.6ex}{\math\sim}}}}
\newcommand\quasirhd{\mbox{\raisebox{0.31ex}{$\rhd$}\hspace{-0.75em}\raisebox{-0.6ex}{$\sim$}}}

\mathcommand\rhdrhd{\rhd$\hspace{-0.35em}$\rhd}
\mathcommand\lhdlhd{\lhd$\hspace{-0.21em}$\lhd}

\mathcommand\quasilhdquasilhd{\quasilhd$\hspace{-0.13em}$\quasilhd}

\newcommand\hiddensubSS{_{_{\rm SS}}}
\mathcommand\antisubsum     {\rhd\hiddensubSS}
\mathcommand\notantisubsum  {\ntriangleright\hiddensubSS}
\mathcommand\subsum         {\lhd\hiddensubSS}
\mathcommand\notsubsum      {\ntriangleleft\hiddensubSS}
\mathcommand\antisubsumeq   {\trianglerighteq\hiddensubSS}
\mathcommand\subsumeq       {\trianglelefteq\hiddensubSS}
\mathcommand\quasisubsum    {\,\quasilhd\raisebox{0.1ex}{$\hiddensubSS$}}
\mathcommand\antiquasisubsum{\,\quasirhd\raisebox{0.1ex}{$\hiddensubSS$}}
\mathcommand\quasiquasisubsum{\quasisubsum\!\!\quasisubsum}
\mathcommand\antiquasiquasisubsum{\antiquasisubsum\!\!\!\antiquasisubsum}

\newcommand\hiddensubH{_{_{\rm H}}}
\newcommand\hiddensubCONS{_{_\CONS}}
\mathcommand\hql   {\,\lesssim\hiddensubH}
\mathcommand\consql{\,\lesssim\hiddensubCONS}
\mathcommand\hl    {\,<       \hiddensubH}
\mathcommand\hleq  {\,\leq    \hiddensubH}
\mathcommand\consl {\,<       \hiddensubCONS}
\mathcommand\conseq{\,\approx \hiddensubCONS}

\newcommand\Inthesequel{In what follows}

\newcommand\cons {{\rm cons}}
\mathcommand\sigconsV{\sig/\cons/\V}
\mathcommand\sigconsR{\sig/\cons/\R}
\mathcommand\primesigconsV{\sig'\!/\cons'\!/\V'}
\mathcommand\primesigconsR{\sig'\!/\cons'\!/\R'}
\newcommand\dunno{{\rm sc}}
\newcommand\DUNNO{{\rm SC}}
\mathcommand\SIGCONS   {\{\SIG,\CONS\}}
\mathcommand\sigsortstimes{\SIGCONS\tight\times\sigsorts}

\mathcommand\TERMSSYM
{\mathcal{T\hspace{-.20em}E\hspace{-.08em}R\hspace{-.16em}M\hspace{-.10em}S}}
\mathapplycommand\condterms{\TERMSSYM}
\mathcommand\kurzregel{((l,r),C)}
\mathcommand\kurzregelprime{((l',r'),C')}
\mathcommand\kurzregelindex[1]{((l_{#1},r_{#1}),C_{#1})}
\mathapplycommand\lhs{\rm lhs}

\mathcommand\red{\redsimple} 

\mathcommand\lemms{L}
\mathcommand\hypos{H}
\mathcommand\goals{G}
\mathcommand\lemmsprime{\lemms'}
\mathcommand\hyposprime{\hypos'}
\mathcommand\goalsprime{\goals'}
\mathcommand\lemmsprimeprime{\lemms''}
\mathcommand\hyposprimeprime{\hypos''}
\mathcommand\goalsprimeprime{\goals''}
\mathcommand\oldtriple            {(\lemms   ,\hypos   ,\goals  )}
\mathcommand\inittriple        {(\emptyset,\emptyset,\goals  )}
\mathcommand\triplehelp[1]     {(\lemms#1,\hypos#1 ,\goals#1)}
\mathcommand\tripleprime       {\triplehelp'}
\mathcommand\triplenogoalsprime{(\lemmsprime,\hyposprime,\emptyset  )}
\mathcommand\tripleprimeprime  {\triplehelp{''}}
\mathcommand\tripleindex[1]    {\triplehelp{_{#1}}}

\mathcommand\constcong[1]{\,\,\sim_{\!_{#1}}\,}

\mathapplycommand\avail{\rm\Av ail}

\def\emph#1{\/ {\itshape#1}\/}
\newcommand\tightemph[1]{\/{\itshape#1}\/}

\input headertheorem
\mathcommand\notconflu{\mathchoice
             {{\hskip1.5pt\nmid\hskip-4.697545pt\downarrow}}
             {{\hskip1.5pt\nmid\hskip-4.65pt\downarrow}}   
             {{\hskip1pt\nmid\hskip-3.494pt\downarrow\hskip1pt}}  
             {{\hskip1pt\nmid\hskip-3.01pt\downarrow\hskip0.5pt}}   
}
\mathcommand\redpara{\mathchoice
           {{\redsimple\hskip-16pt  \shortparallel}\hskip8.5pt}
           {{\redsimple\hskip-16pt  \shortparallel}\hskip8.5pt}
           {{\redsimple\hskip-8.5pt \shortparallel}\hskip6pt}
           {{\redsimple\hskip-7.5pt \shortparallel}\hskip5pt}
}
\mathcommand\antiredpara{\mathchoice
           {{\antired\hskip-14.6pt  \shortparallel}\hskip7pt}
           {{\antired\hskip-14.6pt  \shortparallel}\hskip7pt}
           {{\antired\hskip-8.pt \shortparallel}\hskip5pt}
           {{\antired\hskip-7.pt \shortparallel}\hskip5pt}
}
\mathcommand\revpara{\mathchoice
           {{\redsimple\hskip-16pt  \infty}\hskip4.8pt}
           {{\redsimple\hskip-16pt  \infty}\hskip4.8pt}
           {{\redsimple\hskip-11.5pt\infty}\hskip4pt}
           {{\redsimple\hskip-9.9pt \infty}\hskip3pt}
}
\mathcommand\antirevpara{\mathchoice
           {{\antired\hskip-15.4pt\infty}\hskip4pt}
           {{\antired\hskip-15.4pt\infty}\hskip4pt}
           {{\antired\hskip-10.8pt\infty}\hskip3pt}
           {{\antired\hskip-9.5pt \infty}\hskip3pt}
}
\mathcommand\simpara{\mathchoice
           {{\redsimple\hskip-13pt  \circ}\hskip7pt}
           {{\redsimple\hskip-13pt  \circ}\hskip7pt}
           {{\redsimple\hskip-11.5pt\circ}\hskip4pt}
           {{\redsimple\hskip-9.9pt \circ}\hskip3pt}
}

\mathcommand\ident[1]{\mathsf{#1}}
\newcommand\plussymbol  {\ident{+}}
\newcommand\minussymbol {\ident{-}}
\newcommand\dividesymbol{\ident{/}}
\newcommand\timessymbol {\ident{*}}

\newcommand\set     {\ident{set}}

\newcommand\naturalssymbol{\ident{naturals}}
\newcommand\gensymsymbol{\ident{gensym}}
\mathcommand\mbpsymbol{\ident{m\hspace{-0.055em}b\hspace{-0.045em}p}}

\newcommand\csymbol     {\ident c}
\newcommand\esymbol     {\ident e}
\newcommand\fsymbol     {\ident f}
\newcommand\gsymbol     {\ident g}
\newcommand\hsymbol     {\ident h}
\newcommand\ksymbol     {\ident k}
\newcommand\psymbol     {\ident p}
\newcommand\ssymbol     {\ident s}
\newcommand\Everysymbol {\ident{Every}}
\newcommand\Permsymbol {\ident{Perm}}
\newcommand\RExistssymbol{\ident{Rexists}}
\newcommand\invertsymbol{\ident{invert}}
\newcommand\invsymbol{\ident{inv}}
\newcommand\abssymbol   {\ident{abs}}
\newcommand\cnssymbol   {\ident{cons}}
\mathcommand\cnsindexsymbol[1]{\ident{cons}_{#1}}

\newcommand\lengthsymbol{\ident{length}}
\newcommand\dlsymbol    {\ident{dl}}
\newcommand\dloncesymbol{\ident{delonce}}
\newcommand\rcsymbol    {\ident{rc}}
\newcommand\brsymbol    {\ident{br}}
\newcommand\revtailsymbol{\ident{revtail}}
\newcommand\revsymbol{\ident{rev}}
\newcommand\appendsymbol {\ident{append}}
\newcommand\zeropredicatesymbol{\ident{zerop}}
\newcommand\eqsymbol        {\ident{eq}}
\newcommand\ifthensymbol    {\mbox{\ident{If{}Then}}}
\newcommand\ifthenelsesymbol{\mbox{\ident{If{}ThenElse}}}
\mathcommand\eqindexsymbol        [1]{\eqsymbol        _{#1}}
\mathcommand\ifthenindexsymbol    [1]{\ifthensymbol    _{#1}}
\mathcommand\ifthenelseindexsymbol[1]{\ifthenelsesymbol_{#1}}
\newcommand\orsymbol    {\ident{or}}
\newcommand\andsymbol   {\ident{and}}
\newcommand\leqsymbol   {\ident{leq}}
\newcommand\lessymbol   {\ident{less}}
\newcommand\lexsymbol   {\ident{lex}}
\newcommand\acksymbol   {\ident{ack}}
\newcommand\switchsymbol{\ident{switch}}
\newcommand\swatchsymbol{\ident{swatch}}
\newcommand\diveinssymbol{\ident{div1}}
\newcommand\divzweisymbol{\ident{div2}}
\newcommand\divrestsymbol{\ident{divrest}}
\newcommand\diveinstailsymbol{\ident{div1tail}}
\newcommand\divzweitailsymbol{\ident{div2tail}}

\newcommand\turingmachinesymbol{\ident T}
\newcommand\terminatespsymbol  {\ident{terminatesp}}
\newcommand\statesymbol        {\ident{state}}
\newcommand\cmdsymbol          {\ident{cmd}}
\newcommand\nthsymbol          {\ident{nth}}
\newcommand\doublesymbol       {\ident{double}}

\newcommand\ppsymbol           {\ident{p}}
\newcommand\qpsymbol           {\ident{q}}
\newcommand\Epsymbol           {\ident{E}}
\newcommand\Ppsymbol           {\ident{P}}
\newcommand\Qpsymbol           {\ident{Q}}
\newcommand\Marriessymbol      {\ident{Marries}}
\newcommand\Lovessymbol        {\ident{Loves}}
\newcommand\StolenBysymbol     {\ident{StolenBy}}
\newcommand\Humansymbol        {\ident{Human}}
\newcommand\Evensymbol         {\ident{Even}}
\newcommand\Oddsymbol          {\ident{Odd}}
\newcommand\Primesymbol        {\ident{Prime}}
\newcommand\EveryPairsymbol   {\ident{EveryPair}}
\newcommand\Givesymbol         {\ident{Give}}
\newcommand\Fathersymbol       {\ident{Father}}
\newcommand\Elephantpsymbol    {\ident{Elephant}}
\newcommand\Flowerpsymbol    {\ident{Flower}}
\newcommand\Germanpsymbol      {\ident{German}}
\newcommand\Bicyclepsymbol     {\ident{Bicycle}}
\newcommand\Hugepsymbol        {\ident{Huge}}
\newcommand\Animalpsymbol      {\ident{Animal}}
\newcommand\Malepsymbol        {\ident{Male}}
\newcommand\Boypsymbol        {\ident{Boy}}
\newcommand\Girlpsymbol        {\ident{Girl}}
\newcommand\Femalepsymbol      {\ident{Female}}
\newcommand\Roundpsymbol       {\ident{Round}}
\newcommand\Quadrangularpsymbol{\ident{Quadrangular}}
\newcommand\Metpsymbol         {\ident{Met}}
\newcommand\Bishopsymbol       {\ident{Bishop}}
\newcommand\mindexsymbol[1]{\existsvari w{#1}}

\newcommand\nonnegpsymbol      {\ident{nonnegp}}
\newcommand\wellsymbol         {\ident{well}}
\newcommand\welltailsymbol     {\ident{welltail}}
\newcommand\varsymbol          {\ident{var}}
\newcommand\aritysymbol        {\ident{arity}}

\newcommand\whilesymbol        {\ident{while}}

\newcommand\nullsymbol         {\ident{null}}
\newcommand\hdsymbol           {\ident{hd}}
\newcommand\tlsymbol           {\ident{tl}}
\newcommand\insymbol           {\ident{in}}
\newcommand\applysymbol        {\ident{app}}
\newcommand\termsymbol         {\ident{term}}
\mathcommand\tightim{\longrightarrow}
\mathcommand\im{\ \tightim\ }
\mathcommand\rs{\:\rulesugar\:\:}
\mathcommand\rulesugar{\longleftarrow}

\mathcommand\doublepp[1]      {\doublesymbol   \beginargs{#1}\allargs}
\mathcommand\aritypp[1]      {\aritysymbol   \beginargs{#1}\allargs}
\mathcommand\lengthpp[1]      {\lengthsymbol   \beginargs{#1}\allargs}
\mathcommand\wellpp[1]      {\wellsymbol   \beginargs{#1}\allargs}
\mathcommand\welltailpp[1]      {\welltailsymbol   \beginargs{#1}\allargs}
\mathcommand\varpp[1]      {\varsymbol   \beginargs{#1}\allargs}
\mathcommand\divrestpp[2]    {\divrestsymbol\beginargs{#1}\separgs{#2}\allargs}
\mathcommand\diveinspp[2]    {\diveinssymbol\beginargs{#1}\separgs{#2}\allargs}
\mathcommand\divzweipp[3]    {\divzweisymbol\beginargs{#1}\separgs{#2}
\separgs{#3}\allargs}
\mathcommand\diveinstailpp[4]    {\diveinstailsymbol\beginargs{#1}\separgs{#2}
\separgs{#3}\separgs{#4}\allargs}
\mathcommand\divzweitailpp[6]    {\divzweitailsymbol\beginargs{#1}\separgs{#2}
\separgs{#3}\separgs{#4}\separgs{#5}\separgs{#6}\allargs}
\mathcommand\mbppp[2]         {\mbpsymbol   \beginargs{#1}\separgs{#2}\allargs}
\mathcommand\revpp[1]     
{\revsymbol\beginargs{#1}\allargs}
\mathcommand\revppi[2]     
{\revsymbol^{#1}\beginargs{#2}\allargs}
\mathcommand\revtailpp[2]     
{\revtailsymbol\beginargs{#1}\separgs{#2}\allargs}
\mathcommand\revtailppi[3]
{\revtailsymbol^{#1}\beginargs{#2}\separgs{#3}\allargs}
\mathcommand\Permpp[2]     
{\Permsymbol\beginargs{#1}\separgs{#2}\allargs}
\mathcommand\Permppi[3]
{\Permsymbol^{#1}\beginargs{#2}\separgs{#3}\allargs}
\mathcommand\appendpp[2]      
{\appendsymbol \beginargs{#1}\separgs{#2}\allargs}
\mathcommand\appendppi[3]      
{\appendsymbol^{#1}\beginargs{#2}\separgs{#3}\allargs}
\mathcommand\Everypp[2]      
{\Everysymbol \beginargs{#1}\separgs{#2}\allargs}
\mathcommand\RExistspp[1]      
{\RExistssymbol \beginargs{#1}\allargs}
\mathcommand\appendlongpp[2]      
{\appendsymbol\left(\begin{array}{@{}l@{}}{#1}\separgs\\{#2}\end{array}\right)}
\mathcommand\cnspp[2]         {\cnssymbol   \beginargs{#1}\separgs{#2}\allargs}
\mathcommand\cnsppi[3]       {\cnssymbol^{#1}\beginargs{#2}\separgs{#3}\allargs}
\mathcommand\cnsindexpp[3]
{\cnsindexsymbol{#1}\beginargs{#2}\separgs{#3}\allargs}
\mathcommand\dlpp[2]          {\dlsymbol    \beginargs{#1}\separgs{#2}\allargs}
\mathcommand\dloncepp[2]      {\dloncesymbol\beginargs{#1}\separgs{#2}\allargs}
\mathcommand\dlonceppi[3]{\dloncesymbol^{#1}\beginargs{#2}\separgs{#3}\allargs}
\mathcommand\rcpp[2]          {\rcsymbol    \beginargs{#1}\separgs{#2}\allargs}
\mathcommand\brpp[2]          {\brsymbol    \beginargs{#1}\separgs{#2}\allargs}
\mathcommand\orpp[2]          {\orsymbol    \beginargs{#1}\separgs{#2}\allargs}
\mathcommand\andpp[2]         {\andsymbol   \beginargs{#1}\separgs{#2}\allargs}
\mathcommand\shortcnspp[2]    {\csymbol     \beginargs{#1}\separgs{#2}\allargs}
\mathcommand\tightshortcnspp[2]
{\csymbol\beginargs{#1}\tightsepargs{#2}\allargs}
\mathcommand\spp[1]           {\ssymbol     \beginargs{#1}\allargs}
\mathcommand\sppiterated[2]   {\ssymbol^{#1}\beginargs{#2}\allargs}
\mathcommand\ppp[1]           {\psymbol     \beginargs{#1}\allargs}
\mathcommand\pppiterated[2]   {\psymbol^{#1}\beginargs{#2}\allargs}
\mathcommand\zeropp           {\ident 0}
\mathcommand\Julietpp         {\ident{Juliet}}
\mathcommand\Romeopp          {\ident{Romeo}}
\mathcommand\Ipp              {\ident I}
\mathcommand\onepp            {\ident1}
\mathcommand\twopp            {\ident2}
\mathcommand\threepp          {\ident3}
\mathcommand\invertpp[1]      {\invertsymbol\beginargs{#1}\allargs}
\mathcommand\invpp[1]         {\invsymbol\beginargs{#1}\allargs}
\mathcommand\abspp[1]         {\abssymbol\beginargs{#1}\allargs}
\mathcommand\naturalspp[1]    {\naturalssymbol\beginargs{#1}\allargs}
\mathcommand\gensympp[1]      {\gensymsymbol\beginargs{#1}\allargs}
\mathcommand\nilpp            {\ident{nil}}
\mathcommand\falsepp          {\ident{false}}
\mathcommand\truepp           {\ident{true}}
\mathcommand\FALSEpp          {\ident{FALSE}}
\mathcommand\TRUEpp           {\ident{TRUE}}
\mathcommand\weirdppp         {\ident{weirdp}}
\mathcommand\ambigppp         {\ident{ambigp}}
\mathcommand\zeropredicatepp[1]{\zeropredicatesymbol\beginargs{#1}\allargs}
\mathcommand\cppeins       [1]{\csymbol     \beginargs{#1}\allargs}
\mathcommand\cppzwei       [2]{\csymbol\beginargs{#1}\separgs{#2}\allargs}
\mathcommand\eppeins       [1]{\esymbol     \beginargs{#1}\allargs}
\mathcommand\fppeins       [1]{\fsymbol     \beginargs{#1}\allargs}
\mathcommand\fppeinsindex  [2]{\fsymbol_{#1}\beginargs{#2}\allargs}
\mathcommand\fppeinsiterated[2]{\fsymbol^{#1}\beginargs{#2}\allargs}
\mathcommand\gppeins       [1]{\gsymbol     \beginargs{#1}\allargs}
\mathcommand\gppzwei       [2]{\gsymbol     \beginargs{#1}\separgs{#2}\allargs}
\mathcommand\hppeins       [1]{\hsymbol     \beginargs{#1}\allargs}
\mathcommand\kppeins       [1]{\ksymbol     \beginargs{#1}\allargs}
\mathcommand\appzero          {\ident a}
\mathcommand\bppzero          {\ident b}
\mathcommand\cppzero          {\ident c}
\mathcommand\dppzero          {\ident d}
\mathcommand\eppzero          {\ident e}
\mathcommand\eqindexpp[3]{\eqindexsymbol{#1}\beginargs{#2}\separgs{#3}\allargs}
\mathcommand\ifthenindexpp
[3]{\ifthenindexsymbol{#1}\beginargs{#2}\separgs{#3}\allargs}
\mathcommand\ifthenelseindexpp
[4]{\ifthenelseindexsymbol{#1}\beginargs{#2}\separgs{#3}\separgs{#4}\allargs}
\mathcommand\eqpp[2]{\eqsymbol\beginargs{#1}\separgs{#2}\allargs}
\mathcommand\leqpp[2]{\leqsymbol\beginargs{#1}\separgs{#2}\allargs}
\mathcommand\lespp[2]{\lessymbol\beginargs{#1}\separgs{#2}\allargs}
\mathcommand\lexpp[3]{\lexsymbol\beginargs{#1}\separgs{#2}\separgs{#3}\allargs}
\mathcommand\ackpp[2]{\acksymbol\beginargs{#1}\separgs{#2}\allargs}
\mathcommand\switchpp[1]{\switchsymbol\beginargs{#1}\allargs}
\mathcommand\swatchpp[1]{\swatchsymbol\beginargs{#1}\allargs}
\mathcommand\whilepp[2]{\whilesymbol\beginargs{#1}\separgs{#2}\allargs}
\mathcommand\nullpp[1]{\nullsymbol\beginargs{#1}\allargs}
\mathcommand\nullppiterated[2]{\nullsymbol^{#1}\beginargs{#2}\allargs}
\mathcommand\hdpp[1]{\hdsymbol\beginargs{#1}\allargs}
\mathcommand\hdppiterated[2]{\hdsymbol^{#1}\beginargs{#2}\allargs}
\mathcommand\tlpp[1]{\tlsymbol\beginargs{#1}\allargs}
\mathcommand\tlppiterated[2]{\tlsymbol^{#1}\beginargs{#2}\allargs}
\mathcommand\inpp[2]{\insymbol\beginargs{#1}\separgs{#2}\allargs}
\mathcommand\inppiterated[3]{\insymbol^{#1}\beginargs{#2}\separgs{#3}\allargs}
\mathcommand\applypp[2]{\applysymbol\beginargs{#1}\separgs{#2}\allargs}
\mathcommand\applyppiterated
[3]{\applysymbol^{#1}\beginargs{#2}\separgs{#3}\allargs}
\mathcommand\termpp[2]{\termsymbol\beginargs{#1}\separgs{#2}\allargs}
\mathcommand\setpp[1]{\set\beginargs{#1}\allargs}

\mathcommand\Tpp[6]{\turingmachinesymbol\beginargs{#1}\separgs{#2}\separgs
{#3}\separgs{#4}\separgs{#5}\separgs{#6}\allargs}
\mathcommand\Tppseven[7]{\turingmachinesymbol\beginargs{#1}\separgs{#2}\separgs
{#3}\separgs{#4}\separgs{#5}\separgs{#6}\separgs{#7}\allargs}
\mathcommand\foreverppp[6]{\ident{foreverp}\beginargs{#1}\separgs{#2}\separgs
{#3}\separgs{#4}\separgs{#5}\separgs{#6}\allargs}
\mathcommand\terminatesppp[6]{\terminatespsymbol\beginargs{#1}\separgs
{#2}\separgs{#3}\separgs{#4}\separgs{#5}\separgs{#6}\allargs}
\mathcommand\terminatespppone[1]{\terminatespsymbol \beginargs{#1}\allargs}
\mathcommand\statepp
[3]{\statesymbol\beginargs{#1}\separgs{#2}\separgs{#3}\allargs}
\mathcommand\tightstatepp
[3]{\statesymbol\beginargs{#1}\tightsepargs{#2}\tightsepargs{#3}\allargs}
\mathcommand\cmdpp
[3]{\cmdsymbol  \beginargs{#1}\separgs{#2}\separgs{#3}\allargs}
\mathcommand\tightcmdpp
[3]{\cmdsymbol  \beginargs{#1}\tightsepargs{#2}\tightsepargs{#3}\allargs}
\mathcommand\stoppp           {\ident{stop}}
\mathcommand\leftpp           {\ident{left}}
\mathcommand\rightpp          {\ident{right}}
\mathcommand\nthpp         [2]{\nthsymbol  \beginargs{#1}\separgs{#2}\allargs}
\mathcommand\pppp          [1]{\ppsymbol\beginargs{#1}            \allargs}
\mathcommand\qppp          [2]{\qpsymbol\beginargs{#1}\separgs{#2}\allargs}
\mathcommand\Eppp          [1]{\Epsymbol\beginargs{#1}            \allargs}
\mathcommand\Epppzwei      [2]{\Epsymbol\beginargs{#1}\separgs{#2}\allargs}
\mathcommand\Pppp          [1]{\Ppsymbol\beginargs{#1}            \allargs}
\mathcommand\Ppppdrei      
[3]{\Ppsymbol\beginargs{#1}\separgs{#2}\separgs{#3}\allargs}
\mathcommand\Ppppvier
[4]{\Ppsymbol\beginargs{#1}\separgs{#2}\separgs{#3}\separgs{#4}\allargs}
\mathcommand\Qppp          [2]{\Qpsymbol\beginargs{#1}\separgs{#2}\allargs}
\mathcommand\Qpppeins      [1]{\Qpsymbol\beginargs{#1}\allargs}
\mathcommand\Qpppdrei      
[3]{\Qpsymbol\beginargs{#1}\separgs{#2}\separgs{#3}\allargs}
\mathcommand\Fatherpp      [2]{\Fathersymbol\beginargs{#1}\separgs{#2}\allargs}
\mathcommand\Marriespp     [2]{\Marriessymbol\beginargs{#1}\separgs{#2}\allargs}
\mathcommand\Lovespp       [2]{\Lovessymbol\beginargs{#1}\separgs{#2}\allargs}
\mathcommand\StolenBypp    [2]
{\StolenBysymbol\beginargs{#1}\separgs{#2}\allargs}
\mathcommand\Humanpp       [1]{\Humansymbol\beginargs{#1}\allargs}
\mathcommand\Evenpp        [1]{\Evensymbol\beginargs{#1}\allargs}
\mathcommand\Evenppi       [2]{\Evensymbol^{#1}\beginargs{#2}\allargs}
\mathcommand\Oddpp         [1]{\Oddsymbol\beginargs{#1}\allargs}
\mathcommand\Primepp       [1]{\Primesymbol\beginargs{#1}\allargs}
\mathcommand\EveryPairpp  [2]{\EveryPairsymbol\beginargs{#1}\separgs
{#2}\allargs}
\mathcommand\mindexppeins  [2]{\mindexsymbol{#1}\beginargs{#2}\allargs}
\mathcommand\Givepp        [3]{\Givesymbol
\beginargs{#1}\separgs{#2}\separgs{#3}\allargs}
\mathcommand\mindexppzwei  [3]{\mindexsymbol
{#1}\beginargs{#2}\separgs{#3}\allargs}
\mathcommand\mindexppdrei  [4]{\mindexsymbol
{#1}\beginargs{#2}\separgs{#3}\separgs{#4}\allargs}

\mathcommand\nonnegppp     [1]{\nonnegpsymbol\beginargs{#1}\allargs}

\mathcommand\anonymouscsymbol{c}
\mathcommand\anonymouscindexsymbol[1]{\anonymouscsymbol_{#1}}
\mathcommand\anonymousfsymbol{f}
\mathcommand\anonymouscpp
[2]{\anonymouscsymbol\beginargs{#1}\separgs\ldots\separgs{#2}\allargs}
\mathcommand\anonymouscindexpp
[3]{\anonymouscindexsymbol{#1}\beginargs{#2}\separgs\ldots\separgs{#3}\allargs}
\mathcommand\anonymousfpp
[2]{\anonymousfsymbol\beginargs{#1}\separgs\ldots\separgs{#2}\allargs}
\mathcommand\coerceindexpp[3]{[#3]_{#1}^{#2}}

\mathcommand\Elephantppp    [1]{\Elephantpsymbol\beginargs{#1}\allargs}
\mathcommand\Flowerppp      [1]{\Flowerpsymbol  \beginargs{#1}\allargs}
\mathcommand\Bicycleppp     [1]{\Bicyclepsymbol \beginargs{#1}\allargs}
\mathcommand\Germanppp      [1]{\Germanpsymbol  \beginargs{#1}\allargs}
\mathcommand\Hugeppp        [1]{\Hugepsymbol    \beginargs{#1}\allargs}
\mathcommand\Animalppp      [1]{\Animalpsymbol  \beginargs{#1}\allargs}
\mathcommand\Maleppp        [1]{\Malepsymbol    \beginargs{#1}\allargs}
\mathcommand\Boyppp         [1]{\Boypsymbol     \beginargs{#1}\allargs}
\mathcommand\Girlppp        [1]{\Girlpsymbol    \beginargs{#1}\allargs}
\mathcommand\Femaleppp      [1]{\Femalepsymbol  \beginargs{#1}\allargs}
\mathcommand\Roundppp       [1]{\Roundpsymbol   \beginargs{#1}\allargs}
\mathcommand\Bishoppp       [1]{\Bishopsymbol   \beginargs{#1}\allargs}
\mathcommand\Quadrangularppp[1]{\Quadrangularpsymbol  \beginargs{#1}\allargs}
\mathcommand\Metppp[2]{\Metpsymbol     \beginargs{#1}\separgs{#2}\allargs}

\newcommand\bound     {{\rm bound}}
\newcommand\free      {{\rm free}}

\mathcommand\Vtripleindex[3]{\V\!_{{#1},\,{#2},\,{#3}}}
\mathcommand\Vdoubleindex[2]{\V\!_{{#1},\,{#2}}}
\mathcommand\Vsingleindex[1]{\V\!_{{#1}}}

\mathcommand\Erel[1]{\Gammaoffont\!_{#1}}
\mathcommand\Urel[1]{\Deltaoffont_{#1}}

\newcommand\exRsub  {\math R-sub\-sti\-tu\-tion  on \nolinebreak\Vsome}

\newcommand\cc{choice-condition}
\newcommand\CC{Choice-Condition}

\newcommand\vc{vari\-able-con\-di\-tion}
\newcommand\VC{Vari\-able-Con\-di\-tion}

\newcommand\semanticobject{\mbox{\math\Sigmaoffont-}struc\-ture}

\mathcommand\theRprimefromstrongtoweak{
  \inparenthesesinlinetight{
     \domres\id{\Vwall\cup\Vsome\setminus\RAN\varsigma}
     \nottight{\nottight\uplus}
     \reverserelation\varsigma
  }
  \nottight{\circ}
  \ranres
    {\transclosureinline R}
    {\Vwall\cup\Vsome\setminus\RAN\varsigma}
  \nottight{\nottight{\nottight{\uplus}}}
  \Vsome\tighttimes\Vsall
}

\mathcommand\deltaminus{\delta^-}
\mathcommand\deltaplus{\delta^+}
\mathcommand\deltaplusplus{\delta^{+^+}}
\mathcommand\deltastar{\delta^*}
\mathcommand\deltastarstar{\delta^{*^*}}
\newcommand\varihelper[1]{free \discretionary
{\mbox{\math{#1}-vari-}}{\mbox{able}}{\mbox{\math{#1}-variable}}}

\newcommand\fev {\varihelper\gamma}
\newcommand\fuv {\varihelper\delta}

\mathcommand\Vall     {\Vsingleindex\indexdelta         }
\mathcommand\Vwall    {\Vsingleindex\indexdeltaminu     }
\mathcommand\Vsall    {\Vsingleindex\indexdeltaplus     }
\mathcommand\Vgsome   {\Vsingleindex\indexgammaplus     }
\mathcommand\Vsome    {\Vsingleindex\indexgamma         }
\mathcommand\Vfree    {\Vsingleindex\indexfree          }
\mathcommand\Vbound   {\Vsingleindex\indexbound         }
\mathcommand\Vsomesall{\Vsingleindex\indexgammadeltaplus}

\mathapplycommand\VARall      {\VARsingleindex\indexdelta         }
\mathapplycommand\VARwall     {\VARsingleindex\indexdeltaminu     }
\mathapplycommand\VARsall     {\VARsingleindex\indexdeltaplus     }
\mathapplycommand\VARgsome    {\VARsingleindex\indexgammaplus     }
\mathapplycommand\VARsome     {\VARsingleindex\indexgamma         }
\mathapplycommand\VARfree     {\VARsingleindex\indexfree          }
\mathapplycommand\VARbound    {\VARsingleindex\indexbound         }
\mathapplycommand\VARsomesall {\VARsingleindex\indexgammadeltaplus}
\mathcommand\displayVARsall[1]{\VARsingleindex\indexdeltaplus
\!\!\!\:\left(\begin{array}{@{}c@{}}#1\end{array}\right)}

\mathcommand\rigidvari     [2]{#1_{#2}^\indexgammadeltaplus}
\mathcommand\existsvari    [2]{#1_{#2}^\indexgamma    }
\mathcommand\forallvari    [2]{#1_{#2}^\indexdelta    }
\mathcommand\freevari      [2]{#1_{#2}^\indexfree     }
\mathcommand\wforallvari   [2]{#1_{#2}^\indexdeltaminu}
\mathcommand\sforallvari   [2]{#1_{#2}^\indexdeltaplus}
\mathcommand\gexistsvari   [2]{#1_{#2}^\indexgammaplus}
\mathcommand\boundvari     [2]{#1_{#2}}
\mathcommand\vari          [2]{#1_{#2}}
\mathcommand\wforallvarilow[2]{#1_{#2}^
{\raisebox{-.82ex}{\math\indexdeltaminu}}}

\newcommand\indexhelper[1]{{\scriptscriptstyle#1\:\!\!}}
\newcommand\indexdeltaplus
{\indexhelper{\delta^{\raisebox{-.17ex}{\fvesf\hskip-0.14em +}}}}
\newcommand\indexdeltaminu
{\indexhelper{\delta^{\mbox{\fvesf\hskip-0.14em\rule[.2ex]{.7em}{.15ex}}}}}
\newcommand\indexgammaplus
{\indexhelper{\gamma^{\mbox{\fvesf\hskip-0.14em +}}}}
\newcommand\indexgammadeltaplus
{\indexhelper{\gamma\delta^{\raisebox{-.17ex}{\fvesf\hskip-0.14em +}}}}

\newcommand\indexdelta{\indexhelper\delta}
\newcommand\indexgamma{\indexhelper\gamma}
\newcommand\indexfree
{{\scriptscriptstyle\free}}
\newcommand\indexbound
{{\scriptscriptstyle\bound}}

\newcommand\Wellfsymb{\ident{Wellf}}
\mathapplycommand\Wellfpp{\Wellfsymb}

\mathcommand\beginargs{(}
\mathcommand\allargs  {)}
\mathcommand\separgs  {,\,}
\mathcommand\tightsepargs{,}

\mathcommand\minusppnoparentheses  [2]{{#1}\,\minussymbol\,{#2}}
\mathcommand\tightminusppnoparentheses  [2]{{#1}\minussymbol{#2}}
\mathcommand\divideppnoparentheses [2]{{#1}\,\dividesymbol\,{#2}}
\mathcommand\plusppnoparentheses   [2]{{#1}\,\plussymbol \,{#2}}
\mathcommand\plusppnoparenthesesi  [3]{{#2}\,\plussymbol^{#1}\,{#3}}
\mathcommand\tightplusppnoparentheses   [2]{{#1}\plussymbol{#2}}
\mathcommand\timesppnoparentheses  [2]{{#1}\,\timessymbol\,{#2}}
\mathcommand\undppnoparentheses    [2]{{#1}\und            {#2}}
\mathcommand\oderppnoparentheses   [2]{{#1}\oder           {#2}}
\mathcommand\impliesppnoparentheses[2]{{#1}\implies        {#2}}
\mathcommand\leqinfixppnoparentheses[2]{{#1}\,\tight\leq\,{#2}}
\mathcommand\geqinfixppnoparentheses[2]{{#1}\,\tight\geq\,{#2}}
\mathcommand\dividepp [2]{(\divideppnoparentheses {#1}{#2})}
\mathcommand\minuspp  [2]{(\minusppnoparentheses  {#1}{#2})}
\mathcommand\pluspp   [2]{(\plusppnoparentheses   {#1}{#2})}
\mathcommand\timespp  [2]{(\timesppnoparentheses  {#1}{#2})}
\mathcommand\undpp    [2]{(\undppnoparentheses    {#1}{#2})}
\mathcommand\oderpp   [2]{(\oderppnoparentheses   {#1}{#2})}
\mathcommand\impliespp[2]{(\impliesppnoparentheses{#1}{#2})}

\usepackage{url}

\newlength{\mybibitemsep}
\newcommand\setmybibitemsep[1]{\setlength{\mybibitemsep}{#1}}

\setmybibitemsep{0.5\topsep}

\newcommand\includenetreferences{y}

\newcommand\referencessize{\normalsize}

\newcommand\mybibbaselinestretch{0.96}

\newcommand\mybibsection[1]{\section*{#1}\if 
\addcontentslineofbibsection\addcontentsline{toc}{section}{#1}\fi}
\newcommand\addcontentslineofbibsection{y}

\newcommand\mybibtitle[1]{{\em #1\@.}}
\newcommand\mybibhardnodate
[1]{{\def\myfirstarg{#1}\ifx\empty\myfirstarg{}\else#1.\ \fi}}
\newcommand\mybibsoft
[2]{{\def\myfirstarg{#1}\def\mysecondarg{#2}\ifx\empty\myfirstarg{}\else
\if y\includenetreferences\writemynetsource#1\else\if x\includenetreferences
\ifx\empty\mysecondarg\writemynetsource#1\else{}\fi\else
\if n\includenetreferences{}\else\typeout
{WARNING includenetreferences from headerbiblio.tex has silly definition}
\fi\fi\fi\fi}}

\def\writemynetsource[#1,#2,#3,#4://#5]{{\tt\sloppy\ 
\url{#4://#5} \discretionary
{(\ignorespaces#2\,\ignorespaces#1\mbox{$\!$},\mbox{$\!$}}%
{\ignorespaces#3)\mbox{$\!$}.}%
{(\ignorespaces#2\,\ignorespaces#1\mbox{$\!$},\,\ignorespaces
#3)\mbox{$\!$}.}}}

\catcode`\@=11

\newcommand\resetlongbibstyle{%
\def\bibitem{\@ifnextchar[\@lbibitem\@bibitem}
\def\@lbibitem[##1]##2{\item[\@biblabel{##1}\hfill]\if@filesw
      {\let\protect\noexpand
       \immediate
       \write\@auxout{\string\bibcite{##2}{##1}}}\fi\ignorespaces}
\def\@bibitem##1{\item\if@filesw \immediate\write\@auxout
       {\string\bibcite{##1}{\the\value{\@listctr}}}\fi\ignorespaces}
\def\bibcite{\@newl@bel b}
\let\citation\@gobble
\let\bibdata=\@gobble
\let\bibstyle=\@gobble
\def\bibliography##1{%
  \if@filesw
    \immediate\write\@auxout{\string\bibdata{##1}}%
  \fi
  \@input@{\jobname.bbl}}
\def\bibliographystyle##1{%
  \ifx\@begindocumenthook\@undefined\else
    \expandafter\AtBeginDocument
  \fi
    {\if@filesw
       \immediate\write\@auxout{\string\bibstyle{##1}}%
     \fi}}
\def\nocite##1{\@bsphack
  \@for\@citeb:=##1\do{%
    \edef\@citeb{\expandafter\@firstofone\@citeb}%
    \if@filesw\immediate\write\@auxout{\string\citation{\@citeb}}\fi
    \@ifundefined{b@\@citeb}{\G@refundefinedtrue
        \@latex@warning{Citation `\@citeb' undefined}}{}}%
  \@esphack}
\expandafter\let\csname b@*\endcsname\@empty
\DeclareRobustCommand\cite{%
  \@ifnextchar [{\@tempswatrue\@citex}{\@tempswafalse\@citex[]}}
\DeclareRobustCommand\citet{%
  \@ifnextchar [{\@tempswatrue\@citex}{\@tempswafalse\@citex[]}}
\def\@tempswafalse{\let\if@tempswa\iffalse}
\def\@tempswatrue{\let\if@tempswa\iftrue}
\let\if@tempswa\iffalse
\def\@cite##1##2{##1\if@tempswa , ##2\fi}
\def\@citex[##1]##2{%
  \let\@citea\@empty
  \@cite{\@for\@citeb:=##2\do
    {\@citea\def\@citea{,\penalty\@m\ }%
     \edef\@citeb{\expandafter\@firstofone\@citeb}%
     \if@filesw\immediate\write\@auxout{\string\citation{\@citeb}}\fi
     \@ifundefined{b@\@citeb}{\mbox{\reset@font\bfseries ?}%
       \G@refundefinedtrue
       \@latex@warning
         {Citation `\@citeb' on page \thepage \space undefined}}%
       {\csname b@\@citeb\endcsname}}}{##1}}
\def\@biblabel##1{##1}
\def\mybibitem##1##2##3##4##5##6##7##8##9
{\item
 \if@filesw
      {\let\protect\noexpand
       \immediate
       \write\@auxout{\string\bibcite{##1}{##7\discretionary
{}{}{\,}(##3##9)}}}\fi\ignorespaces
##2 (##3##9). \mybibtitle{##4} \mybibhardnodate{##5}\mybibsoft
{##6}{##5}$\!\!$\par}
\def\thebibliography##1{\mybibsection{\refname}%
\@mkboth{\uppercase{\refname}}{\uppercase{\refname}}
\def\baselinestretch{\mybibbaselinestretch}%
\list{}{\labelwidth\z@
    \leftmargin 1.5pc
    \itemindent-\leftmargin}
    \referencessize
    \parindent\z@
    \parskip\mybibitemsep\relax
    \def\newblock{\hskip .11em plus .33em minus .07em}
    \sloppy\clubpenalty4000\widowpenalty4000
    \sfcode`\.=1000\relax}
\let\endthebibliography=\endlist
}

\newcommand\resetshortbibstyle{%
\def\bibitem{\@ifnextchar[\@lbibitem\@bibitem}
\def\@lbibitem[##1]##2{\item[\@biblabel{##1}\hfill]\if@filesw
      {\let\protect\noexpand
       \immediate
       \write\@auxout{\string\bibcite{##2}{##1}}}\fi\ignorespaces}
\def\@bibitem##1{\item\if@filesw \immediate\write\@auxout
       {\string\bibcite{##1}{\the\value{\@listctr}}}\fi\ignorespaces}
\def\bibcite{\@newl@bel b}
\let\citation\@gobble
\def\@citex[##1]##2{%
  \let\@citea\@empty
  \@cite{\@for\@citeb:=##2\do
    {\@citea\def\@citea{,\penalty\@m\ }%
     \edef\@citeb{\expandafter\@firstofone\@citeb}%
     \if@filesw\immediate\write\@auxout{\string\citation{\@citeb}}\fi
     \@ifundefined{b@\@citeb}{\mbox{\reset@font\bfseries ?}%
       \G@refundefinedtrue
       \@latex@warning
         {Citation `\@citeb' on page \thepage \space undefined}}%
       {\hbox{\csname b@\@citeb\endcsname}}}}{##1}}
\let\bibdata=\@gobble
\let\bibstyle=\@gobble
\def\bibliography##1{%
  \if@filesw
    \immediate\write\@auxout{\string\bibdata{##1}}%
  \fi
  \@input@{\jobname.bbl}}
\def\bibliographystyle##1{%
  \ifx\@begindocumenthook\@undefined\else
    \expandafter\AtBeginDocument
  \fi
    {\if@filesw
       \immediate\write\@auxout{\string\bibstyle{##1}}%
     \fi}}
\def\nocite##1{\@bsphack
  \@for\@citeb:=##1\do{%
    \edef\@citeb{\expandafter\@firstofone\@citeb}%
    \if@filesw\immediate\write\@auxout{\string\citation{\@citeb}}\fi
    \@ifundefined{b@\@citeb}{\G@refundefinedtrue
        \@latex@warning{Citation `\@citeb' undefined}}{}}%
  \@esphack}
\expandafter\let\csname b@*\endcsname\@empty
\def\@cite##1##2{[{##1\if@tempswa , ##2\fi}]}
\def\@biblabel##1{[##1]}
\DeclareRobustCommand\cite{%
  \@ifnextchar [{\@tempswatrue\@citex}{\@tempswafalse\@citex[]}}
\def\@tempswafalse{\let\if@tempswa\iffalse}
\def\@tempswatrue{\let\if@tempswa\iftrue}
\let\if@tempswa\iffalse
\def\mybibitem##1##2##3##4##5##6##7##8##9
{\bibitem[##8##9]{##1}##2 (##3). 
\mybibtitle{##4} \mybibhardnodate{##5}\mybibsoft{##6}{##5}$\!\!$\par}
\def\thebibliography##1{\mybibsection{\refname}%
\@mkboth{\uppercase{\refname}}{\uppercase{\refname}}%
\def\baselinestretch{\mybibbaselinestretch}%
\referencessize
\list
 {[\arabic{enumi}]}
 {\settowidth\labelwidth{[##1]}\leftmargin\labelwidth
 \advance\leftmargin\labelsep
 \usecounter{enumi}}
 \parskip\mybibitemsep\relax
 \def\newblock{\hskip .11em plus .33em minus .07em}
 \sloppy\clubpenalty4000\widowpenalty4000
 \sfcode`\.=1000\relax}
\let\endthebibliography=\endlist
}

\newcommand\resetnumberbibstyle{%
\def\bibitem{\@ifnextchar[\@lbibitem\@bibitem}
\def\@lbibitem[##1]##2{\item[\@biblabel{##1}\hfill]\if@filesw
      {\let\protect\noexpand
       \immediate
       \write\@auxout{\string\bibcite{##2}{##1}}}\fi\ignorespaces}
\def\@bibitem##1{\item\if@filesw \immediate\write\@auxout
       {\string\bibcite{##1}{\the\value{\@listctr}}}\fi\ignorespaces}
\def\bibcite{\@newl@bel b}
\let\citation\@gobble
\def\@citex[##1]##2{%
  \let\@citea\@empty
  \@cite{\@for\@citeb:=##2\do
    {\@citea\def\@citea{,\penalty\@m\ }%
     \edef\@citeb{\expandafter\@firstofone\@citeb}%
     \if@filesw\immediate\write\@auxout{\string\citation{\@citeb}}\fi
     \@ifundefined{b@\@citeb}{\mbox{\reset@font\bfseries ?}%
       \G@refundefinedtrue
       \@latex@warning
         {Citation `\@citeb' on page \thepage \space undefined}}%
       {\hbox{\csname b@\@citeb\endcsname}}}}{##1}}
\let\bibdata=\@gobble
\let\bibstyle=\@gobble
\def\bibliography##1{%
  \if@filesw
    \immediate\write\@auxout{\string\bibdata{##1}}%
  \fi
  \@input@{\jobname.bbl}}
\def\bibliographystyle##1{%
  \ifx\@begindocumenthook\@undefined\else
    \expandafter\AtBeginDocument
  \fi
    {\if@filesw
       \immediate\write\@auxout{\string\bibstyle{##1}}%
     \fi}}
\def\nocite##1{\@bsphack
  \@for\@citeb:=##1\do{%
    \edef\@citeb{\expandafter\@firstofone\@citeb}%
    \if@filesw\immediate\write\@auxout{\string\citation{\@citeb}}\fi
    \@ifundefined{b@\@citeb}{\G@refundefinedtrue
        \@latex@warning{Citation `\@citeb' undefined}}{}}%
  \@esphack}
\expandafter\let\csname b@*\endcsname\@empty
\def\@cite##1##2{[{##1\if@tempswa , ##2\fi}]}
\def\@biblabel##1{[##1]}
\DeclareRobustCommand\cite{%
  \@ifnextchar [{\@tempswatrue\@citex}{\@tempswafalse\@citex[]}}
\def\@tempswafalse{\let\if@tempswa\iffalse}
\def\@tempswatrue{\let\if@tempswa\iftrue}
\let\if@tempswa\iffalse
\def\mybibitem##1##2##3##4##5##6##7##8##9
{\bibitem{##1}##2 (##3). 
\mybibtitle{##4} \mybibhardnodate{##5}\mybibsoft{##6}{##5}$\!\!$\par}
\def\thebibliography##1{\mybibsection{\refname}%
\@mkboth{\uppercase{\refname}}{\uppercase{\refname}}
\def\baselinestretch{\mybibbaselinestretch}%
\referencessize
\list
 {[\arabic{enumi}]}
 {\settowidth\labelwidth{[##1]}\leftmargin\labelwidth
 \advance\leftmargin\labelsep
 \usecounter{enumi}}
 \parskip\mybibitemsep\relax
 \def\newblock{\hskip .11em plus .33em minus .07em}
 \sloppy\clubpenalty4000\widowpenalty4000
 \sfcode`\.=1000\relax}
\let\endthebibliography=\endlist
}

\newcommand\setlongbibstyle  {\AtBeginDocument{\resetlongbibstyle  }}

\catcode`\@=12

\mathcommand\primweig{{\rm Weight}}
\mathcommand\weig {{\rm Weight}(\V)}
\mathcommand\Xweig{{\rm Weight}(\X)}
\mathcommand\Yweig{{\rm Weight}(\Y)}
\mathcommand\algebraweig{{\rm Weight}(\Vsig^\algebra)}

\mathcommand\Anoprime{A}
\mathcommand\Bnoprime{B}
\mathcommand\Cnoprime{C}
\mathcommand\Dnoprime{D}
\mathcommand\Enoprime{E}
\mathcommand\Bprime{B'}
\mathcommand\Dprime{D'}

\mathcommand\Meins{\aleph}
\mathcommand\Mzwei{\beth}
\mathcommand\Mdrei{\daleth}
\mathcommand\Meinsprime{\Meins'}
\mathcommand\Mzweiprime{\Mzwei'}
\mathcommand\Mdreiprime{\Mdrei'}

\mathcommand\FMeins{(\Feins,\Meins)}
\mathcommand\FMzwei{(\Fzwei,\Mzwei)}
\mathcommand\FMdrei{(\Fdrei,\Mdrei)}
\mathcommand\FMeinsprime{(\Feinsprime,\Meinsprime)}
\mathcommand\FMdreiprime{(\Fdreiprime,\Mdreiprime)}

\mathcommand\Beins{\kappa}
\mathcommand\Bzwei{\chi}

\mathcommand\Ieins{\algebra}
\mathcommand\Izwei{\balgebra}

\mathcommand\Seins{\tau}
\mathcommand\Seinsprime{\tau'}
\mathcommand\Szwei{\pi}
\mathcommand\iS   {\iota\!_\algebra}

\mathcommand\SBeins{\Seins,\Beins}
\mathcommand\SBzwei{\Szwei,\Bzwei}

\mathcommand\SIBeins{\Seins,\Ieins
}
\mathcommand\SIBzwei{\Szwei,\Izwei
}

\mathcommand\MSBeins{(\Meins\Seins,\Beins)}
\mathcommand\MSBzwei{(\Mzwei\Szwei,\Bzwei)}

\mathcommand\Ceins{(\Feins,\SIBeins)}
\mathcommand\Czwei{(\Fzwei,\SIBzwei)}

\mathcommand\liteins{\lambda  }
\mathcommand\litzwei{\lambda' }
\mathcommand\litdrei{\lambda''}

\mathapplycommand\GENDOM{\rm GENDOM}

\mathcommand\sepfw{;\ \ }

\newcommand \tallall   {\forall}
\newcommand \tallsome  {\exists}
\newcommand \all       {{\scriptscriptstyle\tallall }}
\newcommand \some      {{\scriptscriptstyle\tallsome}}

\mathcommand\boundfreeallsome{\{\bound,\free,\tallall,\tallsome\}}

\mathcommand\Vfreesig  {\Vdoubleindex\free \SIG }
\mathcommand\Vfreecons {\Vdoubleindex\free \CONS}
\mathcommand\Vallcons  {\Vdoubleindex\all  \CONS}
\mathcommand\Vallsig   {\Vdoubleindex\all  \SIG }
\mathcommand\Vsomesig  {\Vdoubleindex\some \SIG }
\mathcommand\Vboundsig {\Vdoubleindex\bound\SIG }
\mathcommand\Vboundcons{\Vdoubleindex\bound\CONS}
\mathcommand\Vboundsigindex [1]{\Vtripleindex\bound\SIG {#1}}
\mathcommand\Vboundconsindex[1]{\Vtripleindex\bound\CONS{#1}}

\mathcommand\sigVfree{\Vfreesig\tight\uplus\Vfreecons}

\mathdoubleapplycommand\reduce{\rm SUPERPOSE}
\mathtripleapplycommand\narrow{\rm NARROW}

\newcommand\REFUT{{\rm refut}}
\mathcommand\grtrefut{>_{\!\!_\REFUT}}

\mathcommand\wrule[6]{\begin{array}[c]{@{}llll@{}}(#1&,#2&,#3&)\\\cline
{1-4}(#4&,#5&,#6&)\\\end{array}}

\mathcommand\criticalpeaklongform
{((t_0,C_0,\Lambda_0),\ (t_1,C_1,\Lambda_1),\ \hat t,\ \sigma,\ p)}

\mathcommand\foresteins{F}
\mathcommand\forestzwei{F'}
\mathcommand\foresteinspair{(F ,R )}
\mathcommand\forestzweipair{(F',R')}
\newcommand\Opensymbol{{\rm Goals}}
\mathcommand\Open[1]{\Openofset{\{#1\}}}
\mathapplycommand\Openofset{\Opensymbol}

\newcommand\orelse{or \nolinebreak else:}

\mathcommand\cextriple{\pair e\algebra}
\renewcommand\exRsub{existential \math R-sub\-sti\-tu\-tion}
\newcommand\expansionrule[6]{\LINEmath{~~~~~~~~\begin{array}[c]{c}#1\\\hline
\mediumheadroom#2\\\end{array}~~~~~~~~~~#3~~~~#4~~~~#5}}
\newcommand\strongexpansionrule[6]{\LINEmath{~~~~~~~~\begin
{array}[c]{c}#1\\\hline\mediumheadroom#2\\\end{array}~~~~~~~~~~\begin
{array}[c]{l}#3\\#4\\#5\end{array}}}
\newcommand\strongvalidityinA[2]{\math{#2}-strong \math{#1}-validity}
\newcommand\stronglyvalidinA[2]{\math{#2}-strongly \math{#1}-valid}
\newcommand\stronglyvalid[3]{\math{#3}-strongly \pair{#1}{#2}-valid}

\mathcommand\synconseins{S}
\mathcommand\synconszwei{S'}
\mathcommand\infoeins{\pi}
\mathcommand\infozwei{\pi'}
\mathcommand\synconsinfoeins{(\synconseins,\infoeins)}
\mathcommand\synconsinfozwei{(\synconszwei,\infozwei)}
\mathcommand\inductionorderinglesssim{\lesssim_{\rm ind}}
\mathcommand\inductionorderinggtrsim {\gtrsim}
\mathcommand\inductionorderingless   {<}
\mathcommand\inductionorderinggtr    {>}

\mathcommand\strictlyfounded{\searrow}
\mathcommand\antistrictlyfounded{\swarrow}
\mathcommand\notstrictlyfounded{\not\searrow}
\mathcommand\notantistrictlyfounded{\hskip.3em\setminus\hskip-1.1em\swarrow}
\mathcommand\founded{\curvearrowright}
\mathcommand\antifounded{\curvearrowleft}
\mathcommand\strictquasifoundedindex[1]
{\,\strictlyfounded\mbox{\math/}\!\founded_{#1}\,}

\newcommand\K{{\rm K}}

\mathcommand\Feins{\Gamma}
\mathcommand\Fzwei{\Delta}
\mathcommand\Fdrei{\Pi}  
\mathcommand\Fvier{\Lambda}
\mathcommand\Ffuen{\Theta}
\mathcommand\Fsech{\Xi}
\mathcommand\Feinsprime{\Feins'}
\mathcommand\Fzweiprime{\Fzwei'}
\mathcommand\Fdreiprime{\Fdrei'}
\mathcommand\Feinsprimeprime{\Feins''}

\mathcommand\Feinszwei{\Feins\!\Fzwei}

\newcommand\getformulaname{{\rm logic}}
\mathcommand\getformula[1]{\getformulasofset{\{#1\}}}
\mathapplycommand\getformulasofset{\getformulaname}

\newcommand\hiddenIOsubs[1]{_{#1}}

\mathcommand\iql    [1]{\;\lesssim \hiddenIOsubs{#1}\,}
\mathcommand\iqb    [1]{\;\gtrsim  \hiddenIOsubs{#1}\,}
\mathcommand\il     [1]{\,<        \hiddenIOsubs{#1}\,}
\mathcommand\ileq   [1]{\,\leq     \hiddenIOsubs{#1}\,}
\mathcommand\ib     [1]{\,>\!\!    \hiddenIOsubs{#1}\,}
\mathcommand\notib  [1]{\,\ngtr\!\!\hiddenIOsubs{#1}\,}
\mathcommand\notil  [1]{\,\nless   \hiddenIOsubs{#1}\,}
\mathcommand\iapprox[1]{\,\approx  \hiddenIOsubs{#1}\,}

\setlongbibstyle
\newcommand\paper{paper}
\begin{document}
{\flushbottom
\thispagestyle{empty}
\setcounter{page}{1}

\noindent
\begin{minipage}{\textwidth}
\mbox{}

\vspace{10.0em}
\begin{center}
{\LARGE\bf
Full First-Order Sequent and Tableau Calculi
With Preservation of Solutions \\and
the Liberalized \math\delta-Rule
but\\
Without Skolemization 
\\\mbox{}
}
{
\\\mbox{}
\\\mbox{}
\\
Claus-Peter Wirth
\\\mbox{}
\\
\email
\\
\www
\\\mbox{}
\\\mbox{}
\\
}
{\small
Research Report 698/1998
\\
\url{http://www.ags.uni-sb.de/~cp/p/ftp98/long.html}
\\
\mbox{}\\
December~13, 1998
\\\mbox{}\\
Revised April~1, 1999 
\\
(Typos corrected, 
 Proof of \lemmref{lemma geserthmimproved} and Abstract shortened, 
 References extended)
\\\mbox{}\\
Revised \Sep~29, 1999 
\\
(Typos corrected)
\\
\mbox{}\\\SEKIedition\\
\mbox{}\\
\uniDO
\\
Informatik V
\\
D-44221 Dortmund
\\
Germany
\\
}

\end{center}
\end{minipage}

\vspace{\fill}
{\footnotesize

\noindent
{\bf Abstract:}
\renewcommand{\baselinestretch}{0.8}
We present a combination 
of raising, explicit variable dependency representation, 
the liberalized \math\delta-rule, and
preservation of solutions
for first-order deductive theorem proving.
Our main motivation is to provide the foundation for our work on
inductive theorem proving.

}

}\raggedbottom
\vfill\pagebreak
\setcounter{tocdepth}{2}
\pagestyle{empty}
\setcounter{page}{2}
\samepage{
\tableofcontents\nopagebreak
\vfill\vfill
 
}
\vfill\pagebreak
\pagestyle{myheadings}
\pagenumbering{arabic}
\setcounter{page}{1}

\section{Introduction}
\label{section introduction}

\yestop
\yestop
\noindent
The paper organizes as follows:
After explaining the technical terms of the title in 
\sectref{section introduction} and the remaining basic notions 
in \sectref{sect basic},
we start to explicate the differences between our two versions of 
calculi in \sectref{section two versions}.
The weak version is explained in \sectref{section weak}.
The changes necessary for the strong version 
in order to admit liberalization of the \math\delta-rule
are explained in \sectref{section strong}.
After concluding in \sectref{section conclusion}
we append all the proofs, references, and notes.

\yestop
\yestop
\subsection{Without Skolemization}
\label{section without skolemization}

\yestop
\yestop
\noindent
In this \paper\ we discuss how to analytically
prove first-order theorems in contexts where
Skolemization is not appropriate.
Skolemization has at least three problematic
aspects. 

\begin{enumerate}

\yestop
\item
Skolemization \signatureenlarge s the signature or introduces
higher-order variables.
Unless special care is taken,
this may introduce objects into empty universes
and change the notion of term-generatedness or
Herbrand models.
Above that, 
the Skolem functions occur
in answers to goals
or solutions of constraints\footnotemark\
which in general 
cannot be translated into the original 
signature.
For a detailed discussion of 
these problems \cf\ \cite{miller}.

\yestop
\item
Skolemization results in the following simplified 
quantification structure:
\begin{quote}
For all Skolem functions \math{\vec u} there
are solutions to the \fev s \nolinebreak\math{\vec e} 
(\ie\ the free variables of \cite{fitting}) 
such that the quantifier-free
theorem \math{T(\vec e,\vec u)} is valid.
\\
Short:\LINEmath{\forall\vec u\stopq\exists\vec e\stopq T(\vec e,\vec u).}
\end{quote}

Since the state of a proof attempt
is often represented as the conjunction of the branches of 
a tree (\eg\ in sequent or (dual) tableau calculi),
the \fev s become ``rigid'' or ``global'',
\ie\ a solution for a \fev\ must solve all occurrences
of this variable in the whole proof tree.
This is because,
for \math{B_0,\ldots,B_n} denoting the branches of the proof tree,

\LINEmath{
  \forall\vec u\stopq\exists\vec e\stopq
  \inparenthesesinline{
    B_0
    \und
    \ldots
    \und
    B_n
  }
}

is logically strictly stronger
than
\bigmath{
  \forall\vec u\stopq
  \inparenthesesinline{
    \exists\vec e\stopq B_0
    \nottight\und
    \ldots
    \nottight\und
    \exists\vec e\stopq B_n
  }
.}

\pagebreak

Moreover, with this quantification structure it
does not seem to be possible to do
inductive theorem proving 
by finding, 
for each assumed counterexample,
another counterexample that is strictly smaller in 
some wellfounded ordering.\footnotemark\ 
The reason for this is the following.
When we have some counterexample \math{\vec u}
for \math{T(\vec e,\vec u)}
(\ie\ there is no \nolinebreak\math{\vec e} such that
 \math{T(\vec e,\vec u)}
 is valid)
then for different \nolinebreak\math{\vec e} different branches 
\nolinebreak\math{B_i} in the 
proof tree may cause the invalidity of the conjunction.
If we have applied induction hypotheses in more than one
branch, for different \math{\vec e} we get different smaller
counterexamples. What we would need, however, is one single
smaller counterexample for all \math{\vec e}.

\yestop
\item 
Skolemization increases the size of the formulas.
(Note that in most calculi the only relevant part of Skolem terms is
 the top symbol and the set of occurring variables.)

\yestop
\yestop
\end{enumerate} 

\yestop
\yestop
\noindent
The first and second problematic aspects disappear when
one uses\emph{raising} (\cf\ \cite{miller}) instead of Skolemization.
Raising is a dual of Skolemization and simplifies the
quantification structure to something like: 
\begin{quote}
There are raising functions \math{\vec e} such that for all
possible values of the \fuv s \nolinebreak\math{\vec u }
(\ie\ the nullary constants or ``parameters'')
the quantifier-free
theorem \math{T(\vec e,\vec u)} is valid.
\\
Short:\LINEmath{\exists\vec e\stopq\forall\vec u\stopq T(\vec e,\vec u).}
\end{quote}
Note that due to the two duality switches ``unsatisfiability/validity''
and ``Skolemization/\linebreak raising'', in this \paper\ raising
will look much like Skolemization in refutational theorem proving.
The inverted order of universal and existential quantification of 
raising (compared to Skolemization) is advantageous because now

\noindent
\LINEmath{
  \exists\vec e\stopq\forall\vec u\stopq
  \inparenthesesinline{
    B_0
    \und
    \ldots
    \und
    B_n
  }
}

\noindent
is indeed logically equivalent to
\bigmath{
  \exists\vec e\stopq
  \inparenthesesinline{
    \forall\vec u\stopq B_0
    \nottight\und
    \ldots
    \nottight\und
    \forall\vec u\stopq B_n
  }
.}

Furthermore, inductive theorem proving works well:
When, for some \math{\vec e}, we have some counterexample \math{\vec u}
for \math{T(\vec e,\vec u)}
(\ie\ 
 \math{T(\vec e,\vec u)}
 is invalid)
then one branch \math{B_i} in the 
proof tree must cause the invalidity of the conjunction.
If this branch is closed, then it contains the application of 
an induction hypothesis that is invalid for this \math{\vec e}
and the \math{\vec u'} resulting from the instantiation of
the hypothesis.
Thus, \math{\vec u'} together with the induction hypothesis 
provides the strictly smaller counterexample
we are searching for for this \math{\vec e}.

The third problematic aspect disappears when 
the dependency of variables is explicitly 
represented in a\emph\vc, \cf\ \cite{kohlhasetableaux}.
This idea actually has a long history,
\cf\ \cite{prawitzimproved},
\cite{kanger}, \cite{bibel}.
Moreover, the use of \vc s admits the free existential variables to 
be first-order.

\pagebreak

\yestop
\subsection{Sequent and Tableau Calculi}

\yestop
\noindent
In \cite{smullyan}, rules for analytic theorem proving are
classified as 
\mbox{\math\alpha-,}
\mbox{\math\beta-,}
\mbox{\math\gamma-,} and
\math\delta-rules
independently from a concrete calculus.

\begin{description}

\item[\math\alpha-rules] describe the simple and the 

\item[\math\beta-rules] the case-splitting propositional proof steps.

\item[\math\gamma-rules] show existential properties,
either by exhibiting a term witnessing to the existence
or else by introducing a special kind of variable,
called ``dummy'' in \cite{prawitzimproved} and \cite{kanger},
and ``free variable'' in footnote~11 of \cite{prawitzimproved}
and in \cite{fitting}.
We will call these variables\emph{\fev s}.
By the use of \fev s we can delay the choice of a
witnessing term until the state of the proof attempt
gives us more information which choice is likely to result in 
a successful proof.
It is the important addition of \fev s that makes the major difference
between the free variable calculi of 
\cite{fitting} and the calculi of \cite{smullyan}.
Since there use to be infinitely many possibly witnessing terms
(and different branches may need different ones),
the \math\gamma-rules (under assistance of the \math\beta-rules) 
often destroy the possibility to decide
validity because they enable infinitely many 
\math\gamma-rule applications to the same formula.

\item[\math\delta-rules] show universal properties simply
with the help of a new symbol, called a ``parameter'', about
which nothing is known.
Since the present \fev s must not be instantiated with 
this new parameter, 
in the standard framework of Skolemization and unification 
the parameter is given the present \fev s as arguments.
In this \paper, however, we will use nullary parameters, 
which we call\emph{\fuv s}.
These variables are not {free} in the sense that
they may be chosen freely, but in the sense that they are not bound
by any quantifier. 
Our \fuv s are similar to the parameters of \cite{kanger}
because a \fev\ may not be instantiated with all of them.
We will store the information on the dependency between
\fev s and \fuv s in\emph{\vc s}.
\end{description}

\pagebreak

\subsection{Preservation of Solutions}

Users even of pure Prolog are not so much interested 
in theorem proving as they are in answer computation.
The theorem they want to prove usually contains some
free existential variables that are instantiated during 
a proof attempt. When the proof attempt is successful,
not only the input theorem is known to be valid but also
the instance of the theorem with the substitution built-up
during the proof. 
Since the knowledge of mere existence is much less useful
than the knowledge of a term that witnesses to this existence
(unless this term is a only free existential variable),
theorem proving should---if possible---always provide
these witnessing terms.
Answer computation is no problem
in Prolog's Horn logic because it is so simple.
But also for the more difficult clausal logic, answer computation is possible.
\Cf\ \eg\ \cite{answercomputation}, where tableau calculi are used for
answer computation in clausal logic.
Answer computation becomes even harder when we
consider full first-order logic instead of clausal logic.
When \math\delta-steps occur in a proof,
the introduced free universal variables
may provide no information on what kind of 
object they denote.
Their excuse may be that they cannot do this
in terms of computability or \math\lambda-terms.
Nevertheless, they can provide this information in 
form of \hilbert's \math\varepsilon-terms, and 
the strong versions of our calculi will do so.
When full first-order logic is considered,
one should focus on\emph{preservation of solutions} 
instead of computing answers.
By this we mean at least the following property:
\begin{quote}
All solutions that transform a proof attempt
for a proposition into a closed proof 
(\ie\ the closing substitutions for the \fev s)
are also solutions of the original proposition.
\end{quote}
This again is closely related to inductive theorem proving:
Suppose that we finally have shown that for the reduced form 
\math{R(\vec e,\vec u)} (\ie\ the state of the proof attempt)
of the original theorem \math{T(\vec e,\vec u)}
(\cf\ the discussion in \sectref{section without skolemization}),
there is some solution \math{\vec e} such that
for each counterexample \math{\vec u} of 
\math{R(\vec e,\vec u)} there is a counterexample
\math{\vec u'} for the original theorem and that this \math{\vec u'}
is strictly smaller than \math{\vec u} in some wellfounded ordering.
In this case we have proved \math{T(\vec e,\vec u)} only if the 
solution \math{\vec e} for the reduced form 
\math{\forall\vec u\stopq R(\vec e,\vec u)} 
is also a solution for the original theorem 
\math{\forall\vec u\stopq T(\vec e,\vec u)}. 

\pagebreak

\subsection{The Liberalized \math\delta-rule}

\yestop
\yestop
\noindent
We use `\math\uplus' for the union of disjoint classes and `\id' for the
identity function.
For a class \nolinebreak\math R we define\emph{domain},\emph{range}, 
and\emph{restriction to} 
and\emph{image}\footnotemark\  
and\emph{reverse-image of a class \math A} by

\noindent
\LINEmath{
 \begin{array}[t]{l l l l}
   \DOM R
  &:=
  &\setwith{\!a}{\exists b\stopq          (a,b)\tightin R\!}
  &;
  \\\RAN R
  &:=
  &\setwith{\!b}{\exists a\stopq          (a,b)\tightin R\!}
  &;
  \\\domres R A
  &:=
  &\setwith{\!(a,b)\tightin R}{a\tightin A\!}
  &;
  \\\relapp R A
  &:=
  &\setwith{\!b}{\exists a\tightin A\stopq(a,b)\tightin R\!}
  &;
  \\\revrelapp R B
  &:=
  &\setwith{\!a}{\exists b\tightin B\stopq(a,b)\tightin R\!}
  &.
  \\
 \end{array}
}

\yestop
\yestop
\noindent
We define a\emph{sequent} to be a list of formulas.\footnotemark\
The\emph{conjugate} of a formula\ \nolinebreak\math A 
(written:\ \nolinebreak\ \nolinebreak\overline A\ \nolinebreak)
is 
the formula\ \nolinebreak\math B 
if\ \nolinebreak\math A is of the form\ \nolinebreak\math{\neg B},
and the formula\ \nolinebreak\math{\neg A} otherwise.
Note that the conjugate of the conjugate of a formula 
is the original formula again, unless it has the 
form \nolinebreak\math{\neg\neg B}.

In the tradition of \cite{gentzen} we assume the symbols for\emph{\fev s}
(\ie\ the free variables of \cite{fitting}),\emph{\fuv s}
(\ie\ nullary parameters),\emph{bound variables}
(\ie\ variables for quantified use only), 
and the\emph{constants}
(\ie\ the function (and predicate) symbols from the signature)
to come from four disjoint sets \Vsome, \Vall, \Vbound, and \math\Sigmaoffont.
We assume each of \Vsome, \Vall, \Vbound\
to be infinite (for each sort)
and set \math{\Vfree:=\Vsome\tightuplus\Vall}.
Moreover, 
due to the possibility to rename bound variables \wrog,
we do not permit quantification on variables that occur already
bound in a formula;\label{section quantification restriction} \ie\ \eg\
\math{\forall x\tight:\,A} is only a formula in our sense 
if \math A does not contain a quantifier on \math x like
\math{\forall x} or \math{\exists x}.
The simple effect is that our \math\gamma- and \math\delta-rules
can simply replace\emph{all} occurrences of \math x.
For a term, formula, 
sequent \Feins\ 
\etc,
`\VARsome\Feins', \nolinebreak`\VARall\Feins', \nolinebreak`\VARbound\Feins',
`\VARfree\Feins' 
denote the sets of variables from
\Vsome, \nolinebreak\Vall, \nolinebreak\Vbound, \Vfree\ 
occurring in \nolinebreak\Feins, \resp.
For a substitution
\math\sigma\ 
we denote with `\math{\Feins\sigma}' 
the result of replacing in \nolinebreak\Feins\ 
each 
variable \nolinebreak\math x 
in \DOM\sigma\ with \math{\sigma(x)}.
Unless stated otherwise,
we tacitly assume that each 
substitution \math\sigma\
satisfies 
\ \math{
  \VARbound{\DOM\sigma\cup\RAN\sigma}
}
\math=
\nolinebreak\math{
  \emptyset
,} \ 
such that no bound variables can be replaced 
and no additional variables become bound (\ie\ captured) 
when applying \nolinebreak\math\sigma.

\yestop
\yestop
\yestop
\yestop
\noindent
A\emph\vc\ \math R is a subset of \bigmath{\Vsome\times\Vall.}
Roughly speaking, 
\bigmath{
 (\existsvari x{},\forallvari y{})
 \tightin
 R
}
says that \existsvari x{} is older than \forallvari y{},
so that we must not instantiate the
\fev\ \existsvari x{} with a term containing \forallvari y{}.

While the benefit of the introduction of 
\fev s in \math\gamma-rules is to delay the choice 
of a witnessing term,
it is sometimes unsound to instantiate such a
\fev\ \existsvari x{} with a term containing a \fuv\ 
\forallvari y{}
that was introduced later than \existsvari x{}:

\pagebreak

\yestop
\begin{example}
\\\LINEmath{
  \exists x\stopq
  \forall y\stopq
  (x\tightequal y)
}

\noindent
is not deductively valid.
We can start a proof attempt via:
\\\math\gamma-step:
\\\LINEmath{
  \forall y\stopq
  (\existsvari x{}\tightequal y)
.}\\
\math\delta-step:
\\\LINEmath{
  (\existsvari x{}\tightequal\forallvari y{})
.}

\noindent
Now, if we were allowed to substitute the \fev\ \existsvari x{} 
with the \fuv\ \forallvari y{}, we would get the tautology
\bigmath{
  \inparenthesesinlinetight{
    \forallvari y{}\tightequal\forallvari y{}
  }
,}
\ie\ we would have proved an invalid formula.
In order to prevent this, the \math\delta-step has
to record \math{(\existsvari x{},\forallvari y{})}
in the \vc, which disallows the instantiation step.
\end{example}

\yestop
\yestop
\yestop
\noindent
In order to restrict the possible instantiations as 
little as possible, we should keep our \vc s as small
as possible.
\cite{kanger} and \cite{bibel} are quite generous in that they
let their \vc s become quite big:

\yestop
\begin{example}\label{example liberalized}\sloppy
\\\LINEmath{
  \exists x\stopq
  \inparentheses{
      \Pppp x
    \nottight\oder
      \forall y \stopq
      \neg\Pppp y
  }
}
\\
can be proved the following way:
\\
\math\gamma-step:
\\\LINEmath{
  \inparentheses{
      \Pppp{\existsvari x{}}
    \nottight\oder
      \forall y \stopq
      \neg\Pppp y
  }
.}
\\\math\alpha-step:
\\\LINEmath{
      \Pppp{\existsvari x{}}
  \comma\ 
      \forall y \stopq
      \neg\Pppp y
.}
\\\math\delta-step:
\\\LINEmath{
      \Pppp{\existsvari x{}}
  \comma
      \neg\Pppp{\forallvari y{}}
.}
\\
Instantiation step:
\\\LINEmath{
      \Pppp{\forallvari y{}}
  \comma
      \neg\Pppp{\forallvari y{}}
.}

\noindent
The last step is not allowed in the above citations,
so that another \math\gamma-step must be applied to
the original formula in order to prove it.
Our instantiation step, however, is perfectly sound:
Since \existsvari x{} does not occur in 
\bigmath{
      \forall y \stopq
      \neg\Pppp y
,}
the free variables
\existsvari x{} and \forallvari y{}
do not depend on each other 
and 
there is no reason to insist on 
\existsvari x{} \nolinebreak being older than \nolinebreak\forallvari y{}.
Note that moving-in the existential quantifier transforms the
original formula into the logically equivalent formula
\bigmath{
      \exists x\stopq
      \Pppp x
    \nottight\oder
      \forall y \stopq
      \neg\Pppp y
,}
which (after a preceding \math\alpha-step)
enables the 
\math\delta-step introducing \forallvari y{}
to come before the 
\math\gamma-step introducing \existsvari x{}.
\end{example}

\yestop
\yestop
\yestop
\noindent
Keeping small the \vc s generated by the \math\delta-rule
results in non-elementary reduction of the size of smallest
proofs. This ``liberalization of the \math\delta-rule''
has a history ranging from \cite{smullyan} over
\cite{deltaplus} to \cite{baazdelta}.
While the liberalized \math\delta-rule of \cite{smullyan}
is already able to prove the formula of \examref{example liberalized}
with a single \math\gamma-step,
it is much more restrictive than the more liberalized
\math\delta-rule of \cite{baazdelta}.

\yestop
\noindent
Note that liberalization of the \math\delta-rule is not
simple because it easily results in unsound calculi,
\cf\ \cite{kohlhasetableaux} \wrt\ our \examref{ex kohlhase fail}
and \cite{kohlhasetableauxupdated} \wrt\ our \examref{example R'' so big}.
The difficulty lies with instantiation steps that 
relate previously unrelated variables:

\yestop
\begin{example}\label{ex kohlhase fail}\sloppy

\noindent
\LINEmath{
      \exists x\stopq
      \forall y\stopq
      \Qppp x y
    \nottight\oder
      \exists u\stopq
      \forall v\stopq
      \neg\Qppp v u
}

\noindent
is not deductively valid (to wit, let \Qpsymbol\ be the
identity relation on a non-trivial universe).

\yestop
\noindent
Consider the following proof attempt:
One \math\alpha-, two \math\gamma-, and two liberalized 
\math\delta-steps result in 

\noindent
\LINEmath{
  \Qppp{\existsvari x{}}{\forallvari y{}}
  \comma
  \neg\Qppp{\forallvari v{}}{\existsvari u{}}
}(\math\ast)

\noindent
with \vc\

\noindent
\LINEmath{
  R
  \nottight{\nottight{:=}}
  \{
    \pair
      {\existsvari x{}}
      {\forallvari y{}}
    \comma
    \pair
      {\existsvari u{}}
      {\forallvari v{}}
  \}
.}(\#)

\noindent
(Note that the non-liberalized \math\delta-rule
would additionally have produced
\pair{\existsvari x{}}{\forallvari v{}}
or
\pair{\existsvari u{}}{\forallvari y{}}
or both,
depending on the order of the proof steps.)

When we now instantiate \existsvari x{} with \forallvari v{},
we relate the previously unrelated variables \existsvari u{}
and \forallvari y{}. 
Thus, our new goal 

\noindent
\LINEmath{
  \Qppp{\forallvari v{}}{\forallvari y{}}
  \comma
  \neg\Qppp{\forallvari v{}}{\existsvari u{}}
}

\noindent
must be equipped with the new \vc\
\math{
  \{
    \pair
      {\existsvari u{}}
      {\forallvari y{}}
  \}
.}
Otherwise we could instantiate \existsvari u{}
with \forallvari y{}, resulting in the tautology
\bigmath{
  \Qppp{\forallvari v{}}{\forallvari y{}}
  \comma
  \neg\Qppp{\forallvari v{}}{\forallvari y{}}
.}

Note that in the standard framework of Skolemization and unification, 
this new \vc\ is automatically generated by the occur-check of unification:
When we instantiate 
\existsvari x{} with 
\math{
  \forallvari v{}\funarg{\existsvari u{}}
}
in

\noindent
\LINEmath{
  \Qppp
    {\existsvari x{}}
    {\forallvari y{}\funarg{\existsvari x{}}}
  \comma
  \neg
  \Qppp
    {\forallvari v{}\funarg{\existsvari u{}}}
    {\existsvari u{}}
}

\noindent
we get

\noindent
\LINEmath{
  \Qppp
    {\forallvari v{}\funarg{\existsvari u{}}}
    {\forallvari y{}\funarg{\forallvari v{}\funarg{\existsvari u{}}}}
  \comma
  \neg
  \Qppp
    {\forallvari v{}\funarg{\existsvari u{}}}
    {\existsvari u{}}
,}

\noindent
which cannot be reduced to a tautology because
\math{
  \forallvari y{}\funarg{\forallvari v{}\funarg{\existsvari u{}}}
}
and 
\existsvari u{}
cannot be unified.

When we instantiate the variables 
\existsvari x{} and \existsvari u{} 
in the sequence (\math\ast)
in parallel
via 

\noindent
\LINEmath{
  \sigma
  :=
  \{
    \existsvari x{}\tight\mapsto\forallvari v{}
  ,\ 
    \existsvari u{}\tight\mapsto\forallvari y{}
  \}
,}(\$)

\noindent
we have to 
check whether the newly imposed \vc s are consistent with the
substitution itself. 
In particular, a cycle
as given (for the \math R of (\#)) by

\noindent
\LINEmath{
  \forallvari y{}
  \nottight{\nottight{\reverserelation\sigma}}
  \existsvari u{}
  \nottight{\nottight{R}}
  \forallvari v{}
  \nottight{\nottight{\reverserelation\sigma}}
  \existsvari x{}
  \nottight{\nottight{R}}
  \forallvari y{}
}

\noindent
must not exist.
Although this sounds fairly difficult,
the formal treatment is quite simple.
\end{example}

\pagebreak

\section{Basic Notions, Notations, and Assumptions}
\label{sect basic}

\noindent
We make use of ``\opt\ldots'' for stating two
definitions, lemmas, or theorems (and their proofs \etc)\ in one, 
where the parts between `\opt{' and `}' are
optional and are meant to be all included or all omitted.
`\N' denotes the set of and `\math\prec' the ordering on natural numbers.
We define
\bigmath{\posN:=\setwith{n\tightin\N}{0\tightnotequal n}.}

Let `\math R' 
denote a binary relation.
\math R is said to be a\emph{relation on \math A}
\udiff\ \bigmath{\DOM R\cup\RAN R\nottight{\nottight\subseteq} A.}
\math R is\emph{irreflexive} \udiff\  
\bigmath{
  \id\cap R=\emptyset
.} 
It is \math A{\em-reflexive\/} \udiff\  
\bigmath{
  \domres\id A\subseteq R
.} 
Simply speaking of a\emph{reflexive} relation
we refer to the biggest \math A that is appropriate in the local context,
and referring to this \math A
we write \math{R^0} 
to ambiguously denote 
\math{\domres\id A}.
Furthermore, we write \math{R^1} 
to denote
\math R%
. 
For \math{n\in\posN} we 
write \math{R^{n+1}} 
to denote
\bigmath{R^{n}\tight\circ R
,} 
such that \math{R^{n}} 
denotes the \math n step relation
for \math R%
.
%
The\emph{transitive closure} of 
\math R
is 
\bigmath{\transclosureinline R:=\bigcup_{n\in\posN}R^n.}
The\emph{reflexive \& transitive closure} of 
\math R
is 
\bigmath{
  \refltransclosureinline R
  :=
  \bigcup_{n\in\N}R^n
.}
%
The\emph{reverse}\footnotemark\ of 
\math R
will be denoted with 
\reverserelation R%
.
\ \math R \nolinebreak is\emph{terminating} \udiff\ there is no
\FUNDEF s\N{\DOM R} with \bigmath{s_i\nottight R s_{i+1}}
for all \math{i\in\N}.

Furthermore, we use `\math\emptyset' to denote the empty set as well as the
empty function or empty word.
By an (irreflexive)\emph{ordering} `\math <' (on \math A)
we mean an irreflexive and transitive binary relation (on \math A),
sometimes called ``strict partial ordering'' \etc\ by other authors. 
A\nolinebreak\emph{reflexive ordering `\/\math\leq' on} \math A 
is an \math A-reflexive, 
antisymmetric, and transitive relation on \nolinebreak\math A.
The\emph{reflexive ordering on \math A of an ordering}~\math < 
is \math{(\tight<\cup\id)\cap(A\tighttimes A)}.
%
An ordering \nolinebreak\math <  is called\emph{wellfounded} \udiff\
\math > is terminating;
where, as with all our asymmetric relation symbols,  
\bigmath{\tight>\nottight{\nottight{:=}}\reverserelation<.}
%
The\emph{class of total functions from \math A to \math B}
is denoted with \FUNSET A B.
The\emph{class of (possibly) partial functions from \math A to \math B}
is denoted with \PARFUNSET A B.

Validity is expected to be given 
with respect to some \semanticobject\ (\math\Sigmaoffont-algebra)
\nolinebreak\algebra,
assigning a universe (to each sort)
and an appropriate function to each symbol in \math\Sigmaoffont.
For \math{\X\subseteq\Vfree} we denote 
the set of total \algebra-valuations of \X\
(\ie\ functions mapping free variables to 
 objects of the universe of \algebra\
 (respecting sorts))
with \bigmath{\FUNSET\X\algebra} and 
the set of (possibly) partial \algebra-valuations of \X\
with \bigmath{\PARFUNSET\X\algebra.}
For \math{\pi\in\FUNSET\X\algebra} we denote with
`\math{\algebra\tightuplus\pi}'
the extension of \algebra\ to the variables of \X\
which are then treated as nullary constants.
More precisely, we assume the existence of some
evaluation function `\EVALSYM' such that \EVAL{\algebra\tightuplus\pi}
maps any term
over \math{\Sigmaoffont\tightuplus\X} into the universe
of \algebra\ (respecting sorts) 
such that for all \math{x\in\X}:
\bigmath{
  \EVAL{\algebra\tightuplus\pi}
  \funarg x
  \tightequal
  \pi
  \funarg x
.}
Moreover, \EVAL{\algebra\tightuplus\pi} maps
any formula \math B over
\math{\Sigmaoffont\tightuplus\X} to \TRUEpp\ or \FALSEpp,
such that \math B is valid in \math{\algebra\tightuplus\pi} 
\uiff\ \math{\EVAL{\algebra\tightuplus\pi}(B)=\TRUEpp}.
We assume that the\emph{Substitution-Lemma} holds in the sense
  that, for any
  substitution \math\sigma,
  \semanticobject\ \algebra, and valuation
  \math{\pi\in\FUNSET\Vfree\algebra},
  validity of a formula 
  \math B in
  \math{
    \algebra
    \uplus
    \inparenthesesinlinetight{
       (\sigma
        \uplus
        \domres\id{\Vfree\setminus\DOM\sigma})
       \circ
       \EVAL{\algebra\tightuplus\pi}
    }
  }
  is logically equivalent to validity of 
  \math{B\sigma} in \math{\algebra\tightuplus\pi}.
Finally,
we assume that the value of the evaluation function on a term 
or formula \math B
  does not depend on the free variables that do not occur in \math B:
  \ \math{
    \EVAL{\algebra\tightuplus\pi}(B)
    =
    \EVAL{\algebra\uplus\domres\pi{\VARfree B}}(B)
 .}
Further properties of validity or evaluation are definitely not needed.

\vfill

\pagebreak

\section{Two Versions of \VC s}
\label{section two versions}

In this section we formally describe two possible choices for 
the formal treatment of \vc s.
The\emph{weak} 
version works well with the non-liberalized \math\delta-rule. 
The\emph{strong} version is a little more difficult
but can be used for the liberalized versions of the \math\delta-rule. 
The presented material is rather formal, but this cannot be avoided and the 
following sections will be less difficult then.

Several binary relations on free variables will be introduced.
The overall idea is that when \pair x y occurs in such a relation
this means something like ``\math x is older than \nolinebreak\math y''
or ``the value of \math y
depends on or is described in terms of \math x''.

\yestop
\begin{definition}
[\math{E_\sigma}, \math{U_\sigma}]
\label{definition e u ex}

\noindent
For a substitution \math\sigma\ with 
\bigmath{
  \DOM\sigma
  \tightequal
  \Vsome
} 
we define the\emph{existential relation} to be

\noindent
\LINEmath{
  E_\sigma
  \nottight{\nottight{:=}}
  \setwith
    {\pair{x'}x}
    {x'\tightin\VARsome{\sigma\funarg{x}}
     \und
     x\tightin\Vsome
    }
}

\noindent
and the\emph{universal relation} to be

\noindent
\LINEmath{
  U_\sigma
  \nottight{\nottight{:=}}
  \setwith
    {\pair y x}
    {y\tightin\VARall{\sigma\funarg{x}}
     \und
     x\tightin\Vsome
    }
.}
\end{definition}

\yestop
\begin{definition}[\opt{Strong} Existential \math R-Substitution]
\label{definition ex r sub}
\\
Let \math R be a \vc. 
\\
\bigmath\sigma\ is an\emph{\exRsub}
\udiff\  
\math\sigma\ is a substitution with
\bigmath{
  \DOM\sigma
  \tightequal
  \Vsome
} 
for which
\bigmath{
 U_\sigma\circ R
}
is irreflexive.
\\
\bigmath\sigma\ is a\emph{strong existential \math R-substitution}
\udiff\  
\math\sigma\ is a substitution with
\bigmath{
  \DOM\sigma
  \tightequal
  \Vsome
} 
for which
\bigmath{
 \transclosureinline{\inparenthesesinlinetight{U_\sigma\circ R}}
}
is a wellfounded ordering.
\end{definition}

\yestop
\noindent
Note that, regarding syntax, 
\bigmath{\pair{\existsvari x{}}{\forallvari y{}}\tightin R} 
is intended to mean that an \exRsub\ \nolinebreak\math\sigma\ 
may not replace \math{\existsvari x{}} with a term in which 
\math{\forallvari y{}} occurs, 
\ie\ \bigmath{\pair{\forallvari y{}}{\existsvari x{}}\tightin U_\sigma} 
must be disallowed,
\ie\ \bigmath{U_\sigma\tight\circ R} must be irreflexive.
Thus, 
the definition of a (weak) \exRsub\
is quite straightforward.
The definition of a\emph{strong} \exRsub\
requires an additional transitive closure because the strong version then
admits a smaller \math R. 
To see this, take from \examref{ex kohlhase fail} the
\vc\ \math R of (\#) and the \math\sigma\ of \nolinebreak(\$). 
As explained there, \math\sigma\ must not be a
strong existential \math R-substitution
due to the cycle
\bigmath{
  \forallvari y{}
  \nottight{U_\sigma}
  \existsvari u{}
  \nottight{R}
  \forallvari v{}
  \nottight{U_\sigma}
  \existsvari x{}
  \nottight{R}
  \forallvari y{}
}
which just contradicts the irreflexivity of 
\bigmath{
  \inparenthesesinlinetight{
    U_\sigma
    \tight\circ
    R
  }^2
.}
Note that in practice \wrog\ \math{U_\sigma} and \math R can always be chosen 
to be finite, so that irreflexivity of 
\transclosureinline{\inparenthesesinlinetight{U_\sigma\tight\circ R}}
is then equivalent to 
\transclosureinline{\inparenthesesinlinetight{U_\sigma\tight\circ R}}
being a wellfounded ordering.

\pagebreak

\yestop
\noindent
After application of 
a \opt{strong} \exRsub\ \math\sigma,
in case of 
\bigmath{
  \pair{\existsvari x{}}{\forallvari y{}}\tightin R
,}
we have to ensure that 
\math{\existsvari x{}} is not replaced with \nolinebreak\math{\forallvari y{}}
via a future application of another \opt{strong} \exRsub\ that replaces
a \fev\ \math{\existsvari u{}} occurring in \math{\sigma(\existsvari x{})} 
with \nolinebreak\math{\forallvari y{}}.
In this case, the new \vc\ has to contain 
\pair{\existsvari u{}}{\forallvari y{}}.
This means that
\math{
  E_\sigma
  \tight\circ
  R
}
must be a subset of the updated \vc. For the weak version this
is already enough. 
For the strong version we have to
add an arbitrary number of steps with
\math{U_\sigma\tight\circ R} again.

\begin{definition}[\opt{Strong} \math\sigma-Update]\label{definition update}

\noindent
Let \math R be a \vc\ and \math\sigma\ be an \opt{strong} \exRsub.

\noindent
The \opt{\tightemph{strong}}\emph{\math\sigma-update of \math R} is 
\bigmath{
  E_\sigma
  \tight\circ
  R
  \ 
  \opt
    {\nottight\circ
     \refltransclosureinline
       {\inparenthesesinlinetight
          {U_\sigma\tight\circ R}}}
.}
\end{definition}

\yestop
\begin{example}

\noindent
In the proof attempt of \examref{ex kohlhase fail}
we applied the strong \exRsub\
\\\linemath{
  \sigma'
  :=
  \{{\existsvari x{}}\tight\mapsto{\forallvari v{}}\}
  \uplus
  \domres\id{\Vsome\setminus\{\existsvari x{}\}}
}
where
\bigmath{
  R
  \tightequal
  \{
    \pair{\existsvari x{}}{\forallvari y{}}
  ,\ 
    \pair{\existsvari u{}}{\forallvari v{}}
  \}
.}
Note that
\\\linemath{
  U_{\sigma'}
  \tightequal
  \{\pair{\forallvari v{}}{\existsvari x{}}\}
}
and
\\\LINEmath{
  E_{\sigma'}
  \tightequal
  \domres\id{\Vsome\setminus\{\existsvari x{}\}}
.}
\\
Thus:
\\\LINEmath{
  \begin{array}[t]{l l l}
       E_{\sigma'}\tight\circ R
       \circ
       \inparenthesesinlinetight{
       U_{\sigma'}\tight\circ R
       }
       ^
       0
      &=
      &\{\pair{\existsvari u{}}{\forallvari v{}}\}
    \\
       E_{\sigma'}\tight\circ R
       \circ
       \inparenthesesinlinetight{
       U_{\sigma'}\tight\circ R
       }
       ^
       1
      &=
      &\{\pair{\existsvari u{}}{\forallvari y{}}\}
    \\
       E_{\sigma'}\tight\circ R
       \circ
       \inparenthesesinlinetight{
       U_{\sigma'}\tight\circ R
       }
       ^
       2
      &=
      &\emptyset
    \\
  \end{array}
}

\noindent
The strong
\math{\sigma'}-update of \math R is then the new \vc\
\\\LINEmath{
  \{
    \pair{\existsvari u{}}{\forallvari v{}}
  \comma
    \pair{\existsvari u{}}{\forallvari y{}}
  \}
.}
\end{example}

\pagebreak

\yestop
\yestop
\noindent
Let \algebra\ be some \semanticobject.
We now define a semantic counterpart of our existential
\math R-substitutions, which we will call 
``existential \math{(\algebra,R)}-valuation''.
Suppose that 
\ \math e\nolinebreak\ \nolinebreak\ 
maps each \fev\ not directly to an object of \algebra\ 
(of the same sort),
but can additionally read the values of some \fuv s under
an \algebra-valuation \math{\pi\in\FUNSET\Vall\algebra},
\ie\ \ \math e\nolinebreak\ \nolinebreak\ 
gets some \math{\pi'\in\PARFUNSET\Vall\algebra}
with \bigmath{\pi'\tightsubseteq\pi} as a second argument;
short:
\bigmath{
  \FUNDEF 
    e
    \Vsome
    {\inparenthesesinlinetight
      {\FUNSET
        {\inparenthesesinlinetight
          {\PARFUNSET\Vall\algebra}
        }
        \algebra
      }
    }
.}
Moreover, for each \fev\ \math x, we require 
the set of read \fuv s (\ie\ \DOM{\pi'}) to be 
identical for all \bigmath{\pi;}
\ie\
there has to be some ``semantic relation''
\math{S_e\subseteq\Vall\tighttimes\Vsome}
such that
for all \math{x\in\Vsome}: 

\noindent
\LINEmath{
  \FUNDEF
    {e\funarg x}
    {\inparenthesesinlinetight
       {\FUNSET
          {\revrelappsin{S_e}x}
          \algebra}}
    \algebra     
.}

\noindent
Note that, for each \math e, at most one semantic relation
exists, namely

\noindent
\LINEmath{
  S_e
  \nottight{\nottight{:=}}
  \setwith
    {\pair y x}
    {y\tightin\DOM
       {\bigcup\funarg
         {\DOM{e\funarg x}}}
     \und
     x\tightin\Vsome}
.}

\yestop
\begin{definition}
[\math{S_e}, \opt{Strong} Existential \pair\algebra R-Valuation, 
 \math\epsilon]
\label{definition exval}

\noindent
Let \math R be a \vc,
\algebra\ a \semanticobject,
and
\bigmath{
  \FUNDEF 
    e
    \Vsome
    {\inparenthesesinlinetight
      {\FUNSET
        {\inparenthesesinlinetight
          {\PARFUNSET\Vall\algebra}
        }
        \algebra
      }
    }
.}

\noindent 
The\emph{semantic relation of \math e} is
\bigmath{
  S_e
  \nottight{\nottight{:=}}
  \setwith
    {\pair y x}
    {y\tightin\DOM
       {\bigcup\funarg
         {\DOM{e\funarg x}}}
     \und
     x\tightin\Vsome}
.}

\noindent
\math e \ is an\emph{existential \pair\algebra R-valuation}
\udiff\
\bigmath{
   S_e\circ R
}
is irreflexive and, for all \math{x\in\Vsome}, 
\\\LINEmath{
  \FUNDEF
    {e\funarg x}
    {\inparenthesesinlinetight
       {\FUNSET
          {\revrelappsin{S_e}x}
          \algebra}}
    \algebra     
.}

\noindent
\math e \ is a\emph{strong existential \pair\algebra R-valuation}
\udiff\
\bigmath{
   \transclosureinline{\inparenthesesinlinetight{S_e\circ R}}
}
is a wellfounded ordering and, for all \math{x\in\Vsome}, 
\\\LINEmath{
  \FUNDEF
    {e\funarg x}
    {\inparenthesesinlinetight
       {\FUNSET
          {\revrelappsin{S_e}x}
          \algebra}}
    \algebra     
.}

\noindent
Finally, for applying \opt{strong} existential \pair\algebra R-valuations in a
uniform manner, we define the function
\\\linemath{\footroom
  \FUNDEF
    \epsilon
    {~~(\FUNSET\Vsome{(\FUNSET{(\PARFUNSET\Vall\algebra)}\algebra)})~~}
    {~~(\FUNSET{(\FUNSET\Vall\algebra)}{(\FUNSET\Vsome\algebra)})~~}
}
by 
(~\math{e\in\FUNSET\Vsome{(\FUNSET{(\PARFUNSET\Vall\algebra)}\algebra)}},
 ~~\math{\pi\in\FUNSET\Vall\algebra},
 ~~\math{x\in\Vsome}~)
\\\LINEmath{\headroom
  \epsilon(e)(\pi)(x)
  :=e(x)(\domres\pi{\revrelappsin{S_e}x})
.}

\end{definition}

\yestop
\yestop
\begin{lemma}\label{lemma changing R}
Let \math R be a \vc.
\begin{enumerate}

\noitem
\item
Let \math{R'} be a \vc\ with \bigmath{R\tightsubseteq R'.}\\
For each \opt{strong} existential \pair\algebra{R'}-valuation 
\bigmath{e'} there is some 
\\
\opt{strong} existential \pair\algebra R-valuation
\bigmath e 
such that 
\bigmath{\epsilon\funarg e\tightequal\epsilon\funarg{e'}.}

\item
Let \math\sigma\ be a \opt{strong} \exRsub\ 
and \math{R'} the \opt{strong} 
\math\sigma-update of \math R.
\\
For each \opt{strong} existential \pair\algebra{R'}-valuation 
\bigmath{e'} there is some 
\\
\opt{strong} existential \pair\algebra R-valuation
\bigmath e 
such that 
for all \math{\pi\in\FUNSET\Vall\algebra}:

\noindent
\LINEmath{
  \epsilon
  \funarg e
  \funarg\pi
  \nottight{\nottight{\nottight{\nottight{=}}}}
  \sigma
  \nottight{\nottight{\circ}}
  \EVAL{\algebra\uplus\epsilon\funarg{e'}\funarg\pi\uplus\pi}  
.}

\end{enumerate}
\end{lemma}
\vfill\pagebreak
\section{The Weak Version}
\label{section weak}

\yestop
\yestop
\noindent
We are now going to define \math R-validity of a set of sequents with 
free variables,
in terms of 
validity of a formula 
(where the free variables are treated as nullary constants).

\yestop
\begin{definition}[Validity]\label{definition weak validity} 

\noindent
Let \math R be a \vc,
\algebra\ a \semanticobject, and \math G a set of sequents.

\noindent
\goals\ is\emph{\math R-valid in \nolinebreak\algebra} 
\udiff\
there is an
existential \math{(\algebra,R)}-valuation \bigmath e such that
\goals\ is \pair e\algebra-valid.

\noindent
\goals\ 
is\emph{\pair e\algebra-valid} \udiff\
\goals\ is \trip\pi e\algebra-valid
for all \math{\pi\in\FUNSET\Vall\algebra}.

\noindent
\goals\ 
is\emph{\trip\pi e\algebra-valid} \udiff\ 
\ \math G is valid in
  \math{
     \algebra
     \nottight\uplus
     \epsilon
     \funarg e
     \funarg\pi
     \nottight\uplus
     \pi
  }.

\noindent
\math G is\emph{valid in \algebra} 
\udiff\ \Feins\ is valid in \algebra\
for all
\math{
  \Feins\in\goals
}.

\noindent
A sequent \Feins\ is\emph{valid in \algebra} 
\udiff\
there is some formula listed in \nolinebreak\Feins\ that is valid in 
\nolinebreak\algebra.

\noindent
Validity in a class of \semanticobject s
is understood as validity in each of the \semanticobject s of that class.

\noindent
If we omit the reference to a special \semanticobject\
we mean validity (or reduction, \cf\ below) 
in some fixed class \nolinebreak\K\
of \semanticobject s, \eg\ the class of all \math\Sigmaoffont-structures 
(\math\Sigmaoffont-algebras)
or the class of  Herbrand \math\Sigmaoffont-structures
(term-generated \math\Sigmaoffont-algebras), \cf\ \cite{wgcade} for more
interesting classes for establishing inductive validities.
\yestop
\end{definition}

\yestop
\begin{lemma}[Anti-Monotonicity of Validity in \math R]
\label{lemma reduces to 5a}\\
Let \math G be a set of sequents
and \math R and \math{R'} \vc s with \bigmath{R\tightsubseteq R'.}
Now:\\ 
If   \math{G} is \math{R'}-valid in \nolinebreak\algebra,
then \math{G} is \math{R }-valid in \nolinebreak\algebra.
\end{lemma}

\yestop
\begin{example}[Validity]\label{ex validity}
\sloppy

\noindent
For 
\math{\existsvari x{}\in\Vsome}, 
\math{\forallvari y{}\in\Vall},
the sequent 
\math{
  \existsvari x{}\boldequal\forallvari y{}
} 
is \math\emptyset-valid in any \algebra\
because we can choose 
\math{S_e:=\Vall\tighttimes\Vsome}
and
\math{
  e
  \funarg{\existsvari x{}}
  \funarg\pi
  :=
  \pi
  \funarg{\forallvari y{}}
}
resulting in 
\bigmath{
   \epsilon
   \funarg e 
   \funarg\pi
   \funarg{\existsvari x{}}
   =
   e
   \funarg{\existsvari x{}}
   \funarg{\domres\pi{\revrelappsin{S_e}{\existsvari x{}}}}
   =
   e
   \funarg{\existsvari x{}}
   \funarg{\domres\pi\Vall}
   =
   \pi
   \funarg{\forallvari y{}}
.}
This means that \math\emptyset-validity of 
\math{
  \existsvari x{}\boldequal \forallvari y{}
} 
is the same as validity of 
\bigmath{
  \forall y
  \stopq
  \exists x
  \stopq
  x\boldequal y
.}
Moreover, note that
\math{
  \epsilon(e)(\pi)
}
has access to the \math\pi-value of 
\math{\forallvari y{}}
just as a raising function \math f for \nolinebreak\math x
in the raised (\ie\ dually Skolemized) version 
\math{ 
  f(\forallvari y{})
  \boldequal 
  \forallvari y{}
}
of
\bigmath{
  \forall y
  \stopq
  \exists x
  \stopq
  x\boldequal y
.}

Contrary to this, for \math{R:=\Vsome\tighttimes\Vall},
the same formula
\math{
  \existsvari x{}
  \boldequal 
  \forallvari y{}
} 
is not
\math R-valid in general
because then the required irreflexivity of 
\math{S_e\tight\circ R} implies 
\bigmath{S_e\tightequal\emptyset,}
and 
\math{
  e
  \funarg{\existsvari x{}}
  \funarg{\domres\pi{\revrelappsin{S_e}{\existsvari x{}}}}
  =
  e
  \funarg{\existsvari x{}}
  \funarg{\domres\pi\emptyset}
  =
  e
  \funarg{\existsvari x{}}
  \funarg\emptyset
}
cannot depend on 
\math{\pi\funarg{\forallvari y{}}}
anymore.
This means that \math{(\Vsome\tighttimes\Vall)}-validity of 
\math{
  \existsvari x{}
  \boldequal 
  \forallvari y{}
} 
is the same as validity of 
\bigmath{
  \exists x
  \stopq
  \forall y
  \stopq
  x\boldequal y
.}
Moreover, note that
\math{
  \epsilon(e)(\pi)
}
has no access to the \math\pi-value of 
\math{\forallvari y{}}
just as a raising function 
\math c for \nolinebreak\math x
in the raised version 
\math{ 
  c\boldequal \forallvari y{}
}
of
\bigmath{
  \exists x
  \stopq
  \forall y
  \stopq
  x\boldequal y
.}

For a more general example let
\math{
  \goals=
  \setwith
    {A_{i,0}\ldots A_{i,n_i-1}}
    {i\tightin I}
},
where for \math{i\in I} and \math{j\tightprec n_i} the \math{A_{i,j}} 
are formulas 
with \fev s from \math{\vec x}
and
\fuv s from \math{\vec y}.
Then \math{(\Vsome\tighttimes\Vall)}-validity of \goals\ means validity of
\bigmath{
  \exists\vec x
  \stopq
  \forall\vec y
  \stopq
  \forall i
  \tightin I
  \stopq
  \exists j
  \tightprec n_i
  \stopq
  A_{i,j}
;}
whereas \math\emptyset-validity of \goals\ means validity of
\bigmath{
  \forall\vec y
  \stopq
  \exists\vec x
  \stopq
  \forall i
  \tightin I
  \stopq
  \exists j
  \tightprec n_i
  \stopq
  A_{i,j}
.}

\end{example}

\pagebreak

\yestop
\yestop
\yestop
\noindent
Besides the notion of validity 
we need the notion of reduction.
Roughly speaking, 
a set \math{G_0} of sequents
reduces to 
a set \math{G_1} of sequents
if 
validity of \math{G_1}
implies
validity of \math{G_0}.
This, however, is too weak for our purposes here
because we are not only interested in validity but also
in preserving the solutions for the \fev s:
For inductive theorem proving,
answer computation, and constraint solving
it becomes important
that the solutions of \math{G_1} 
are also solutions of \math{G_0}.

\begin{definition}[Reduction]

\noindent
\math{G_0}\emph
{\math R-reduces to \nolinebreak\math{G_1} in \algebra}
\udiff\
for all 
existential \pair\algebra R-valuations \math e:

\noindent
\LINEnomath{
  if 
  \math{G_1} is \pair e\algebra-valid 
  then
  \math{G_0} is \pair e\algebra-valid, too.
}

\end{definition}

\yestop
\yestop
\begin{lemma}[Reduction]\label{lemma reduces to}

\noindent
Let \math R, \math{R'} be \vc s;
\algebra\ a \semanticobject;
\math{G_0}, \math{G_1}, \math{G_2}, 
and \math{G_3} sets of sequents.
Now:
\begin{enumerate}
\notop
\item{\bf(Validity)}
\\
If 
\math{G_0} 
\math R-reduces to \math{G_1} in \nolinebreak\algebra\
and
\math{G_1} is \math R-valid 
in \nolinebreak\algebra,
\\
then 
\math{G_0} is \math R-valid 
in \nolinebreak\algebra, too.

\item{\bf(Reflexivity)}
\\
In case of
\ \math{
  G_0\tightsubseteq G_1
}:
\ 
\math{G_0}
\math R-reduces to \math{G_1} 
in \nolinebreak\algebra.

\item{\bf(Transitivity)}
\\
If 
\math{G_0} \math R-reduces to \math{G_1} in \nolinebreak\algebra\
and
\math{G_1} \math R-reduces to \math{G_2} in \nolinebreak\algebra,
\\
then \math{G_0} \math R-reduces to \math{G_2} in \nolinebreak\algebra.

\item{\bf(Additivity)}
\\
If 
\math{G_0} \math R-reduces to \math{G_2} 
in \nolinebreak\algebra\
and
\math{G_1} \math R-reduces to \math{G_3} 
in \nolinebreak\algebra,
\\
then \math{G_0\tightcup G_1} \math R-reduces 
to \math{G_2\tightcup G_3} 
in \nolinebreak\algebra.

\item
{\bf(Monotonicity in \math R)}
\\
In case of  
\ \math{
  R\tightsubseteq R'
}: \ 
If \math{G_0} \math R-reduces to 
\math{G_1} in \nolinebreak\algebra,
then
\math{G_0} \math{R'}-reduces to 
\math{G_1} in \nolinebreak\algebra.

\item{\bf(Instantiation)}

For an \exRsub\ 
\nolinebreak\math\sigma,
and \math{R'} the \math\sigma-update of \math R:
\begin{enumerate}
\notop
\item
If 
\math{G_0\sigma} is \math{R'}-valid 
in \nolinebreak\algebra,
then
\math{G_0} is \math R-valid in 
\nolinebreak\algebra.

\item
If 
\math{G_0} \math R-reduces to 
\math{G_1} in \nolinebreak\algebra,
then
\math{G_0\sigma} \math{R'}-reduces 
to \math{G_1\sigma} 
in \nolinebreak\algebra. 

\end{enumerate}
\end{enumerate}
\end{lemma}

\pagebreak

\yestop
\noindent
Now we are going 
to abstractly
describe deductive sequent and tableau calculi.
We will later show that the usual
deductive first-order calculi are instances of our abstract calculi.
The benefit of the abstract version is that every
instance is automatically sound. Due to the small
number of inference rules in deductive first-order calculi
and the locality of soundness, this abstract version is not really
necessary. For inductive calculi, however, due to a bigger
number of inference rules 
(which usually have to be improved now and then)
and the globality of soundness, such an abstract version
is very helpful, \cf\ \cite{wirthbecker}, \cite{wirthdiss}.

\begin{definition}[Proof Forest]\label{definition proof forest}
\\
A\emph{(deductive) proof forest in a sequent} 
(\orelse\emph{tableau})\emph{calculus}
is a pair
\pair\foresteins R
where 
\math R \nolinebreak is a \vc\ 
and \foresteins\ is a set of pairs
\pair\Feins t, where
\Feins\ is a sequent 
and \math t is a tree\footnotemark\
whose nodes are labeled with sequents (\orelse\ formulas).
\end{definition}
Note that the tree \math t is intended to represent a proof attempt for 
\nolinebreak\Feins. The nodes of \nolinebreak\math t are labeled with formulas
in case of a tableau calculus and with sequents in case of a sequent
calculus. While the sequents at the nodes of a tree in a 
sequent calculus stand
for themselves, in a tableau calculus all the ancestors have to be
included to make up a sequent and, moreover, the formulas at the 
labels are in negated form:

\begin{definition}[\Openofset{}, \Ax, Closedness]
\label{definition open to closedness}

\noindent
`\Openofset{T}' denotes the set of sequents labeling the leaves
of the trees in the set \math T
(\orelse\ the set of sequents resulting from listing 
 the conjugates of the formulas
 labeling a branch from a leaf to the root in a tree in \math T).

\noindent
\Inthesequel, we assume \Ax\ to be some set of\emph{axioms}.
By this we mean that \Ax\ is 
\math{\Vsome\tighttimes\Vall}-valid.
(\Cf\ the last sentence in \defiref{definition weak validity}.)

\noindent
The tree \nolinebreak\math t is\emph{closed}
\udiff\
\bigmath{\Open{t}\subseteq\Ax.}
\end{definition}

\yestop
\noindent
The readers may ask themselves why we consider a proof forest
instead of a single proof tree only. The possibility to have
an empty proof forest provides a nicer starting point.
Besides that, if we have trees 
\math{\pair\Feins t,\ \pair\Feinsprime{t'}\in\foresteins}
we can apply \Feins\ as a lemma in the tree \math{t'}
of \Feinsprime, provided that the lemma application relation is 
acyclic. For deductive theorem proving the availability of 
lemma application is not really necessary. For inductive theorem proving,
however, lemma and induction hypothesis application of this
form becomes necessary.

\begin{definition}[Invariant Condition]
\\
The\emph{invariant condition on \pair\foresteins R} is that
\math{\{\Feins\}} \math R-reduces to \Open{t}
for all \math{\pair\Feins t\in\foresteins}.
\end{definition}

\begin{theorem}\label{theorem closed means valid}
\\
Let the proof forest \pair\foresteins R satisfy the above invariant condition.
Let 
\bigmath{
  (\Feins,t)\tightin\foresteins
.}
\\
If\/ \bigmath t \nolinebreak is closed,
then\/ \Feins\ is \math R-valid.
\end{theorem}

\pagebreak

\yestop
\yestop
\yestop
\begin{theorem}\label{theorem abstract deductive calculi}

\noindent
The above invariant condition is always satisfied 
when we start with an empty proof forest 
\math{\pair\foresteins R:=\pair\emptyset\emptyset} and then
iterate only the following kinds of modifications 
of\/ \pair\foresteins R (resulting in \pair\forestzwei{R'}):

\begin{description}

\notop\halftop
\item[Hypothesizing:]
Let \math{R'} be a \vc\ with 
\bigmath{R\tightsubseteq R'.}
Let \Feins\ be a sequent.
Let \math t \nolinebreak be the tree with a single node only,
which is labeled with \Feins\
(\orelse\ with a single branch only, such that \Feins\ 
 is the list of the conjugates of the formulas
 labeling the branch from the leaf to the root).
Then we may set 
\bigmath{
  \forestzwei\nottight{:=}\foresteins\cup\{(\Feins,t)\}
.} 

\item[Expansion:]
Let \math{(\Feins,t)\in\foresteins}.
Let \math{R'} be a \vc\ with \bigmath{R\tightsubseteq R'.}
Let \math l be a leaf in \nolinebreak\math t.
Let \Fzwei\ be the label of\/ \nolinebreak\math l
(\orelse\ result from listing the conjugates of the formulas
 labeling the branch from \math l to the root of\/ \nolinebreak\math t).
Let \goals\ be a finite set of sequents
.
Now if\/ \math{\{\Fzwei\}} \math{R'}-reduces to \math G 
(\orelse\ \setwith{\Fvier\Fzwei}{\Fvier\tightin\goals}),
then we may set
\bigmath{
  \forestzwei
  \nottight{:=}
  (\foresteins
   \tightsetminus
   \{(\Feins,t)\}
  )
  \cup
  \{(\Feins,t')\}
} 
where \math{t'} results from \math t by adding
to the former leaf\/ \math l, 
exactly for each sequent \Fvier\ in \goals, 
a new child node labeled with \Fvier\
(\orelse\ a new child branch such that \Fvier\ \nolinebreak
 is the list of the conjugates of the formulas
 labeling the branch from the leaf to the new child node of\/ \nolinebreak\math l).

\item[Instantiation:]
Let \math\sigma\ be an \exRsub.
Let \math{R'} be the \math\sigma-update of \math R.
Then we may set
\bigmath{
  \forestzwei
  \nottight{:=}
  \foresteins\sigma
.} 

\notop\halftop
\end{description}
\end{theorem}

\yestop
\yestop
\yestop
\noindent
While Hypothesizing and Instantiation steps are self-explanatory,
Expansion steps are parameterized by a sequent \Fzwei\
and a set of sequents \math G such that 
\math{\{\Fzwei\}} \math{R'}-reduces to \math G.
For tableau calculi, however,
this set of sequents must actually have the form
\setwith{\Fvier\Fzwei}{\Fvier\tightin\goals}
because an Expansion step cannot remove formulas from
ancestor nodes. This is because these formulas are also
part of the goals associated with other leaves in the proof tree.
Therefore, although tableau
calculi may save repetition of formulas,
sequent calculi have substantial advantages:
Rewriting of formulas in place is always possible,
and we can remove formulas that are redundant
\wrt\ the other formulas in a sequent. 
But this is not our subject here. 
For the below examples of \math\alpha-, \math\beta-, \math\gamma-, and
\math\delta-rules we will use the sequent calculi presentation
because it is a little more explicit.
When we write

\noindent
\expansionrule\Fzwei{\Fdrei_0\quad\ldots\quad\Fdrei_{n-1}}{R''}{}{}{}

\noindent
we want to denote a sub-rule of the Expansion rule
which is given by \bigmath{G:=\{\Fdrei_0,\ldots,\Fdrei_{n-1}\}}
and \bigmath{R':=R\cup R''.}
This means that for this rule really being a sub-rule of
the Expansion rule we have to show that
\math{\{\Fzwei\}} \math{R'}-reduces to \math G.
By \lemmref{lemma reduces to}(5) and because \math R does not 
matter here, it suffices that we actually show that 
\math{\{\Fzwei\}} \math{R''}-reduces to \math G.
Moreover, note that in old times when trees grew upwards,
\gentzenname\ would have written 
\bigmath{\Fdrei_0\quad\ldots\quad\Fdrei_{n-1}} 
above the line and \math\Fzwei\ below, 
such that passing the line meant implication.
In our case, passing the line means reduction.

\pagebreak

\noindent
Let \math{A} and \math{B} be formulas,
\math{\Gamma} and \math{\Pi} sequents,
\math{x\in\Vbound}, 
\bigmath{
  \existsvari x{}
  \nottight\in
  \Vsome
  \setminus
  \VARsome{A,\Gamma\Pi}
,}\footnotemark\ 
and
\bigmath{
  \forallvari x{}
  \nottight\in
  \Vall
  \setminus
  \VARall{A,\Gamma\Pi}
.}

\noindent
\math\alpha-rules:
\\
\expansionrule
{\Gamma~~\inparenthesesinlinetight{A\tightoder B}~~\Pi}
{A~~B~~\Gamma~~\Pi}
\emptyset{}{}{}

\noindent
\expansionrule
{\Gamma~~\neg\inparenthesesinlinetight{A\tightund B}~~\Pi}
{\headroom\overline{\,A\,}~~\headroom\overline{\,B\,}~~\Gamma~~\Pi}
\emptyset{}{}{}

\noindent
\expansionrule
{\Gamma~~\neg\neg A~~\Pi}
{A~~\Gamma~~\Pi}
\emptyset{}{}{}
\\

\noindent
\math\beta-rules:
\\
\expansionrule
{\Gamma~~\inparenthesesinlinetight{A\tightund B}~~\Pi}
{A~~\Gamma~~\Pi~~~~~~~~~~~~B~~\Gamma~~\Pi}
\emptyset{}{}{}

\noindent
\expansionrule
{\Gamma~~\neg\inparenthesesinlinetight{A\tightoder B}~~\Pi}
{\headroom\overline{\,A\,}~~\Gamma~~\Pi~~~~~~~~~~~~\headroom\overline{\,B\,}~~\Gamma~~\Pi}
\emptyset{}{}{}
\\

\noindent
\math\gamma-rules:
\\
\expansionrule
{\Gamma~~\exists x\tight:\, A~~\Pi}
{A\{x\tight\mapsto\existsvari x{}\}~~\Gamma~~\exists x\tight:\, A~~\Pi}
\emptyset{}{}{}

\noindent
\expansionrule
{\Gamma~~\neg\forall x\tight:\, A~~\Pi}
{\majorheadroom\overline{\,A\{x\tight\mapsto\existsvari x{}\}\,}~~\Gamma~~\neg\forall x\tight:\, A~~\Pi}
\emptyset{}{}{}
\\

\noindent
\math\delta-rules:
\\
\expansionrule
{\Gamma~~\forall x\tight:\, A~~\Pi}
{A\{x\tight\mapsto\forallvari x{}\}~~\Gamma~~\Pi}
{\VARsome{A,\Gamma\Pi}\times\{\forallvari x{}\}}
{}{}{}

\noindent
\expansionrule
{\Gamma~~\neg\exists x\tight:\, A~~\Pi}
{\majorheadroom\overline{\,A\{x\tight\mapsto\forallvari x{}\}\,}~~\Gamma~~\Pi}
{\VARsome{A,\Gamma\Pi}\times\{\forallvari x{}\}}
{}{}{}\\

\yestop
\begin{theorem}\label{theorem sub-rules}
\\
The above examples of 
\math\alpha-, \math\beta-, \math\gamma-, and
\math\delta-rules 
are all sub-rules of the Expansion rule of the sequent calculus
of \theoref{theorem abstract deductive calculi}.
\end{theorem}
\vfill\pagebreak
\section{The Strong Version}
\label{section strong}

\yestop
\noindent
The additional solutions (or existential substitutions)
of the strong version
(which admit additional proofs compared to the weak version) 
do not add much difficulty
when one is interested in validity only, \cf\ \eg\ \cite{deltaplus}.
When also the preservation of solutions is required, however,
the additional substitutions pose some problems because the 
new solutions may tear some \fuv s out of their contexts:

\yestop
\begin{example}[Reduction \& Liberalized \math\delta-Steps]
\label{example eta}

\noindent
In \examref{example liberalized}
a liberalized \math\delta-step reduced

\noindent
\LINEmath{
      \Pppp{\existsvari x{}}
  \comma\ 
      \forall y \stopq
      \neg\Pppp y
}
\\to
\\\LINEmath{
      \Pppp{\existsvari x{}}
  \comma
      \neg\Pppp{\forallvari y{}}
}

\noindent
with empty \vc\ \math{R:=\emptyset.}

\noindent
The latter sequent is \pair e\algebra-valid for the strong existential
\pair\algebra R-valuation \math e given by
\\\LINEmath{
  e
  \funarg{
    \existsvari x{}
  }
  \funarg\pi
  :=
  \pi
  \funarg{
    \forallvari y{}
  }
.}

\noindent
The former sequent, however, is not \pair e\algebra-valid 
when \math{\Ppsymbol^\algebra\funarg a} is true 
and \math{\Ppsymbol^\algebra\funarg b} is false
for some \math a, \math b from the universe of \algebra.
To see this, take some \math\pi\ with 
\math{\pi\funarg{\forallvari y{}}:=b}.
\end{example}

\yestop
\yestop
\yestop
\noindent
How can we solve the problem exhibited in \examref{example eta}?
\Ie\ how can we change the notion of reduction 
such that the liberalized \math\delta-step becomes a reduction step?

\begin{enumerate}

\yestop
\item
The approach we tried first was to allow a slight modification of \bigmath e
to \bigmath{e'} such that 
\bigmath{
  e'\funarg{\existsvari x{}}\funarg\pi\tightequal a
.}
This trial finally failed because it was not possible to preserve reduction
under Instantiation-steps.

\Eg, an Instantiation-step with the strong existential 
\math R-substitution \math{\{\existsvari x{}\tight\mapsto\forallvari y{}\}}
transforms the reduction of
\examref{example eta} into the reduction of 

\noindent
\LINEmath{
      \Pppp{\forallvari y{}}
  \comma\ 
      \forall y \stopq
      \neg\Pppp y
}
\\
to
\\\LINEmath{
      \Pppp{\forallvari y{}}
  \comma
      \neg\Pppp{\forallvari y{}}
.}

\noindent
Taking \math\pi, \math e, and \algebra\ as in \examref{example eta},
the new latter sequent is still \pair e\algebra-valid.
There is, however, no modification \bigmath{e'} of \bigmath e
such that the new former sequent is \trip\pi{e'}\algebra-valid.

Thus, with this approach,
reduction could not be preserved by Instantiation-steps.

Moreover, the modification of \bigmath e does not go together well
with our requirement of preservation of solutions.

\pagebreak

\yestop
\item
Learning from this, the second approach we tried was to 
allow a slight modification of \nolinebreak\bigmath\pi\ instead.
\Eg, for the reduction step of \examref{example eta}, 
we would require the existence of some 
\math{\eta\in\FUNSET{\{\forallvari y{}\}}\algebra}
such that the former sequent is 
\trip
  {\domres\pi{\Vall\setminus\{\forallvari y{}\}}\tight\uplus\eta}
  e
  \algebra
-valid
instead of
\trip\pi e\algebra-valid.
Choosing \math{\eta:=\{\forallvari y{}\tight\mapsto a\}}
would solve the problem of \examref{example eta} then:
Indeed, the former sequent is 
\trip
  {\domres\pi{\Vall\setminus\{\forallvari y{}\}}\tight\uplus\eta}
  e
  \algebra
-valid 
because for the \bigmath e of \examref{example eta} we have
\bigmath{
  e
  \funarg{
    \existsvari x{}
  }
  \funarg{\domres\pi{\Vall\setminus\{\forallvari y{}\}}\tight\uplus\eta}
  =
  \inparenthesesinlinetight{\domres\pi{\Vall\setminus\{\forallvari y{}\}}\tight\uplus\eta}
  \funarg{
    \forallvari y{}
  }
  =
  a
.}

Moreover, with this approach, 
reduction is preserved under Instantiation-steps.

The problems with this approach arise, however,
when one asks whether there has
to be a single \math\eta\ for all \math\pi\ or, for each \math\pi,
a different \math\eta.

If we require a single \math\eta, 
we cannot model liberalized \math\delta-steps
where another \fuv, say \forallvari z{}, occurs in the principal formula,
as, \eg, in the reduction of 

\noindent
\LINEmath{
      \forallvari z{}
      \boldequal
      \existsvari x{}
  \comma\ 
      \forall y \stopq
        \forallvari z{}
        \boldunequal
        y
}
\\to
\\\LINEmath{
      \forallvari z{}
      \boldequal
      \existsvari x{}
  \comma
        \forallvari z{}
        \boldunequal
        \forallvari y{}
}

\noindent
with empty \vc.
In this case, for the \bigmath e of \examref{example eta}
(which gives \existsvari x{} the value of \forallvari y{})
the \math{\eta\in\FUNSET{\{\forallvari y{}\}}\algebra}
must change when the \math\pi-value of \forallvari z{} changes:
\Eg, for 
\math{
  \pi
  :=
  \{
    \forallvari y{}
    \tight\mapsto
    a
  \comma
    \forallvari z{}
    \tight\mapsto
    b
  \}
}
we need \math{\eta\funarg{\forallvari y{}}:=b},
while for 
\math{
  \pi
  :=
  \{
    \forallvari y{}
    \tight\mapsto
    b
  \comma
    \forallvari z{}
    \tight\mapsto
    a
  \}
}
we need \math{\eta\funarg{\forallvari y{}}:=a}.
Indeed, in the reduction above,
\forallvari y{} is functionally dependent on
\forallvari z{}.

If, on the other hand, we admit a different \math\eta\ for each
\math\pi, the transitivity of reduction (\cf\ \lemmref{lemma reduces to}(3))
gets lost.

Thus, the only solution can be that \math\eta\ depends on some values
of \math\pi\ and not on others. Since the abstract treatment of this
gets very ugly and does not extract much information on the solution
of \fev s of the original theorem from a completed proof, we prefer to
remember what role the \fuv s introduced by liberalized \math\delta-steps
really play. And this is what the following definition is about.

\yestop
\end{enumerate}

\vfill

\yestop
\begin{definition}[\CC, Extension]

\noindent
\math C is a\emph{\pair R <-\cc}
\udiff\
\math C is a (possibly) partial function from \Vall\ into the
set of formulas,
\math R is a \vc,
\math < is a wellfounded ordering on \Vall\
with \bigmath{\inparenthesesinlinetight{R\circ\tight<}\subseteq R,}
and, for all \math{\forallvari y{}\in\DOM C}:

\noindent
\LINEnomath{
  \bigmath{
    \forallvari z{}
     <
    \forallvari y{}  
  }
  for all 
  \math{
    \forallvari z{}
    \in
    \VARall{C\funarg{\forallvari y{}}}
    \tightsetminus
    \{\forallvari y{}\}
  }
}
\\and
\\\LINEnomath{
  \bigmath{
    \existsvari u{}
    \nottight R
    \forallvari y{}  
  }
  for all 
  \math{
    \existsvari u{}
    \in
    \VARsome{C\funarg{\forallvari y{}}}
  }.
}

\noindent
\trip{C'}{R'}{\tight<'} is an\emph{extension of} \trip C R < \udiff\
\bigmath{C\tightsubseteq C',}
\bigmath{R\tightsubseteq R',}
and \math{C'} is a \pair{R'}{\tight<'}-\cc.
\end{definition}
Note that \math\emptyset\ is a \pair R\emptyset-\cc\ for
any \vc\ \nolinebreak\math R.
For the meaning of \cc s \cf\ \defiref{definition compatibility}.

\pagebreak

\begin{definition}[Extended Strong \math\sigma-Update]\label{definition ex str s up}

\noindent
Let \math C be a \pair R <-\cc\ and \math\sigma\ a strong existential
\math R-substitution.

\noindent
The\emph{extended strong \math\sigma-update 
\bigmath{\trip{C'}{R'}{<'}} of \bigmath{\trip C R <}}
is given by\\
\LINEmath{
  C'
  \nottight{\nottight{:=}}
  \setwith{\pair x{B\sigma}}{\pair x B\tightin C}
,}\\
\LINEnomath{
  \bigmath{R'} is the strong \math\sigma-update of \bigmath{R,}
}\\
\LINEmath{
  \tight{<'}
  \nottight{\nottight{\nottight{:=}}}
  \tight<
  \circ
  \refltransclosureinline{\inparenthesesinlinetight{U_\sigma\tight\circ R}}
  \nottight{\nottight\cup}
  \transclosureinline{\inparenthesesinlinetight{U_\sigma\tight\circ R}}
.}
\end{definition}
\begin{lemma}[Theorem~62 in \cite{geserthmimproved}]
\label{lemma geserthmimproved}
\\
If \math A and \math B are two terminating relations
with 
\bigmath{
  A\tight\circ B
  \nottight{\nottight\subseteq}
  A
  \nottight{\nottight\cup}
  B 
  \tight\circ
  \refltransclosureinline{\inparenthesesinlinetight{A\cup B}}
,}
\\
then \bigmath{A\cup B} is terminating, too.
\end{lemma}
\begin{lemma}[Extended Strong \math\sigma-Update]\label{lemm ex str s up}\\
Let \math C be a \pair R <-\cc, \math\sigma\ a strong existential
\math R-substitution, and \trip{C'}{R'}{<'} the extended strong 
\math\sigma-update of \trip C R <. Now:
\math{C'} is a \pair{R'}{<'}-\cc.
\end{lemma}

\begin{definition}[Compatibility]\label{definition compatibility}
\\
Let \math C be a \pair R <-\cc,
\algebra\ a \semanticobject,
and \math e a strong existential \pair\algebra R-valuation.
\\
We say that \math\pi\ is \emph{\pair e\algebra-compatible with \math C}
\udiff\
\bigmath{\pi\tightin\FUNSET\Vall\algebra}
and for each \math{\forallvari y{}\in\DOM C}:
\begin{quote}
If \math{C\funarg{\forallvari y{}}} is \trip\pi e\algebra-valid,
\\
then \math{C\funarg{\forallvari y{}}} is 
\trip
  {\domres\pi{\Vall\setminus\{\forallvari y{}\}}\uplus\eta}
  e
  \algebra
-valid
for all \math{\eta\in\FUNSET{\{\forallvari y{}\}}\algebra}.
\end{quote}
\end{definition}

\yestop
\noindent
Note that \pair e\algebra-compatibility of 
\math\pi\ 
with 
\math{\{\pair{\forallvari y{}}B\}}
means that a different choice for the \math\pi-value of 
\forallvari y{} does not destroy the validity of the formula \math B
in
\bigmath{
  \algebra
  \nottight\uplus
  \epsilon
  \funarg e
  \funarg\pi
  \nottight\uplus
  \pi
,}
or that \math{\pi\funarg{\forallvari y{}}} is chosen such that
\math B becomes invalid if such a choice is possible, which is
closely related to \hilbert's \math\varepsilon-operator
(~\math{
  \forallvari y{}
  \nottight{\nottight=}
  \varepsilon y\stopq
  \inparenthesesinlinetight{
    \neg B\{\forallvari y{}\tight\mapsto y\}
  }
}~).

\yestop\yestop\yestop\yestop\noindent
We are now going to proceed like in the previous section, 
but using the strong versions instead of the weak ones.

\yestop
\begin{definition}[Strong Validity]

\noindent
Let \math C be a \pair R <-\cc,
\algebra\ a \semanticobject, and \math G a set of sequents.

\noindent
\math G is\emph{\stronglyvalidinA R C in \nolinebreak\algebra} 
\udiff\
there is a strong
existential \math{(\algebra,R)}-valuation \bigmath e such that
\goals\ is \stronglyvalid e\algebra C.

\noindent
\goals\ 
is\emph{\stronglyvalid e\algebra C} \udiff\
\goals\ is \trip\pi e\algebra-valid 
for each \math\pi\ that is \pair e\algebra-compatible with \nolinebreak\math C.

\noindent
The rest is given by \defiref{definition weak validity}.
\end{definition}

\yestop
\begin{lemma}[Anti-Monotonicity in \math R and Monotonicity in \math C]
\label{lemma strong reduces to 5a}\\
Let \math G be a set of sequents, \math{C} a \pair{R}{<}-\cc, and
\math{C'} a \pair{R'}{<'}-\cc\
with \bigmath{R\tightsubseteq R'} and \bigmath{C'\tightsubseteq C.}
Now:\\ 
If   \math{G} is \stronglyvalidinA{R'}{C'} in \nolinebreak\algebra,
then \math{G} is \stronglyvalidinA R C     in \nolinebreak\algebra.
\end{lemma}

\pagebreak

\yestop
\begin{example}[Strong Validity]\sloppy

\noindent
Note that 
\math\emptyset-validity does not differ from 
\strongvalidityinA\emptyset\emptyset\ and 
that
\math{\Vsome\tighttimes\Vall}-validity does not differ from 
\strongvalidityinA{\Vsome\tighttimes\Vall}\emptyset.
This is because the notions of weak and strong existential valuations
do not differ in these cases. 
Therefore, \examref{ex validity}
is also an example for strong validity.

Although \strongvalidityinA R\emptyset\ always implies (weak) 
\math R-validity 
(because each strong existential \pair\algebra R-valuation is
 a (weak) existential \pair\algebra R-valuation),
for \math R not being one of the extremes \math\emptyset\
and \math{\Vsome\tighttimes\Vall},
(weak) \math R-validity and \strongvalidityinA R\emptyset\ 
differ from each other.
\Eg\ the sequent (\math\ast) in \examref{ex kohlhase fail}
is (weakly) \math R-valid but not \stronglyvalidinA R\emptyset\
for the \math R \nolinebreak of \nolinebreak(\#):
For 
\math{
  S_e
  :=
  \{
    \pair{\forallvari y{}}{\existsvari u{}}
  ,\ 
    \pair{\forallvari v{}}{\existsvari x{}}
  \}
}
we get 
\bigmath{
  S_e\tight\circ R
  =
    \{
    \pair{\forallvari y{}}{\forallvari v{}}
  ,\ 
    \pair{\forallvari v{}}{\forallvari y{}}
  \}
,}
which is irreflexive.
Since the sequent (\math\ast) is \pair e\algebra-valid for the (weak) 
\math existential \pair\algebra R-valuation \nolinebreak\math e given by
\bigmath{
  e
  \funarg
    {\existsvari x{}}
  \funarg{\domres\pi{\revrelappsin{S_e}{\existsvari x{}}}}
  =
  \pi
  \funarg{\forallvari v{}}
}
and
\bigmath{
  e
  \funarg
    {\existsvari u{}}
  \funarg{\domres\pi{\revrelappsin{S_e}{\existsvari u{}}}}
  =
  \pi
  \funarg{\forallvari y{}}
,}
the sequent (\math\ast) is (weakly) \math R-valid in \algebra\@. \ 
But 
\bigmath{
  \inparenthesesinlinetight{S_e\tight\circ R}^2
}
is not irreflexive,
so that this \bigmath e is no\emph{strong} 
existential \pair\algebra R-valuation,
which means that the sequent (\math\ast) cannot be 
\stronglyvalidinA R\emptyset\ in general.

For nonempty \math C, however, we must admit that 
\strongvalidityinA R C is hard to understand. 
We have to make sure that
\strongvalidityinA R C can be easily understood in terms
of \strongvalidityinA{R'}\emptyset\ for some \math{R'}, which again implies
(weak) \math{R'}-validity and \math\emptyset-validity. 
Note that this difficulty did not arise in the weak version
because \lemmref{lemma reduces to 5a} states anti-monotonicity of 
(weak) \math R-validity in \nolinebreak\math R, 
whereas \lemmref{lemma strong reduces to 5a} 
states anti-monotonicity of \strongvalidityinA R C in \math R but only
monotonicity of \strongvalidityinA R C in \math C.
\end{example}

\yestop
\yestop
\begin{lemma}[Compatibility and Validity]\label{lemma compatibility}

\noindent
Let \bigmath\algebra\ be a \semanticobject,
\bigmath C a \pair R <-\cc,
and
\bigmath e a strong existential \pair\algebra R-valuation.
\\
Define
\bigmath{
  \tight\lhd
  :=
  \transclosureinline{\inparenthesesinlinetight{S_e\cup R\cup\tight<}}
.}
\begin{enumerate}

\noitem
\item
\math\lhd\ is a wellfounded ordering on \Vfree.

\item
There is a function 
\bigmath{
  \FUNDEF
    \xi
    {\inparentheses{
       \FUNSET
         {\inparenthesesinlinetight{\Vall\tightsetminus\DOM C}}
         \algebra}}
    {\inparentheses{
       \FUNSET
         {\DOM C}
         \algebra}}
}
such that, 
\\
for all 
\math{
  \pi,\pi'
  \in
  \FUNSET
    {\inparenthesesinlinetight{
       \Vall\tightsetminus\DOM C}}
    \algebra
,}
\bigmath{\pi\uplus\xi_\pi} is \pair e\algebra-compatible with \math C,
and, 
\\
for \math{x\in\DOM C},
\bigmath{
  \domres\pi{\revrelappsin\lhd x}
  =
  \domres{\pi'}{\revrelappsin\lhd x}
}
implies
\bigmath{
  \xi_\pi\funarg x
  =
  \xi_{\pi'}\funarg x
.}

\item
Let \math G be a set of sequents
and
\math{
 \varsigma
 \in
 \FUNSET
   {\inparenthesesinlinetight{\VARall G\cap\DOM C}}
   {\inparenthesesinlinetight{\Vsome\tightsetminus\VARsome G}}
}
be injective.
\begin{enumerate}
\item
If 
\math G is \stronglyvalid e\algebra C,
then 
\math{G\varsigma} is 
\stronglyvalidinA{R'}\emptyset\ in \nolinebreak\algebra\
\\
for 
\bigmath{
  R'
  \nottight{\nottight{\nottight{:=}}}
  \domres R{\Vsome\setminus\RAN\varsigma}
  \ \nottight{\nottight\cup}
  \displaystyle\bigcup_{y\in\RAN\varsigma}
    \{y\}
    \times
    \relappsin\unlhd{\reverserelation\varsigma\funarg y}
  \nottight{\nottight{\nottight\cup}}
  \Vsome\times\DOM C
,}
\\
where 
\math\unlhd\nolinebreak\ is the reflexive ordering on \nolinebreak\Vall\ 
of \math\lhd.
\item
If 
\math G is 
\stronglyvalidinA R C
in \nolinebreak\algebra,
then 
\math{G\varsigma} is
\stronglyvalidinA{\domres R{\Vsome\setminus\RAN\varsigma}}\emptyset\
in \nolinebreak\algebra\
and even
\stronglyvalidinA{R''}\emptyset\ in \nolinebreak\algebra\
\\
for 
\bigmath{
  R''
  \nottight{\nottight{\nottight{:=}}}
  \domres R{\Vsome\setminus\RAN\varsigma}
  \ \nottight{\nottight\cup}
  \displaystyle\bigcup_{y\in\RAN\varsigma}
    \{y\}
    \times
    \relappsin{\tight <}{\reverserelation\varsigma\funarg y}
  \nottight{\nottight{\nottight\cup}}
  \Vsome\times\DOM C
.}
\end{enumerate}
\end{enumerate}
\end{lemma}

\pagebreak

\yestop
\yestop
\begin{definition}[Strong Reduction]

\noindent
Let \math C be a \pair R <-\cc,
\algebra\ a \semanticobject, and \math{G_0}, \math{G_1} sets of sequents.

\noindent
{\em \math{G_0} strongly \pair R C-reduces to \nolinebreak\math{G_1} in \algebra\/}
\udiff\
for each strong existential \pair\algebra R-valuation \math e
and each \math\pi\ that is \pair e\algebra-compatible with \math C:

\noindent
\LINEnomath{
  if \math{G_1} is \trip\pi e\algebra-valid,
  then 
  \math{G_0} is \trip\pi e\algebra-valid.
}
\end{definition}

\yestop
\yestop
\begin{lemma}[Strong Reduction]\label{lemma strong reduces to}

\noindent
Let \math C be a \pair R <-\cc;
\algebra\ a \semanticobject;
\math{G_0}, \math{G_1}, \math{G_2}, 
and \math{G_3} sets of sequents.
Now:
\begin{enumerate}

\noitem
\item{\bf(Validity)}

\noindent
Assume that \math{G_0} strongly \pair R C-reduces to \math{G_1} 
in \nolinebreak\algebra\@. Now:

If   \math{G_1} is \stronglyvalid e\algebra C
for some strong existential \pair\algebra R-valuation \ \nolinebreak\math e,
\\
then \math{G_0} is \stronglyvalid e\algebra C.

If   \math{G_1} is \stronglyvalidinA R C in \nolinebreak\algebra,
then \math{G_0} is \stronglyvalidinA R C in \nolinebreak\algebra.

\item{\bf(Reflexivity)}
\\
In case of
\ \math{
  G_0\tightsubseteq G_1
}:
\ 
\math{G_0}
strongly \pair R C-reduces to \math{G_1} 
in \nolinebreak\algebra.

\item{\bf(Transitivity)}
\\
If \math{G_0} strongly \pair R C-reduces to \math{G_1} in 
\nolinebreak\algebra\
and \math{G_1} strongly \pair R C-reduces to \math{G_2} in 
\nolinebreak\algebra,\\
then \math{G_0} strongly \pair R C-reduces to \math{G_2} 
in \nolinebreak\algebra.

\item{\bf(Additivity)}
\\
If \math{G_0} strongly \pair R C-reduces to \math{G_2} 
in \nolinebreak\algebra\ and
\math{G_1} strongly \pair R C-reduces to \math{G_3} 
in \nolinebreak\algebra,\\
then \math{G_0\tightcup G_1}  strongly \pair R C-reduces
to \math{G_2\tightcup G_3} in \nolinebreak\algebra.

\item{\bf(Monotonicity)}\\
For \trip{C'}{R'}{<'} being an extension of\/ \trip C R <:\\
If \math{G_0} strongly \pair R C-reduces to \math{G_1} 
in \nolinebreak\algebra,
then \math{G_0} strongly \pair{R'}{C'}-reduces to \math{G_1} 
in \nolinebreak\algebra.

\item{\bf(Instantiation)}

\noindent
For a strong \exRsub\ \ \nolinebreak\math{\sigma}, \ and 
the extended strong \math\sigma-update\/ \bigmath{\trip{C'}{R'}{<'}} 
of\/ \trip C R <:
\begin{enumerate}
\noitem
\item
If   \math{G_0\sigma} is \stronglyvalidinA{R'}{C'} in \nolinebreak\algebra,
then \math{G_0}       is \stronglyvalidinA R C     in \nolinebreak\algebra.

\item
If 
\math{G_0} strongly \pair R C-reduces to \math{G_1} in \nolinebreak\algebra,
\\
then \math{G_0\sigma} strongly \pair{R'}{C'}-reduces to \math{G_1\sigma} 
in \nolinebreak\algebra. 
\end{enumerate}
\end{enumerate}
\end{lemma}

\pagebreak

\yestop
\noindent
Now we are going 
to abstractly
describe deductive sequent and tableau calculi.
We will later show that the usual
deductive first-order calculi are instances of our abstract calculi.

\begin{definition}[Strong Proof Forest]\label{definition strong proof forest}
\\
A\emph{strong (deductive) proof forest in a sequent} 
(\orelse\emph{tableau})\emph{calculus}
is a quadruple
\quar\foresteins C R <
where \math C is a \pair R <-\cc\
and \foresteins\ is a set of pairs
\pair\Feins t, where
\Feins\ is a sequent 
and \math t is a tree
whose nodes are labeled with sequents (\orelse\ formulas).
\end{definition}
The notions of \Openofset{}, \Ax, and closedness of 
\defiref{definition open to closedness} are not changed.
Note, however, that the \math{\Vsome\tighttimes\Vall}-validity of
\Ax\ immediately implies the 
\strongvalidityinA{\Vsome\tighttimes\Vall}\emptyset\ of \Ax,
which (by \lemmref{lemma strong reduces to 5a}) is the 
logically strongest
kind of \strongvalidityinA R C.

\begin{definition}[Strong Invariant Condition]
\\
The\emph{strong invariant condition on \quar\foresteins C R <} is that
\math{\{\Feins\}} strongly \pair R C-reduces to \Open{t}
for all \math{\pair\Feins t\in\foresteins}.
\end{definition}

\begin{theorem}\label{theorem strong closed means valid}
\\
Let the strong proof forest
\quar\foresteins C R < satisfy the above strong invariant condition.
Let \bigmath{(\Feins,t)\tightin\foresteins}
and \bigmath t \nolinebreak be closed.
Now:
\\
\Feins\ is \stronglyvalidinA R C
and, for any injective
\math{
 \varsigma
 \in
 \FUNSET
   {\inparenthesesinlinetight{\VARall\Feins\cap\DOM C}}
   {\inparenthesesinlinetight{\Vsome\tightsetminus\VARsome\Feins}}
},\\
\math{\Feins\varsigma} is 
\stronglyvalidinA{\domres R{\Vsome\setminus\RAN\varsigma}}\emptyset\
and even
\stronglyvalidinA{R'}\emptyset\ 
for \\
\bigmath{
  R'
  \nottight{\nottight{\nottight{:=}}}
  \domres R{\Vsome\setminus\RAN\varsigma}
  \ \nottight{\nottight\cup}
  \displaystyle\bigcup_{y\in\RAN\varsigma}
    \{y\}
    \times
    \relappsin{\tight <}{\reverserelation\varsigma\funarg y}
  \nottight{\nottight{\nottight\cup}}
  \Vsome\times\DOM C
.}
\end{theorem}

\yestop
\yestop
\begin{theorem}\label{theorem strong abstract deductive calculi}

\noindent
The above strong invariant condition is always satisfied 
when we start with an empty strong proof forest 
\math{\quar\foresteins C R <:=\quar\emptyset\emptyset\emptyset\emptyset} 
and then
iterate only the following kinds of modifications 
of\/ \quar\foresteins C R < (resulting in \quar\forestzwei{C'}{R'}{<'}):

\begin{description}

\notop\halftop
\item[Hypothesizing:]
Let \math{R':=R\tightcup R''} be a \vc\ with
\bigmath{\inparenthesesinlinetight{R''\tight\circ\tight<}\subseteq R'.}
Set \math{C':=C} and \math{\tight{<'}:=\tight<}.
Let \Feins\ be a sequent.
Let \math t \nolinebreak be the tree with a single node only,
which is labeled with \Feins\
(\orelse\ with a single branch only, such that \Feins\ 
 is the list of the conjugates of the formulas
 labeling the branch from the leaf to the root).
Then we may set 
\bigmath{
  \forestzwei\nottight{:=}\foresteins\cup\{(\Feins,t)\}
.} 

\item[Expansion:]
Let \trip{C'}{R'}{<'} be an extension of \trip C R <.
Let \math{(\Feins,t)\in\foresteins}.
Let \math l be a leaf in \nolinebreak\math t.
Let \Fzwei\ be the label of\/ \nolinebreak\math l
(\orelse\ result from listing the conjugates of the formulas
 labeling the branch from \math l to the root of\/ \nolinebreak\math t).
Let \goals\ be a finite set of sequents.
Now if\/ \math{\{\Fzwei\}} strongly \pair{R'}{C'}-reduces to \math G 
(\orelse\ \setwith{\Fvier\Fzwei}{\Fvier\tightin\goals}),
then we may set
\bigmath{
  \forestzwei
  \nottight{:=}
  (\foresteins
   \tightsetminus
   \{(\Feins,t)\}
  )
  \cup
  \{(\Feins,t')\}
} 
where \math{t'} results from \math t by adding
to the former leaf\/ \nolinebreak\math l, 
exactly for each sequent \Fvier\ in \goals, 
a new child node labeled with \Fvier\
(\orelse\ a new child branch such that \Fvier\ \nolinebreak
 is the list of the conjugates of the formulas
 labeling the branch from the leaf to the new child node 
 of\/ \nolinebreak\math l).

\item[Instantiation:]
Let \math\sigma\ be a strong \exRsub.
Let \trip{C'}{R'}{<'} be the extended strong \math\sigma-update of\/ 
\trip C R <.
Then we may set
\bigmath{
  \forestzwei
  \nottight{:=}
  \foresteins\sigma
.} 
\end{description}
\end{theorem}

\pagebreak

\yestop
\yestop
\yestop
\noindent
While Hypothesizing and Instantiation steps are self-explanatory,
Expansion steps are parameterized by a sequent \Fzwei\
and a set of sequents \math G such that 
\math{\{\Fzwei\}} strongly \pair{R'}{C'}-reduces to \math G
for some extension \trip{C'}{R'}{<'} of \trip C R <.
For the below examples of \math\alpha-, \math\beta-, \math\gamma-, and
\math\delta-rules we will use the sequent calculi presentation
because it is a little more explicit.
When we write
\\\strongexpansionrule
\Fzwei{\Fdrei_0\quad\ldots\quad\Fdrei_{n-1}}{C''}{R''}{<''}{}

\noindent
we want to denote a sub-rule of the Expansion rule
which is given by \bigmath{G:=\{\Fdrei_0,\ldots,\Fdrei_{n-1}\},}
\bigmath{C':=C\cup C'',}
\bigmath{R':=R\cup R'',}
and
\bigmath{\tight{<'}:=\tight<\cup\tight{<''}.}
This means that for this rule really being a sub-rule of
the Expansion rule we have to show that
\math{C'} is a \pair{R'}{<'}-\cc\ and that
\math{\{\Fzwei\}} strongly \pair{R'}{C'}-reduces to \math G.

\yestop
\noindent
Let \math{A} and \math{B} be formulas,
\math{\Gamma} and \math{\Pi} sequents,
\math{x\in\Vbound}, 
\bigmath{
  \existsvari x{}
  \nottight\in
  \Vsome
  \setminus
  \VARsome{A,\Gamma\Pi}
,}\footnotemark\ 
and
\bigmath{
  \forallvari x{}
  \nottight\in
  \Vall
  \setminus
  \inparentheses{
    \VARall{A,\Gamma\Pi}
    \cup
    \DOM <
    \cup
    \DOM C
  }
.}

\noindent
\math\alpha-rules:
\\
\strongexpansionrule
{\Gamma~~\inparenthesesinlinetight{A\tightoder B}~~\Pi}
{A~~B~~\Gamma~~\Pi}
\emptyset\emptyset\emptyset{}
\\
\strongexpansionrule
{\Gamma~~\neg\inparenthesesinlinetight{A\tightund B}~~\Pi}
{\headroom\overline{\,A\,}~~\headroom\overline{\,B\,}~~\Gamma~~\Pi}
\emptyset\emptyset\emptyset{}
\\
\strongexpansionrule
{\Gamma~~\neg\neg A~~\Pi}
{A~~\Gamma~~\Pi}
\emptyset\emptyset\emptyset{}
\\

\noindent
\math\beta-rules:
\\
\strongexpansionrule
{\Gamma~~\inparenthesesinlinetight{A\tightund B}~~\Pi}
{A~~\Gamma~~\Pi~~~~~~~~~~~~B~~\Gamma~~\Pi}
\emptyset\emptyset\emptyset{}
\\
\strongexpansionrule
{\Gamma~~\neg\inparenthesesinlinetight{A\tightoder B}~~\Pi}
{\headroom\overline{\,A\,}~~\Gamma~~\Pi~~~~~~~~~~~~\headroom\overline{\,B\,}~~\Gamma~~\Pi}
\emptyset\emptyset\emptyset{}
\\

\noindent
\math\gamma-rules:
\\
\strongexpansionrule
{\Gamma~~\exists x\tight:\, A~~\Pi}
{A\{x\tight\mapsto\existsvari x{}\}~~\Gamma~~\exists x\tight:\, A~~\Pi}
\emptyset\emptyset\emptyset{}
\\
\strongexpansionrule
{\Gamma~~\neg\forall x\tight:\, A~~\Pi}
{\majorheadroom\overline{\,A\{x\tight\mapsto\existsvari x{}\}\,}~~\Gamma~~\neg\forall x\tight:\, A~~\Pi}
\emptyset\emptyset\emptyset{}
\\

\noindent
Liberalized \math\delta-rules:
\\
\strongexpansionrule
{\Gamma~~\forall x\tight:\, A~~\Pi}
{A\{x\tight\mapsto\forallvari x{}\}~~\Gamma~~\Pi}
{\{\pair{\forallvari x{}}{A\{x\tight\mapsto\forallvari x{}\}}\}}
{\inparenthesesinlinetight{
  \VARsome A
  \cup
  \revrelapp R{\VARall A}
 }
 \times
 \{\forallvari x{}\}
}
{\revrelapp{\tight\leq}{\VARall A}
 \times
 \{\forallvari x{}\}
}
{}
\\\\\\
\strongexpansionrule
{\Gamma~~\neg\exists x\tight:\, A~~\Pi}
{\majorheadroom\overline{\,A\{x\tight\mapsto\forallvari x{}\}\,}~~\Gamma~~\Pi}
{\{\pair{\forallvari x{}}{\headroom\overline{\,A\{x\tight\mapsto\forallvari x{}\}\,}}\}}
{\inparenthesesinlinetight{
  \VARsome A
  \cup
  \revrelapp R{\VARall A}
 }
 \times
 \{\forallvari x{}\}
}
{\revrelapp{\tight\leq}{\VARall A}
 \times
 \{\forallvari x{}\}
}
{}\\


\yestop
\begin{theorem}\label{theorem strong sub-rules}
\\
The above examples of 
\math\alpha-, \math\beta-, \math\gamma-, and
liberalized \math\delta-rules 
are all sub-rules of the Expansion rule of the sequent calculus
of \theoref{theorem strong abstract deductive calculi}.
\end{theorem}

\yestop
\yestop
\noindent
The following example shows that 
\math{R''} of the above liberalized \math\delta-rule must indeed contain 
\bigmath{
  \revrelapp R{\VARall A}
  \times
  \{\forallvari x{}\}
.}
\begin{example}\label{example R'' so big}
\\\LINEmath{
  \exists y\stopq
  \forall x\stopq
  \inparentheses{
    \neg\Qppp x y
    \oder
    \forall z\stopq
    \Qppp x z
  }
}

\noindent
is not deductively valid (to wit, let \Qpsymbol\ be the identity relation on
a non-trivial universe).
\\
\math\gamma-step:
\\\LINEmath{  
\forall x\stopq
  \inparentheses{
    \neg\Qppp x{\existsvari y{}}
    \oder
    \forall z\stopq
    \Qppp x z
  }
}
\\
Liberalized \math\delta-step:
\\\LINEmath{  
  \inparentheses{
    \neg\Qppp {\forallvari x{}}{\existsvari y{}}
    \oder
    \forall z\stopq
    \Qppp {\forallvari x{}} z
  }
}

\noindent
with \cc\ \pair{\forallvari x{}}{  \inparenthesesinlinetight{
    \neg\Qppp {\forallvari x{}}{\existsvari y{}}
    \oder
    \forall z\stopq
    \Qppp {\forallvari x{}} z
  }
} 
and 
\vc\ \pair{\existsvari y{}}{\forallvari x{}}.
\\
\math\alpha-step:
\\\LINEmath{  
    \neg\Qppp {\forallvari x{}}{\existsvari y{}}
    \comma
    \forall z\stopq
    \Qppp {\forallvari x{}} z
}
\\
Liberalized \math\delta-step:
\\\LINEmath{  
    \neg\Qppp {\forallvari x{}}{\existsvari y{}}
    \comma
    \Qppp {\forallvari x{}}{\forallvari z{}}
}

\noindent
with additional \cc\ 
\pair{\forallvari z{}}{\Qppp {\forallvari x{}}{\forallvari z{}}}
and additional \vc\ \pair{\existsvari y{}}{\forallvari z{}}.

Note that the additional \vc\ arises although \existsvari y{}
does not appear in \bigmath{\Qppp {\forallvari x{}} z.}
The reason for the additional \vc\ is 
\bigmath{\existsvari y{}\nottight R\forallvari x{}\in\VARall
{\Qppp {\forallvari x{}} z}.}

The \vc\ \pair{\existsvari y{}}{\forallvari z{}}
is, however, essential for soundness, because without it
we could complete the proof attempt by application of the 
strong existential 
\math{\{\pair{\existsvari y{}}{\forallvari x{}}\}}-substitution
\math{
  \sigma
  :=
  \{\existsvari y{}\tight\mapsto\forallvari z{}\}
  \uplus
  \domres\id{\Vsome\setminus\{\existsvari y{}\}}
}.

\end{example}

\pagebreak

\yestop
\yestop
\yestop
\noindent
Another interesting point is that now that we have achieved our
goal of liberalizing our \math\delta-rule and strictly increasing
our proving possibilities, we must not use our
original non-liberalized \math\delta-rule of \sectref{section weak}
anymore. This sounds quite strange on the first view,
but is simply due to our changed notion of reduction.
More precisely, 

\noindent
Weak \math\delta-rule:\strongexpansionrule
{\Gamma~~\forall x\tight:\, A~~\Pi}
{A\{x\tight\mapsto\forallvari x{}\}~~\Gamma~~\Pi}
\emptyset
{\inparenthesesinlinetight{
  \VARsome{A,\Gamma\Pi}
  \cup
  \revrelapp R{\VARall{A,\Gamma\Pi}}
 }
 \times
 \{\forallvari x{}\}
}
{\revrelapp{\tight\leq}{\VARall{A,\Gamma\Pi}}
 \times
 \{\forallvari x{}\}
}
{}

\noindent
does not describe a sub-rule of the 
Expansion rule of the sequent calculus
of \theoref{theorem strong abstract deductive calculi}.
To see this, let us start with the empty proof tree 
\quar\emptyset\emptyset\emptyset\emptyset\ and then 
hypothesize 
\bigmath{
  \forall x\stopq x\boldequal\zeropp
,}
which we abbreviate with \Feins.
Applying the above weak \math\delta-rule we get
\bigmath{\forallvari x{}\boldequal\zeropp}
as the label of the only leaf in the tree \math t
of the proof tree 
\quar{\pair\Feins t}\emptyset\emptyset\emptyset.
But, while \math{\{\Feins\}} does \math\emptyset-reduce to 
\math{\{\forallvari x{}\boldequal\zeropp\}}
(\ie\ \Open t), 
\math{\{\Feins\}} does not strongly \pair\emptyset\emptyset-reduce 
to \math{\{\forallvari x{}\boldequal\zeropp\}}.
To see this, consider some \math\Sigmaoffont-structure \nolinebreak\algebra\
with non-trivial universe, an arbitrary strong existential
\pair\algebra\emptyset-valuation \math e, and some
\math{\pi\in\FUNSET\Vall\algebra} with 
\bigmath{
  \pi\funarg{\forallvari x{}}\tightequal\zeropp^\algebra
.}
Then \math{\{\forallvari x{}\boldequal\zeropp\}} is
\trip\pi e\algebra-valid, but \math{\{\Feins\}} is not.
If we had applied the liberalized \math\delta-rule instead, 
we would have produced the proof tree
\quar{\pair\Feins t}C\emptyset\emptyset\ with
\bigmath{
  C
  \tightequal
  \{\pair{\forallvari x{}}{\forallvari x{}\boldequal\zeropp}\}
.}
And, indeed, 
\math\pi\ is not \pair e\algebra-compatible with \math C,
and
\math{\{\Feins\}} does strongly \pair\emptyset C-reduce
to \nolinebreak\math{\{\forallvari x{}\boldequal\zeropp\}}.

Note that there is a fundamental difference related to the
occurrence of the universal quantification on \math\pi\
between the notion of (weak) reduction

\noindent\LINEmath{\ldots\inparentheses
{\forall\pi\in\inparenthesesinlinetight{\FUNSET
\Vall\algebra}\stopq G_1\ \trip\pi e\algebra\mbox{-valid}}\implies
\inparentheses{\forall\pi\in\inparenthesesinlinetight{\FUNSET
\Vall\algebra}\stopq G_0\ \trip\pi e\algebra\mbox{-valid}}\ldots}

\noindent and the notion of strong reduction

\noindent\LINEmath{\ldots\forall\pi\in\inparenthesesinlinetight{\FUNSET
\Vall\algebra}\ldots\stopq\inparentheses
{G_1\ \trip\pi e\algebra\mbox{-valid}\implies G_0\ \trip\pi e\algebra
\mbox{-valid}}\ldots.}

\noindent
This difference in the nature of reduction
renders the weak version applicable in areas 
where the strong version is not.
For this reason (and for the sake of stepwise presentation)
we have included the weak version in this \paper\ although the strong version
will turn out to be superior in all aspects of the calculus of 
\theoref{theorem strong sub-rules} treated in this \paper.

This fundamental difference in the nature of reduction cannot be removed:
Suppose to weaken the notion of strong reduction in the following definition:
{\em \math{G_0} 
 quite-strongly \pair R C-reduces to 
 \nolinebreak\math{G_1} in \nolinebreak\algebra}\/
\udiff\
for each strong existential \pair\algebra R-valuation \math e:
  if   \math{G_1} is \stronglyvalid e\algebra C,
  then \math{G_0} is \stronglyvalid e\algebra C.
At first glance, this version seems to be very nice. One nice aspect is that
quite-strong \pair R\emptyset-reduction is so similar to 
(weak) \math R-reduction 
that we could omit the weak version because it would be very unlikely
to find an application of the weak version where the strong version 
would not be applicable.
Another nice aspect is that with quite-strong reduction
we could easily adapt our intended version of inductive
theorem proving as described in \sectref{section without skolemization},
which is not so easy with strong reduction because the 
induction hypotheses application becomes difficult.
But for the (really essential!)
monotonicity of reduction as given in \lemmref{lemma strong reduces to}(5),
quite-strong reduction produces the following two additional
requirements: 
\bigmath{\DOM{C'\tightsetminus C}\cap\VARall{G_1\cup\RAN C}=\emptyset}
and 
\bigmath{\VARsome{G_1}\times\DOM{C'\tightsetminus C}\subseteq R'.}
While the first requirement is unproblematic,
the second one restricts the \math\delta-rule even more,
which is the opposite of our intention behind the strong version,
namely to liberalize the \math\delta-rule.

\vfill
\pagebreak

\yestop\yestop
\yestop\yestop
\noindent
Moreover, note that 
(as far as \theoref{theorem strong sub-rules} is concerned) 
the \cc s
do not have any influence on our proofs and may be discarded.
We could, however, use them 
for the following purposes:
\begin{enumerate}

\yestop
\item
We could use the \cc s in order to weaken our requirements for our set
of axioms \Ax: Instead of 
\strongvalidityinA{\Vsome\tighttimes\Vall}\emptyset\ of \Ax\ the weaker 
\strongvalidityinA{\Vsome\tighttimes\Vall}C          of \Ax\
is sufficient for \theoref{theorem strong closed means valid}.

\yestop
\yestop
\item
If we add a functional behavior to a \cc\ \nolinebreak\math C,
\ie\ if we require that for \math{\pair{\forallvari x{}}A\in C}
the value for \forallvari x{} is not just an arbitrary one from
the set of values that make \math A invalid, but a unique element
of this set given by some choice-function,
then we can use the \cc s for simulating the behavior of the 
\math{\delta^{+^+}}-rule of \cite{deltaplusplus}
by using the same \fuv\ for the same \math C-value 
and by later equating \fuv s whose \math C-values become equal
during the proof.

\yestop
\yestop
\item
Moreover, the \cc s may be used to get more interesting answers:
\begin{example}

\noindent
Starting with the empty proof tree and hypothesizing

\noindent
\LINEmath{
  \forall x\stopq\Qppp x x\comma
  \exists y\stopq\inparentheses{\neg\Qppp y y\und\neg\Pppp y}\comma
  \Pppp{\existsvari z{}}
}

\noindent
with the above rules we can produce a proof tree with
the leaves 

\noindent
\LINEmath{
  \neg\Qppp{\existsvari y{}}{\existsvari y{}}
  \comma
  \Qppp{\forallvari x{}}{\forallvari x{}}
  \comma
  \exists y\stopq\inparentheses{\neg\Qppp y y\und\neg\Pppp y}\comma
  \Pppp{\existsvari z{}}
}
\\
and
\\
\LINEmath{
  \neg\Pppp{\existsvari y{}}
  \comma
  \Qppp{\forallvari x{}}{\forallvari x{}}
  \comma
  \exists y\stopq\inparentheses{\neg\Qppp y y\und\neg\Pppp y}\comma
  \Pppp{\existsvari z{}}
}

\noindent
and the \pair\emptyset\emptyset-\cc\ 
\math{
  \{
    \pair
      {\forallvari x{}}
      {\Qppp{\forallvari x{}}{\forallvari x{}}}
  \}
}.

The strong existential \math\emptyset-substitution
\ \bigmath{
  \{\existsvari y{}\tight\mapsto\forallvari x{}
    \comma
    \existsvari z{}\tight\mapsto\forallvari x{}
  \}
  \nottight{\nottight\uplus}
  \domres\id{\Vsome\setminus\{\existsvari y{},\existsvari z{}\}}
} \
closes the proof tree via an Instantiation step.
The answer \forallvari x{} for our query variable \existsvari z{}
is not very interesting unless we note that the \cc\ tells us to
choose \forallvari x{} in such a way that 
\Qppp{\forallvari x{}}{\forallvari x{}} becomes false.

The rules of our weak version of \sectref{section weak} 
are not only unable to provide any information on \fuv s,
but also unable to prove the hypothesized sequent, because
they can only show

\noindent
\LINEmath{
  \forall x\stopq
  \Qppp x x
  \comma
  \exists y\stopq
  \inparentheses{
     \neg\Qppp y y
     \und
     \neg\Pppp y
  }
  \comma
  \exists z\stopq\Pppp z
}
\\
instead. 
\end{example}

\yestop
\yestop
\noindent
Thus it is obvious that the calculus of \theoref{theorem strong sub-rules}
is not only superior to the calculus of \theoref{theorem sub-rules} \wrt\ 
proving but also \wrt\ answer ``computation''.

\end{enumerate}

\pagebreak

\yestop
\yestop
\yestop
\yestop
\noindent
Finally, note that 
(concerning the calculus of \theoref{theorem strong sub-rules}) 
the ordering \math < is not needed at all
when in the liberalized \math\delta-steps 
we always choose a completely new \fuv\ \forallvari x{}
that does not occur elsewhere
and when
in the Hypothesizing steps we guarantee that
\RAN{R''} contains only new \fuv s 
that have not occurred before. 
The former is reasonable anyhow,
because the \fuv s introduced by previous liberalized \math\delta-steps
cannot be used because they are in \DOM C and the use of a
\fuv\ from the input hypothesis deteriorates the result of 
our proof by giving this \fuv\ an existential meaning 
(because it puts it into \DOM C)
as 
explained in \theoref{theorem strong closed means valid}.
The latter does not seem to be restrictive for any
reasonable application.

\yestop
\yestop
\yestop
\noindent
All in all, when interested in proving only, 
the (compared to the weak version) additional
\cc\ and ordering of the strong version do not produce any
overhead because they can simply be omitted.
This is interesting because \cc s or \hilbert's \math\varepsilon-expressions
are sometimes considered to make proofs quite complicated. 
When interested in answer ``computation'',
however, they could turn out to be useful.

\yestop
\yestop
\yestop
\noindent
\Wrt\ the calculus of \theoref{theorem strong sub-rules}
we thus may conclude that the strong version is generally better
than the weak version and the only overhead seems to be that we have to
compute transitive closures when checking whether a substitution 
\math\sigma\ 
is really a strong existential \math R-substitution
and when computing the strong \math\sigma-update of \math R.
But we actually do not have to compute the transitive closure
at all, because the only essential thing is the circularity-check
which can be done on a bipartite\footnotemark\ graph generating the 
transitive closures.
This checking is in the worst case linear in 

\noindent
\LINEmath{
  \CARD R
  +  
  \displaystyle\sum_{\sigma}
  \inparentheses{
  \CARD{U_\sigma}
  +
  \CARD{E_\sigma}
  }
}

\noindent
and is expected to perform at least as well as
an optimally integrated version 
(\ie\ one without conversion of term-representation)
of the linear
unification algorithm of \cite{patersonwegman} 
in the standard framework of Skolemization and unification.
Note, however, that the checking for strong existential \math R-substitutions
can also be implemented with any other unification algorithm.

Not really computing the transitive closure enables another refinement
that allows us to go even beyond the fascinating\emph{strong Skolemization}
of \cite{strongskolem}. 
The basic idea of \cite{strongskolem} can be translated
into our framework in the following simplified way.

\begin{sloppypar}
Instead of proving 
\ \mbox{\math{
  \forall x\tight:\,
  \inparenthesesinlinetight{
    A\tightoder B
  }
}} \ 
it may be advantageous to prove the stronger 
\ \mbox{\math{
  \forall x\tight:\,
    A
  \oder
  \forall x\tight:\,
    B
},} \ 
because after applications of \math\alpha- and
liberalized \math\delta-rules to
\ \mbox{\math{
  \forall x\tight:\,
    A
  \oder
  \forall x\tight:\,
    B
},} \ 
resulting in
\ \mbox{\math{
    A\{x\tight\mapsto\forallvari x A\}
  \comma
    B\{x\tight\mapsto\forallvari x B\}
},} \ 
the \vc s introduced for \forallvari x A and \forallvari x B
may be smaller than the \vc\ introduced for \forallvari y{}
after applying these rules to
\bigmath{
  \forall x\tight:\,
  \inparenthesesinlinetight{
    A\tightoder B
  }
,}
resulting in
\bigmath{
    A\{x\tight\mapsto\forallvari y{}\}
  \comma
    B\{x\tight\mapsto\forallvari y{}\}
,}
\ie\ 
\revrelappsin R{\forallvari x A} 
and
\revrelappsin R{\forallvari x B} 
may be\emph{proper} subsets of \nolinebreak\revrelappsin R{\forallvari y{}}.
Therefore the proof of  
\bigmath{
  \forall x\tight:\,
    A
  \oder
  \forall x\tight:\,
    B
}
may be simpler than the proof of
\bigmath{
  \forall x\tight:\,
  \inparenthesesinlinetight{
    A\tightoder B
  }
.}
The nice aspect of \cite{strongskolem} is that the proofs of
\bigmath{
  \forall x\tight:\,
    A
}
and
\bigmath{
  \forall x\tight:\,
  \inparenthesesinlinetight{
    A\tightoder B
  }
}
can be done in parallel without extra costs,
such that the bigger \vc\ becomes active only if we decide that 
the smaller \vc\ is not enough to prove 
\bigmath{
  \forall x\tight:\,
    A
}
and we had better prove the weaker
\bigmath{
  \forall x\tight:\,
  \inparenthesesinlinetight{
    A\tightoder B
  }
.}

The disadvantage of the strong Skolemization approach of \cite{strongskolem}, 
however, is that we have to decide 
whether to prove either 
\bigmath{
  \forall x\tight:\,
    A
}
or else
\bigmath{
  \forall x\tight:\,
    B
}
in parallel to
\bigmath{
  \forall x\tight:\,
  \inparenthesesinlinetight{
    A\tightoder B
  }
.}
In terms of \hilbert's \math\varepsilon-operator,
this asymmetry can be understood from the argumentation of 
\cite{strongskolem}, which,
for some new variable \math{z\in\Vbound} and 
\math t denoting the term 
\bigmath{
  \varepsilon z\tight:\,
  \inparenthesesinlinetight{
    \neg A\{x\tight\mapsto z\}
    \und
    \inparenthesesinlinetight{A\oder x\boldequal z}
},}
employs the logical equivalence
of 
\ \mbox{\math{
  \forall x\tight:\,
  \inparenthesesinlinetight{
    A\tightoder B
  }
}} \ 
with
\ \mbox{\math{
  \forall x\tight:\,A
  \oder
  \forall x\tight:\,
  \inparenthesesinlinetight{
    B\{x\tight\mapsto t\}
  }
}} \ 
and then the logical equivalence of 
\ \mbox{\math{
  \forall x\tight:\,A
}} \ 
with
\ \mbox{\math{
  \exists x\tight:\,
  \inparenthesesinlinetight{
    A\{x\tight\mapsto t\}
  }
}.}

Now, if we do not really compute the transitive closures in our
strong version, we can prove 
\bigmath{
    A\{x\tight\mapsto\forallvari x A\}
  \comma
    B\{x\tight\mapsto\forallvari x B\}
}
in parallel and may later decide to prove the stronger
\bigmath{
    A\{x\tight\mapsto\forallvari y{}\}
  \comma
    B\{x\tight\mapsto\forallvari y{}\}
}
instead, 
simply by merging the nodes for \forallvari x A and \forallvari x B 
and substituting \forallvari x A and \forallvari x B with \forallvari y{}.
\end{sloppypar}

\vfill
\yestop\yestop\yestop\yestop

\section{Conclusion}
\label{section conclusion}

\yestop
\noindent
All in all, we have presented an easy to read combination 
of raising, explicit variable dependency representation, 
the liberalized \math\delta-rule, and
preservation of solutions
for first-order deductive theorem proving.
Our motivation was not only to make these subjects more popular, 
but also to provide the foundation for our work on
inductive theorem proving (\cf\ \cite{wirthtableaux}) 
where the preservation of solutions is indispensable.

To our knowledge\footnotemark\ we have presented on the one hand the 
first sound combination of explicit variable dependency representation
and the liberalized \math\delta-rule 
and on the other hand the first framework 
for preservation of solutions in full first-order logic.

Finally, 
the described problems with the development of the strong version
reveal unexpected details on the nature of the liberalized \math\delta-rule,
and the discussion at the end of \sectref{section strong} opens up
several new research directions.

\vfill\vfill\vfill\pagebreak
\section{The Proofs}
\label{section the proofs}

\begin{proofparsepqed}{\lemmref{lemma changing R}}
{\bf(1):}
Since \bigmath{e'} is a \opt{strong} existential 
\pair\algebra{R'}-valuation,
\ \math{S_{e'}\tight\circ R'}
\opt{\math{
  \nottight\circ
  \refltransclosureinline{\inparenthesesinlinetight{S_{e'}\tight\circ R'}}
}} \ 
is irreflexive \opt{and a wellfounded ordering}.
Since \bigmath{R\tightsubseteq R',}
we have 
\bigmath{
   {S_{e'}\tight\circ R}
    \ 
    \opt
     {\nottight\circ
      \refltransclosureinline
       {\inparenthesesinlinetight
         {S_{e'}\tight\circ R}}}
}
\math\subseteq
\bigmath{
  {S_{e'}\tight\circ R'}
    \ 
    \opt
     {\nottight\circ
      \refltransclosureinline
       {\inparenthesesinlinetight
         {S_{e'}\tight\circ R'}}}
.}
Thus 
\bigmath{
   {S_{e'}\tight\circ R}
    \ 
    \opt
     {\nottight\circ
      \refltransclosureinline
       {\inparenthesesinlinetight
         {S_{e'}\tight\circ R}}}
}
is irreflexive \opt{and a wellfounded ordering}, too.
Therefore, setting \math{e:=e'},
we get a \opt{strong} existential
\pair\algebra R-valuation trivially satisfying the requirements.

\noindent
{\bf(2):}
Here we denote concatenation (product) of 
relations `\math\circ' simply by juxtaposition and 
assume it to have higher priority than any other
binary operator.
Let \bigmath{e'} be some \opt{strong}
existential \pair\algebra{R'}-valuation.
Define \math{S_e:=S_{e'}E_\sigma\cup U_\sigma}
and the \opt{strong} existential 
\pair\algebra R-valuation
\bigmath e by 
(\math{x\tightin\Vsome}, 
 \math{
   \pi'\tightin
   \FUNSET
     {\revrelappsin{S_e}x}
     \algebra
 }):
\\\LINEmath{\headroom\footroom
  e
  \funarg x
  \funarg{\pi'}
  :=
  \EVAL{
    \algebra
    \uplus
    \epsilon
    \funarg{e'}
    \funarg\pi
    \uplus
    \pi}
  \funarg{\sigma\funarg x}
}
\\
where \math{\pi\in\FUNSET\Vall\algebra} is an arbitrary extension of
\math{\pi'}.
For this definition to be okay, we have to prove the following 
claims:
\\\underline
  {Claim~1:}
For \math{y\in\VARall{\sigma\funarg x}},
the choice of \math{\pi\supseteq\pi'} does not influence the
value of \math{\pi\funarg y}.
\\\underline
  {Claim~2:}
For \math{x'\in\VARsome{\sigma\funarg x}},
the choice of \math{\pi\supseteq\pi'} does not influence the
value of 
\math{
    \epsilon
    \funarg{e'}
    \funarg\pi
    \funarg{x'}
}.
\\\underline
  {Claim~3:}
For the weak version we have to show that
\math{S_e R} is irreflexive.
\\\underline
  {Claim~4:}
For the strong version we have to show that
\transclosureinline{\inparenthesesinlinetight{S_e R}} is 
a wellfounded ordering.
\\\underline
  {Proof of Claim~1:}
\bigmath{y\tightin\VARall{\sigma\funarg x}}
means
\bigmath{
  \pair y x
  \tightin
  U_\sigma
.}
By definition of \math{S_e} we have
\bigmath{
  \pair y x
  \tightin
  S_e
,}
\ie\
\bigmath{
  y\in
  \revrelappsin{S_e}x
  =
  \DOM{\pi'}
.}
\QED{Claim~1}
\\\underline
  {Proof of Claim~2:}
\bigmath{x'\tightin\VARsome{\sigma\funarg x}}
means
\bigmath{
  \pair{x'}x
  \tightin
  E_\sigma
.}
Thus by definition of \math{S_e}
we have 
\bigmath{
  S_{e'}
  \{\pair{x'}x\}
  \subseteq
  S_e
,}
\ie\
\bigmath{
  \revrelappsin{S_{e'}}{x'}
  \subseteq
  \revrelappsin{S_e}x
  =
  \DOM{\pi'}
.}
Therefore
\bigmath{
    \epsilon
    \funarg{e'}
    \funarg\pi
    \funarg{x'}
  =
    e'
    \funarg{x'}
    \funarg
      {\domres\pi
        {\revrelappsin{S_{e'}}{x'}}}
  =
    e'
    \funarg{x'}
    \funarg{\domres{\pi'}{\revrelappsin{S_{e'}}{x'}}}
.}  
\QED{Claim~2}
\\\underline
  {Proof of Claim~3:}
Since
\bigmath{
    S_e 
    R
  =
    S_{e'}
    E_\sigma 
    R
   \cup
    U_\sigma 
    R
}
and 
\math{U_\sigma R}
is irreflexive 
(as \math\sigma\ is an \exRsub),
it suffices to show irreflexivity of 
\math{
    S_{e'}
    E_\sigma 
    R
.}
Since 
\math{R'} is the \math\sigma-update of \math R,
this is equal to 
\math{
    S_{e'}
    R'
,}
which is irreflexive 
because \bigmath{e'} is an existential 
\pair\algebra{R'}-valuation.
\QED{Claim~3}
\\\underline
  {Proof of Claim~4:}
Since \math\sigma\ is a strong \exRsub,
\transclosureinline{\inparenthesesinlinetight{U_\sigma R}}
is a wellfounded ordering.
Thus, if 
\bigmath{
    \transclosureinline{\inparenthesesinlinetight{S_e R}}
  =
    \transclosureinline{\inparenthesesinlinetight{
         {S_{e'}E_\sigma R}
       \cup
         {U_\sigma R}}}
  =
    \transclosureinline{\inparenthesesinlinetight
      {U_\sigma R}}
    \cup
    \refltransclosureinline{\inparenthesesinlinetight
         {U_\sigma R}}
    \transclosureinline{\inparenthesesinlinetight
      {S_{e'}
       E_\sigma
       R
       \refltransclosureinline{\inparenthesesinlinetight
         {U_\sigma R}}}}
}
is not a wellfounded ordering,
there must be an infinite descending sequence of 
the form
\bigmath{
  y_{2i+2}
  \nottight
    {\refltransclosureinline{\inparenthesesinlinetight{U_\sigma R}}}
  y_{2i+1}
  \nottight
    {\transclosureinline{\inparenthesesinlinetight
      {S_{e'}
       E_\sigma
       R
       \refltransclosureinline{\inparenthesesinlinetight
         {U_\sigma R}}}}}
  y_{2i}
}
for all \math{i\in\N}.
But then
\bigmath{
  y_{2i+3}
  \nottight
    {\transclosureinline{\inparenthesesinlinetight
      {S_{e'}
       E_\sigma
       R
       \refltransclosureinline{\inparenthesesinlinetight
         {U_\sigma R}}}}}
  y_{2i+2}
  \nottight
    {\refltransclosureinline{\inparenthesesinlinetight{U_\sigma R}}}
  y_{2i+1}
,}
which contradicts the wellfoundedness of
\bigmath{
    \transclosureinline{\inparenthesesinlinetight
      {S_{e'}
       E_\sigma
       R
       \refltransclosureinline{\inparenthesesinlinetight
         {U_\sigma R}}}}
    \refltransclosureinline{\inparenthesesinlinetight{U_\sigma R}}
  =
    \transclosureinline{\inparenthesesinlinetight
      {S_{e'}
       E_\sigma
       R
       \refltransclosureinline{\inparenthesesinlinetight
         {U_\sigma R}}}}
  =
    \transclosureinline{\inparenthesesinlinetight
      {S_{e'}R'}}
,}
where the latter step is due to \math{R'} being the
strong \math\sigma-update of \math R.
The latter relation is a wellfounded ordering, however, 
because \bigmath{e'} is a strong existential 
\pair\algebra{R'}-valuation.\QED{Claim~4}

\noindent
Now, for
\bigmath{
  \pi\in\FUNSET\Vall\algebra
}
and 
\bigmath{
  x\in\Vsome
}
we have
\\\LINEmath{
  \epsilon
  \funarg e
  \funarg\pi
  \funarg x
  =
  e
  \funarg x
  \funarg{\domres\pi{\revrelappsin{S_e}x}}
  =
  \EVAL{
    \algebra
    \uplus
    \epsilon
    \funarg{e'}
    \funarg\pi
    \uplus
    \pi
  }
  \funarg{\sigma\funarg x}
}
\\  
\ie
\LINEmath{
  \epsilon
  \funarg e
  \funarg\pi
  =
  \sigma
  \circ
  \EVAL{
    \algebra
    \uplus
    \epsilon
    \funarg{e'}
    \funarg\pi
    \uplus
    \pi
  }
.}
\end{proofparsepqed}

\yestop
\begin{proofqed}{\lemmref{lemma reduces to 5a}}\\
This a trivial consequence of \lemmref{lemma changing R}(1).
\end{proofqed}\pagebreak

\begin{proofqed}{\lemmref{lemma reduces to}}
(1), (2), (3), and (4) are trivial.
Note that (5) is a trivial consequence of \lemmref{lemma changing R}(1).

\noindent
\underline{(6a):}
Suppose that \math{G_0\sigma} is \math{R'}-valid in 
\nolinebreak\algebra.
Then there is some existential \pair\algebra{R'}-valuation \math{e'}
such that \math{G_0\sigma} is \pair{e'}\algebra-valid.
Then, by \lemmref{lemma changing R}(2), 
there is some existential \pair\algebra{R}-valuation \math{e}
such that for all \math{\pi\in\FUNSET\Vall\algebra}:
\bigmath{
  \epsilon
  \funarg e
  \funarg\pi
  =
  \sigma
  \circ
  \EVAL{
    \algebra
    \uplus
    \epsilon
    \funarg{e'}
    \funarg\pi
    \uplus
    \pi
  }
.}
Moreover, for \math{y\in\Vall} we have:
\bigmath{
  \pi
  \funarg y
  =
  \EVAL{
    \algebra
    \uplus
    \epsilon
    \funarg{e'}
    \funarg\pi
    \uplus
    \pi
  }
  \funarg{y}
,}
\\  
\ie
\LINEmath{
      \epsilon
      \funarg e
      \funarg\pi
      \uplus
      \pi
  =
        \inparenthesesinlinetight{
          \sigma
          \uplus
          \domres\id\Vall
        }
        \circ
        \EVAL{
          \algebra
          \uplus
          \epsilon
          \funarg{e'}
          \funarg\pi
          \uplus
          \pi
        }
.}
\\
Thus, for any formula \math B, we have
\\\LINEmath{
  \begin{array}{l l}
    \EVAL{
      \algebra
      \uplus
      \epsilon
      \funarg e
      \funarg\pi
      \uplus
      \pi
    }
    \funarg B
  &=\\
    \EVAL{
      \algebra
      \uplus
      \inparenthesesinlinetight{
        \inparenthesesinlinetight{
          \sigma
          \uplus
          \domres\id\Vall
        }
        \circ
        \EVAL{
          \algebra
          \uplus
          \epsilon
          \funarg{e'}
          \funarg\pi
          \uplus
          \pi
        }
      }
    }
    \funarg B
  &=\\
    \EVAL{
      \algebra
      \uplus
      \epsilon
      \funarg{e'}
      \funarg\pi
      \uplus
      \pi
    }
    \funarg{B\sigma}
    ,
  \end{array}
}
\\
the latter step being due to the Substitution-Lemma.

\noindent
Thus,
for any set of sequents \math{G'}: 

\noindent
\LINEnomath{
  \pair e\algebra-validity of
  \math{G'}
  is logically equivalent to 
  \pair{e'}\algebra-validity of
  \math{G'\sigma}
.}(:\S)

\noindent
Especially, \math{G_0} is \pair e\algebra-valid.
Thus, \math{G_0} is \math R-valid in 
\nolinebreak\algebra.

\noindent
\underline{(6b):}
Let \math{e'} be some existential \pair\algebra{R'}-valuation
and suppose that \math{G_1\sigma} is \pair{e'}\algebra-valid.
Let \math{e} be the existential \pair\algebra{R}-valuation 
given by \lemmref{lemma changing R}(2).
Then, by (\S) in the proof of (6a),
\math{G_1} is \pair{e}\algebra-valid.
By assumption, 
\math{G_0} \nolinebreak \math R-reduces to \nolinebreak\math{G_1}.
Thus, \math{G_0} is \pair e\algebra-valid.
By (\S) in the proof of (6a), this means that
\math{G_0\sigma} is \pair{e'}\algebra-valid.
\end{proofqed}

\begin{proofqed}{\theoref{theorem closed means valid}}\\
Since \Ax\ is \math{\Vsome\tighttimes\Vall}-valid, 
\math t is closed,
and \bigmath{R\subseteq\Vsome\tighttimes\Vall,}
by \lemmref{lemma reduces to}(5), 
\Open t is \math R-valid.
Since 
\bigmath{(\Feins,t)\tightin\foresteins}
and 
\pair\foresteins R satisfies the invariant condition,
\math{\{\Feins\}} \math R-reduces to \Open t.
All in all, by \lemmref{lemma reduces to}(1),
\math{\Feins} is \math R-valid.
\end{proofqed}

\begin{proofqed}{\theoref{theorem abstract deductive calculi}}\\
\pair\emptyset\emptyset\ trivially satisfies
the invariant condition. For the iteration
steps, let \math{\pair\Feinsprimeprime{t''}\in\forestzwei}.
Assuming the invariant condition for 
\pair\foresteins R, we have to show that
\math{\{\Feinsprimeprime\}} \math{R'}-reduces to \Open{t''}.

\noindent
\underline{Hypothesizing:}
In case of \bigmath{\pair\Feinsprimeprime{t''}\tightin\foresteins,}
\math{\{\Feinsprimeprime\}} \math R-reduces to \Open{t''} by
assumption, and then, due to \bigmath{R\tightsubseteq R'}
and \lemmref{lemma reduces to}(5),
\math{\{\Feinsprimeprime\}} \math{R'}-reduces to \Open{t''}.
Otherwise we have 
\bigmath{
  \pair\Feinsprimeprime{t''}
  \tightequal
  \pair\Feins t
.}
Then
\bigmath{
  \{\Feinsprimeprime\}
  =
  \{\Feins\}
  =
  \Open t
  =
  \Open{t''}
.}
Thus, by \lemmref{lemma reduces to}(2),
\math{\{\Feinsprimeprime\}} \math{R'}-reduces to \Open{t''}.

\noindent
\underline{Expansion:}
In case of \bigmath{\pair\Feinsprimeprime{t''}\tightin\foresteins,}
\math{\{\Feinsprimeprime\}} \math R-reduces to \Open{t''} by
assumption, and then, due to \bigmath{R\tightsubseteq R'}
and \lemmref{lemma reduces to}(5),
\math{\{\Feinsprimeprime\}} \math{R'}-reduces to \Open{t''}.
Otherwise we have 
\bigmath{
  \pair\Feinsprimeprime{t''}
  \tightequal
  \pair\Feins{t'}
.}
Since \bigmath{\Open{t}\tightsetminus\{\Fzwei\}\subseteq\Open{t'},}
by \lemmref{lemma reduces to}(2),
\math{\Open{t}\tightsetminus\{\Fzwei\}} \math{R'}-reduces to \Open{t'}.
Thus, since by assumption
\math{\{\Fzwei\}} \math{R'}-reduces to a subset of \Open{t'},
by \lemmref{lemma reduces to}(4)
\Open{t} \math{R'}-reduces to \Open{t'}.
Moreover, due to 
\bigmath{
  \pair\Feins t\tightin\foresteins
,}
by assumption
\math{\{\Feins\}} \nolinebreak \math R-reduces to \Open t.
Thus, by \bigmath{R\tightsubseteq R'} and 
\lemmref{lemma reduces to}(5),
\math{\{\Feins\}} \math{R'}-reduces to \Open t.
Thus, since 
\Open{t} \math{R'}-reduces to \Open{t'},
by \lemmref{lemma reduces to}(3)
\math{\{\Feins\}} \math{R'}-reduces to \Open{t'},
\ie\
\math{\{\Feinsprimeprime\}} \math{R'}-reduces to \Open{t''}.

\noindent
\underline{Instantiation:}
There is some \math{\pair\Feins t\in\foresteins} such that
\bigmath{
  \pair\Feins t\sigma
  =
  \pair\Feinsprimeprime{t''}
.}
By assumption, 
\math{\{\Feins\}} \math R-reduces to \Open t.
By \lemmref{lemma reduces to}(6),
\math{\{\Feins\sigma\}} \math{R'}-reduces to \math{\Open t\sigma},
\ \ie\
\math{\{\Feinsprimeprime\}} \math{R'}-reduces to \Open{t''}.%
\end{proofqed}

\pagebreak

\begin{proofparsepqed}{\theoref{theorem sub-rules}}
Let \algebra\ be an arbitrary \math\Sigmaoffont-structure
(\math\Sigmaoffont-algebra).
We only prove the first example of each kind of rule to be a sub-rule
of the Expansion rule and leave the rest as an exercise.

\noindent
\underline{\math\alpha-rule:}
We have to show that 
\bigmath{
  \{\Gamma~~\inparenthesesinlinetight{A\tightoder B}~~\Pi\}  
}
\math\emptyset-reduces to 
\bigmath{
  \{A~~B~~\Gamma~~\Pi\}
}
in \nolinebreak\algebra.
This is trivial, however, because \pair e\algebra-validity 
of the two sets is logically equivalent for each existential
\pair\algebra\emptyset-valuation \math e.

\noindent
\underline{\math\beta-rule:}
We have to show that 
\bigmath{
  \{\Gamma~~\inparenthesesinlinetight{A\tightund B}~~\Pi\}
}
\math\emptyset-reduces to 
\bigmath{
  \{A~~\Gamma~~\Pi,~~~~~~B~~\Gamma~~\Pi\}
}
in \nolinebreak\algebra.
This is trivial, however, because \pair e\algebra-validity 
of the two sets is logically equivalent for each existential
\pair\algebra\emptyset-valuation \math e.

\noindent
\underline{\math\gamma-rule:}
We have to show that 
\bigmath{
  \{\Gamma~~\exists x\tight:\, A~~\Pi\}
}
\math\emptyset-reduces to 
\bigmath{
  \{A\{x\tight\mapsto\existsvari x{}\}~~\Gamma~~\exists x\tight:\, A~~\Pi\}
}
in \nolinebreak\algebra.
This is the case, however, because \pair e\algebra-validity 
of the two sets is logically equivalent for each existential
\pair\algebra\emptyset-valuation \math e.
The direction from left to right is given because the former
sequent is a sub-sequent of the latter. The other direction,
which is the only one we actually have to show here, is also
clear because \trip\pi e\algebra-validity of 
\bigmath{
  A\{x\tight\mapsto\existsvari x{}\}
}
implies
\trip\pi e\algebra-validity of 
\bigmath{
  \exists x\tight:\, A
.}
Although this is clear, we should be a little more explicit here
because the standard semantic definition of \math\exists\
(\cf\ \eg\ \cite{wirthdiss}, \p~188) 
does not use \fev s 
and is somewhat more complicated than it could be
in terms of \fev s. Moreover, in the note above the theorem 
we remarked that 
the restriction on \existsvari x{} not occurring in the former
sequent is not really necessary. Thus, in order to be more explicit
here, assume that
the latter sequent is \pair e\algebra-valid for some
existential \pair\algebra\emptyset-valuation \math e.
Let \math{\pi\in\FUNSET\Vall\algebra}.
We have to show that the former sequent is \trip\pi e\algebra-valid.
If this is not the case,
\bigmath{
  A\{x\tight\mapsto\existsvari x{}\}
}
must be \trip\pi e\algebra-valid.
Let \math{\forallvari y{}\in\Vall\tightsetminus\VARall A}.
Then,
since
\bigmath{
  A
  \{x\tight\mapsto\forallvari y{}\}
  \{\forallvari y{}\tight\mapsto\existsvari x{}\}
}
is equal to 
\bigmath{
  A\{x\tight\mapsto\existsvari x{}\}
,}
we know that
\bigmath{
  A
  \{x\tight\mapsto\forallvari y{}\}
  \{\forallvari y{}\tight\mapsto\existsvari x{}\}
}
is valid in
\bigmath{\algebra\uplus\epsilon\funarg e\funarg\pi\uplus\pi.}
Then, by the Substitution-Lemma,
\bigmath{
  A
  \{x\tight\mapsto\forallvari y{}\}
}
is valid in
\bigmath{\algebra\uplus\epsilon\funarg e\funarg\pi\uplus\pi'}
for \math{\pi'\in\FUNSET\Vall\algebra} given by
\bigmath{
  \domres{\pi'}{\Vall\setminus\{\forallvari y{}\}}
  :=
  \domres\pi{\Vall\setminus\{\forallvari y{}\}}
}
and
\bigmath{
  \pi'\funarg{\forallvari y{}}
  :=
  \epsilon\funarg e\funarg\pi\funarg{\existsvari x{}}
.}
By the standard semantic definition of \math\exists\
and since quantification on \math x cannot occur
in \math A (as \math{\exists x\tight:\,A} is a formula
in our restricted sense, \cf\ \sectref{section quantification restriction}),
this means that 
\math{
  \exists x\tight:
  \inparenthesesinlinetight{ 
  A
  \{x\tight\mapsto\forallvari y{}\}
  \{\forallvari y{}\tight\mapsto x\}
  }
}
is valid in \math{\algebra\uplus\epsilon\funarg e\funarg\pi\uplus\pi}.
Since \forallvari y{} does not occur in \math A,
this formula is equal to \math{\exists x\tight: A},
which means that the former sequent is \trip\pi e\algebra-valid
as was to be shown.

\noindent
\underline{\math\delta-rule:}
We have to show that 
\bigmath{
  \{\Gamma~~\forall x\tight:\, A~~\Pi\}
}
\math{R''}-reduces to 
\bigmath{
  \{A\{x\tight\mapsto\forallvari x{}\}~~\Gamma~~\Pi\}
}
in \nolinebreak\algebra\ for 
\bigmath{
  R''
  =
  {\VARsome{A,\Gamma\Pi}\times\{\forallvari x{}\}}
.}
Assume that 
the latter sequent is \pair e\algebra-valid for some
existential \math{R''}-valuation \math e.
Let \math{\pi\in\FUNSET\Vall\algebra}.
We have to show that the former sequent is \trip\pi e\algebra-valid.
If some formula in \math{\Gamma\Pi} is 
\trip\pi e\algebra-valid, 
then the former sequent is 
\trip\pi e\algebra-valid, too.
Otherwise, \math{\Gamma\Pi} is not only invalid
in \bigmath{\algebra\uplus\epsilon\funarg e\funarg\pi\uplus\pi,}
but also in 
\bigmath{\algebra\uplus\epsilon\funarg e\funarg\pi\uplus\pi'}
for all \math{\pi'\in\FUNSET\Vall\algebra} with 
\bigmath{
  \domres{\pi'}{\Vall\setminus\{\forallvari x{}\}}
  =
  \domres\pi{\Vall\setminus\{\forallvari x{}\}}
,}
simply because \forallvari x{}
does not occur in \math{\Gamma\Pi}.
Because of 
\bigmath{
  {\VARsome{\Gamma\Pi}\times\{\forallvari x{}\}}
  \subseteq
  R''
,}
we know that \math{\Gamma\Pi} must be even invalid in
\bigmath{
  \algebra\uplus\epsilon\funarg e\funarg{\pi'}\uplus\pi'
.}
Since the latter sequent is assumed to be \pair e\algebra-valid,
this means that \math{A\{x\tight\mapsto\forallvari x{}\}}
is \trip{\pi'} e\algebra-valid.
Because of 
\bigmath{  
  \VARsome{
    A
    \{x\tight\mapsto\forallvari x{}
    \}
  }
  \times
    \{\forallvari x{}
    \}
  =
  \VARsome A
  \times
    \{\forallvari x{}
    \}
  \subseteq
  R''
,}
we know that \math{A\{x\tight\mapsto\forallvari x{}\}} must be even valid in
\bigmath{\algebra\uplus\epsilon\funarg e\funarg\pi\uplus\pi'}
for all \math{\pi'\in\FUNSET\Vall\algebra} with 
\bigmath{
  \domres{\pi'}{\Vall\setminus\{\forallvari x{}\}}
  =
  \domres\pi{\Vall\setminus\{\forallvari x{}\}}
.}
By the standard semantic definition of \math\forall\
(\cf\ \eg\ \cite{wirthdiss}, \p~188) 
and since quantification on \math x cannot occur
in \math A (as \math{\forall x\tight:\,A} is a formula
in our restricted sense, \cf\ \sectref{section quantification restriction}),
this means that 
\math{
  \forall x\tight:
  \inparenthesesinlinetight{ 
  A
  \{x\tight\mapsto\forallvari x{}\}
  \{\forallvari x{}\tight\mapsto x\}
  }
}
is valid in \math{\algebra\uplus\epsilon\funarg e\funarg\pi\uplus\pi}.
Since \forallvari x{} does not occur in \math A,
this formula is equal to \math{\forall x\tight: A},
which means that the former sequent is \trip\pi e\algebra-valid
as was to be shown.
\end{proofparsepqed}\pagebreak

\begin{proofparsepqed}{\lemmref{lemma geserthmimproved}}
Since in \cite{geserthmimproved} Theorem~62 and especially its proof 
(which is used to illustrate the application of the 
 very special framework of 
 that paper)
are not easy to read, we give an easier proof here that
requires fewer set theoretical preconditions 
and uses induction only on 
\math\omega.
It proceeds 
 by showing the existence of a refutational element 
 in a nonempty set of infinite descending sequences.

Set \bigmath{F:=\DOM A\cup\RAN A\cup\DOM B\cup\RAN B.}
We show that
\bigmath{
  C
  :=
  \setwith
    {\FUNDEF t\N F}
    {\forall i\tightin\N\stopq 
     t_i\nottight{\inparenthesesinlinetight{A\tightcup B}}t_{i+1}}
}
is empty.
Otherwise we can choose \math{s\in C} and families 
\math{\inparenthesesinlinetight{D_i}_{i\in\N}} 
and
\math{\inparenthesesinlinetight{E_i}_{i\in\posN}} 
of subsets of \math F 
inductively in the following way:
\\
\bigmath{D_0:=\setwith{t_0}{t\tightin C}.}
Choose \math{s_0} such that it is \math B-irreducible in 
\math{D_0}, \ie\ such that \bigmath{s_0\tightin D_0} and 
there is no \math{t'\in D_0} such that
\bigmath{s_0\nottight B t'.}
\\
For \math{n\in\posN}:
\bigmath{
  D_n
  :=
  \setwith
    {t_n}
    {t\tightin C
     \nottight\und
     \forall i\tightprec n\stopq t_i\tightequal s_i
     \nottight\und
     s_{n-1}\nottight A t_n}
.}
\bigmath{
  E_n
  :=
  \setwith
    {t_n}
    {t\tightin C
     \nottight\und
     \forall i\tightprec n\stopq t_i\tightequal s_i
     \nottight\und
     s_{n-1}\nottight B t_n}
.}
If \math{E_n} is nonempty  we choose \math{s_n} from \math{E_n}.
Otherwise, 
we choose \math{s_n} to be \math B-irreducible in \math{D_n}.

\noindent
Since \bigmath{s\in C} and \math A is terminating, there is some minimal
\math{n\in\N} with \bigmath{s_n\nottight B s_{n+1}.}
We have \bigmath{n\tightsucc 0,}
because otherwise \bigmath{s_0\nottight B s_1\in D_0}
contradicts the choice of \math{s_0}.
Thus, 
\bigmath{
  s_{n-1}
  \nottight{\inparenthesesinlinetight{A\tightsetminus B}}
  s_n
  \nottight B
  s_{n+1}
.}
Since 
\bigmath{
  s_{n-1}
  \nottight{\inparenthesesinlinetight{A\tightsetminus B}}
  s_n
,}
we know that \math{s_n} was chosen not from \math{E_n},
but \math B-irreducible in \math{D_n}.
Due to 
\bigmath{
  A\tight\circ B
  \nottight{\nottight\subseteq}
  A
  \nottight{\nottight\cup}
  B 
  \tight\circ
  \refltransclosureinline{\inparenthesesinlinetight{A\cup B}}
}
we get two possible cases now.

\noindent
\underline{\math{s_{n-1}\nottight A s_{n+1}}:}
Then \math{s_0\ldots s_{n-1}s_{n+1}s_{n+2}\ldots} is an element of \math C.
Thus, \bigmath{s_{n+1}\tightin D_n.}
Due to \bigmath{s_n\nottight B s_{n+1},}
this contradicts \math{s_n} being \math B-irreducible in \math{D_n}.

\noindent
\underline
{\math{
  s_{n-1}
  \nottight
    {\inparenthesesinlinetight{
     B 
     \tight\circ
     \refltransclosureinline{\inparenthesesinlinetight{A\cup B}}
    }}
  s_{n+1}
}:}
Then there are some \math{m\in\N} and some
\\
\math{s_0\ldots s_{n-1}u_0\ldots u_m s_{n+2}s_{n+3}\ldots}
in \math C with 
\bigmath{s_{n-1}\nottight B u_0}
and \bigmath{u_m\tightequal s_{n+1}}.
Thus, \bigmath{u_0\tightin E_n,}
\ie\ \math{E_n} is not empty.
But this contradicts the fact that \math{s_n} was not chosen from \math{E_n}.
\end{proofparsepqed}

\vfill\begin{proofparsepqed}{\lemmref{lemm ex str s up}}
Here we denote concatenation (product) of 
relations `\math\circ' simply by juxtaposition and 
assume it to have higher priority than any other
binary operator.

\noindent
\underline{Claim~1:} 
\bigmath{
  R'\tight{<'}\nottight{\nottight\subseteq} R'
.}
\\\underline{Proof of Claim~1:}
Since \math C is a \pair R <-\cc, we have
\bigmath{
  R\tight{<}\subseteq R
.}
Thus,
\bigmath{
 R'\tight{<'}=
 E_\sigma R\refltransclosureinline{\inparenthesesinlinetight{U_\sigma R}}
 \inparenthesesinlinetight{\tight<
    \refltransclosureinline{\inparenthesesinlinetight{U_\sigma R}}
    \cup\transclosureinline{\inparenthesesinlinetight{U_\sigma R}}}
 =E_\sigma\refltransclosureinline{\inparenthesesinlinetight{R U_\sigma}}R
 \tight<\refltransclosureinline{\inparenthesesinlinetight{U_\sigma R}}
 \cup E_\sigma R\refltransclosureinline{\inparenthesesinlinetight{U_\sigma R}}
 \transclosureinline{\inparenthesesinlinetight{U_\sigma R}}\subseteq
 E_\sigma\refltransclosureinline{\inparenthesesinlinetight{R U_\sigma}}R
 \refltransclosureinline{\inparenthesesinlinetight{U_\sigma R}}\cup
 E_\sigma R\refltransclosureinline{\inparenthesesinlinetight{U_\sigma R}}
 \transclosureinline{\inparenthesesinlinetight{U_\sigma R}}=
 E_\sigma R\refltransclosureinline{\inparenthesesinlinetight{U_\sigma R}}
 \refltransclosureinline{\inparenthesesinlinetight{U_\sigma R}}
 \cup E_\sigma R\transclosureinline{\inparenthesesinlinetight{U_\sigma R}}
 =E_\sigma R\refltransclosureinline{\inparenthesesinlinetight{U_\sigma R}}=R'
.}\QED{Claim~1}

\noindent
\underline{Claim~2:} 
\math{<'} is a wellfounded ordering on \Vall.
\\\underline{Proof of Claim~2:}
Since \math C is a \pair R <-\cc, we know that \math < is a wellfounded
ordering on \Vall\ and
\bigmath{
  R\tight{<}\nottight{\nottight\subseteq} R
.}
\\
Thus
\LINEmath{
  U_\sigma R\tight<
  \nottight{\nottight\subseteq}
  U_\sigma R
,}\\
\LINEmath{
  \transclosureinline{\inparenthesesinlinetight{U_\sigma R}}\tight<
  \nottight{\nottight=}
  \refltransclosureinline{\inparenthesesinlinetight{U_\sigma R}}
  U_\sigma R\tight<
  \nottight{\nottight\subseteq}
  \refltransclosureinline{\inparenthesesinlinetight{U_\sigma R}}
  U_\sigma R
  \nottight{\nottight=}
  \transclosureinline{\inparenthesesinlinetight{U_\sigma R}}
,}\\
and
\LINEmath{
  \tight<\refltransclosureinline{\inparenthesesinlinetight{U_\sigma R}}\tight<
  \nottight{\nottight=}
  \tight<\tight<
  \nottight\cup
  \tight<\transclosureinline{\inparenthesesinlinetight{U_\sigma R}}\tight<
  \nottight{\nottight\subseteq}
  \tight<
  \nottight\cup
  \tight<\transclosureinline{\inparenthesesinlinetight{U_\sigma R}}
  \nottight{\nottight=}
  \tight<\refltransclosureinline{\inparenthesesinlinetight{U_\sigma R}}
.}
\\
Since \math\sigma\ is a strong existential 
\math R-substitution, we know that 
\transclosureinline{\inparenthesesinlinetight{U_\sigma R}}
is a is a wellfounded ordering on \Vall.
By \lemmref{lemma geserthmimproved}
(setting \math{A:=\reverserelation{\tight<}}
 and \math{B:=\reverserelation{\inparenthesesinlinetight{U_\sigma R}}})
by the first of the above containments, we know that
\bigmath{
  \reverserelation{\tight<}
  \cup
  \reverserelation{\inparenthesesinlinetight{U_\sigma R}}
} 
is terminating, which (due to
\ \math{
  \tight{<'}
  \nottight{\nottight=}
  \tight<
  \refltransclosureinline{\inparenthesesinlinetight{U_\sigma R}}
  \cup
  \transclosureinline{\inparenthesesinlinetight{U_\sigma R}}
}~)
means that
\tight{>'} is terminating, too.
Finally \math{<'} is also transitive, since by the above containments:
\\\mbox{}~~~~~~~~~~~~~~~~~\LINEmath{
  \tight<\refltransclosureinline{\inparenthesesinlinetight{U_\sigma R}}
  \tight<\refltransclosureinline{\inparenthesesinlinetight{U_\sigma R}}
  \nottight{\nottight\subseteq}
  \tight<\refltransclosureinline{\inparenthesesinlinetight{U_\sigma R}}
  \refltransclosureinline{\inparenthesesinlinetight{U_\sigma R}}
  \nottight{\nottight=}
  \tight<\refltransclosureinline{\inparenthesesinlinetight{U_\sigma R}}
  \nottight{\nottight\subseteq}
  \tight{<'}
}
\\
and
\LINEmath{
  \tight<
  \refltransclosureinline{\inparenthesesinlinetight{U_\sigma R}}
  \transclosureinline{\inparenthesesinlinetight{U_\sigma R}}
  \nottight{\nottight=}
  \tight<
  \transclosureinline{\inparenthesesinlinetight{U_\sigma R}}
  \nottight{\nottight\subseteq}
  \tight<
  \refltransclosureinline{\inparenthesesinlinetight{U_\sigma R}}
  \nottight{\nottight\subseteq}
  \tight{<'}
}
\\
and
\LINEmath{
  \transclosureinline{\inparenthesesinlinetight{U_\sigma R}}
  \tight<
  \refltransclosureinline{\inparenthesesinlinetight{U_\sigma R}}
  \nottight{\nottight\subseteq}
  \transclosureinline{\inparenthesesinlinetight{U_\sigma R}}
  \refltransclosureinline{\inparenthesesinlinetight{U_\sigma R}}
  \nottight{\nottight=}
  \transclosureinline{\inparenthesesinlinetight{U_\sigma R}}
  \nottight{\nottight\subseteq}
  \tight{<'}
}
\\
and ~~~~~~~~~~~
\LINEmath{
  \transclosureinline{\inparenthesesinlinetight{U_\sigma R}}
  \transclosureinline{\inparenthesesinlinetight{U_\sigma R}}
  \nottight{\nottight\subseteq}
  \transclosureinline{\inparenthesesinlinetight{U_\sigma R}}
  \nottight{\nottight\subseteq}
  \tight{<'}
.}\QED{Claim~2}

\noindent
\underline{Claim~3:}
For all \math{\forallvari y{}\in\DOM{C'}}:
For all 
\math{
  \forallvari z{}
  \in
  \VARall{C'\funarg{\forallvari y{}}}\tightsetminus\{\forallvari y{}\}
}:
\bigmath{
  \forallvari z{}<'\forallvari y{}
.}
\\\underline{Proof of Claim~3:}
Let 
\math{
  \forallvari z{}
  \in
  \VARall{C'\funarg{\forallvari y{}}}\tightsetminus\{\forallvari y{}\}
}.
By the definition of \math{C'} this means
\bigmath{
  \forallvari z{}
  \tightin
  \VARall{C\funarg{\forallvari y{}}}\tightsetminus\{\forallvari y{}\}
}
or there is some \math{\existsvari u{}\in\VARsome{C\funarg{\forallvari y{}}}}
with \bigmath{\forallvari z{}\nottight{U_\sigma}\existsvari u{}.}
Since \math C is a \pair R <-\cc, we have
\bigmath{
  \forallvari z{}<\forallvari y{}
}
or
\bigmath{
  \forallvari z{}\nottight{U_\sigma}\existsvari u{}\nottight R\forallvari y{}
.}
Thus, by definition of \math{<'} we have
\bigmath{
  \forallvari z{}<'\forallvari y{}
.}\QED{Claim~3}

\noindent\underline{Claim~4:}
For all \math{\forallvari y{}\in\DOM{C'}}:
For all \math{\existsvari u{}\in\VARsome{C'\funarg{\forallvari y{}}}}:
\bigmath{
  \existsvari u{}\nottight{R'}\forallvari y{}
.}
\\\underline{Proof of Claim~4:}
Let 
\math{
  \existsvari u{}
  \in
  \VARall{C'\funarg{\forallvari y{}}}
}.
By the definition of \math{C'} 
there is some \math{\existsvari v{}\in\VARsome{C\funarg{\forallvari y{}}}}
with \bigmath{\existsvari u{}\nottight{E_\sigma}\existsvari v{}.}
Since \math C is a \pair R <-\cc, we have
\bigmath{
  \existsvari v{}\nottight R\forallvari y{}
.}
Thus, by definition of \math{R'} we have
\bigmath{
  \existsvari u{}\nottight{R'}\forallvari y{}
.}\QED{Claim~4}
\end{proofparsepqed}

\vfill\begin{proofqed}{\lemmref{lemma strong reduces to 5a}}\\
Since \math{G} is \stronglyvalidinA{R'}{C'} in \nolinebreak\algebra, there 
is some strong existential \pair\algebra{R'}-valuation \nolinebreak\math{e'}
such that \math{G} is \stronglyvalid{e'}\algebra{C'}.
Let \math e be the strong existential \pair\algebra R-valuation
with \bigmath{\epsilon\funarg e=\epsilon\funarg{e'}} given
by \lemmref{lemma changing R}(1) due to \bigmath{R\tightsubseteq R'.}
Let \math\pi\ be \pair e\algebra-compatible with \nolinebreak\math C.
It suffices to show that \math{G} is \trip\pi e\algebra-valid.
Since the notion of \pair e\algebra-compatibility does not depend
on the precise form of \math e besides \math{\epsilon\funarg e},
we know that \math\pi\ is also \pair{e'}\algebra-compatible with 
\nolinebreak\math C.
Due to \bigmath{C'\tightsubseteq C,} 
\math\pi\ is also \pair{e'}\algebra-compatible with \math{C'}.
Finally, since \math{G} is \stronglyvalid{e'}\algebra{C'},
we conclude that \math{G} is \trip\pi{e'}\algebra-valid,
\ie\ \trip\pi e\algebra-valid.
\end{proofqed}

\vfill\begin{proofparsepqed}{\lemmref{lemma compatibility}}
\underline{(1):}
Since \math C is a \pair R <-\cc,
we know that \math< is a wellfounded ordering on \nolinebreak\Vall\
and
\bigmath{
  R\subseteq\Vsome\tighttimes\Vall
.}
Moreover, we have 
\bigmath{
  S_e\subseteq\Vall\tighttimes\Vsome
}
and
\bigmath{
  \Vsome
  \tightcap
  \Vall
  =
  \emptyset
.}
Thus, if \math\lhd\ is not wellfounded,
then there is an infinitely descending sequence 
of the form
\bigmath{
  y_{2i+2}
  \nottight{S_e}
  y_{2i+1}
  \nottight{\inparenthesesinlinetight{
    R
    \circ
    \tight\leq
  }}
  y_{2i}
}
for all \math{i\in\N}.
Since \math C is a \pair R <-\cc,
we know that 
\bigmath{
  \inparenthesesinlinetight{R\circ\tight<}\subseteq R
.}
Thus, we get 
\bigmath{
  y_{2i+2}
  \nottight{S_e}
  y_{2i+1}
  \nottight{R}
  y_{2i}
}
for all \math{i\in\N}.
This means that 
\transclosureinline{\inparenthesesinlinetight{S_e\tight\circ R}}
is not wellfounded, which contradicts the assumption that
\math e is a strong existential \pair\algebra R-valuation.

\noindent
\underline{(2):}
Let 
\math{
  \pi\in\FUNSET{\inparenthesesinlinetight{\Vall\tightsetminus\DOM C}}\algebra
}.
By noetherian induction on \math\lhd\
and with the help of a choice function we can define some 
\math{\varrho\in\FUNSET\Vfree\algebra} in the following way:
For \math{x\in\Vsome}:
\bigmath{
  \varrho\funarg x
  :=
  e\funarg x\funarg{\domres\varrho{\revrelappsin{S_e}x}}
.}
For \math{x\in\Vall\tightsetminus\DOM C}:
\bigmath{
  \varrho\funarg x
  :=
  \pi\funarg x
.}
For \math{x\in\DOM C}:
\bigmath{
  \varrho\funarg x
  :=
  a
,}
where \math a is an element of the universe of \algebra\ such that,
if possible, 
\math{C\funarg x} is not \trip{\domres\varrho\Vall}e\algebra-valid.
For this definition to be okay, 
we have to show, for each \math{x\in\Vfree}, that
\math{\varrho\funarg x} is defined in terms of 
\bigmath{
  \domres\varrho{\revrelappsin\lhd x}
.}
In case of \bigmath{x\tightin\Vsome,} this is obvious
because \bigmath{S_e\tightsubseteq\tight\lhd.}
In case of \bigmath{x\tightin\Vall\tightsetminus\DOM  C,} this is trivial.
Thus, let \math{x\in\DOM  C}.
Since \math C is a \pair R <-\cc,
we have
  \bigmath{
    \forallvari z{}
     <
    x
  }
  for all 
  \math{
    \forallvari z{}
    \in
    \VARall{C\funarg x}
    \tightsetminus
    \{x\}
  }
and
  \bigmath{
    \existsvari u{}
    \nottight R
    x
  }
  for all 
  \math{
    \existsvari u{}
    \in
    \VARsome{C\funarg x}
  }.
Thus, since \bigmath{R\tightsubseteq\tight\lhd,}
by induction hypothesis,
\trip{\domres\varrho\Vall}e\algebra-validity of \math{C\funarg x}
means validity of \math{C\funarg x} 
in \nolinebreak\math{\algebra\tight\uplus\varrho}.
Moreover, since \bigmath{\tight<\tightsubseteq\tight\lhd,}
we know that 
\math{\varrho\funarg x} is defined in terms of 
\math{
  \domres\varrho{\VARfree{C\funarg x}\setminus\{x\}}
  \subseteq
  \domres\varrho{\revrelappsin\lhd x}
}.
Finally, we define \math{\xi_\pi:=\domres\varrho{\DOM C}}.

For showing that \math{\pi\tight\uplus\xi_\pi} is \pair e\algebra-compatible
with \math C, let \math{\forallvari y{}\in\DOM C} and suppose that
\math{C\funarg{\forallvari y{}}} is 
\trip
  {\pi\tight\uplus\xi_\pi}
  e
  \algebra
-valid,
\ie\
\trip
  {\domres\varrho\Vall}
  e
  \algebra
-valid.
Thus, by definition of \math\varrho,
we know that,
for all \math{\eta\in\FUNSET{\{\forallvari y{}\}}\algebra},
\
\math{C\funarg{\forallvari y{}}} \nolinebreak is 
\trip
  {\domres\varrho{\Vall\setminus\{\forallvari y{}\}}
   \tight\uplus\eta}
  e
  \algebra
-valid,
\ie\
\trip
  {\pi\tight\uplus\domres{\xi_\pi}{\Vall\setminus\{\forallvari y{}\}}
   \tight\uplus\eta}
  e
  \algebra
-valid. The rest is trivial.

\noindent
\underline{(3a):}
Let \math\xi\ be given as in (2).
Define \math{e'} via 
\\
\bigmath{
  e'\funarg x\funarg\tau
  :=
  \xi_\pi\funarg{\reverserelation\varsigma\funarg x}
}
(\math{x\tightin\RAN\varsigma}, 
 \math{
   \tau
   \in
   \FUNSET
     {\inparenthesesinlinetight{
        \inparenthesesinlinetight{\Vall\tightsetminus\DOM C}
        \cap
        \revrelappsin\lhd{\reverserelation\varsigma\funarg x}
     }}
   \algebra
 }, 
 where
 \math{
   \pi
   \in
   \FUNSET{\inparenthesesinlinetight{\Vall\tightsetminus\DOM C}}\algebra
 }  
 an arbitrary extension of \nolinebreak\math\tau
)
and
\\
\bigmath{
  e'\funarg x\funarg\tau
  :=
  e
  \funarg x
  \funarg{
    \domres
      {\inparenthesesinlinetight{\pi\tight\uplus\xi_\pi}}
      {\revrelappsin{S_e}x}
  }
}
(\math{x\tightin\Vsome\tightsetminus\RAN\varsigma},
 \math{
   \tau
   \in
   \FUNSET
     {\inparenthesesinlinetight{
       \inparenthesesinlinetight{\Vall\tightsetminus\DOM C}
       \cap
       \revrelappsin\lhd x}}
     \algebra
 },
 where
 \math{
  \pi
  \in
  \FUNSET
    {\inparenthesesinlinetight{
       \Vall\tightsetminus\DOM C}}
    \algebra
 } 
 an arbitrary extension of \nolinebreak\math\tau
).
\\
Note that this definition is okay because the choice of 
\math\pi\ does not matter:
For the first \math\pi\ this is directly given by (2).
For the second \math\pi\ we have:
\bigmath{
  \domres\pi{\revrelappsin{S_e}x}
  \subseteq
  \domres\pi{\revrelappsin\lhd x}
  \subseteq
  \tau
,}
and, 
for \math{y\in\DOM C\cap\revrelappsin{S_e}x},
by (2),
\math{\xi_\pi\funarg y} is already determined by
\bigmath{
  \domres\pi{\revrelappsin\lhd y}
  \subseteq
  \domres\pi{\revrelappsin\lhd x}
  \subseteq
  \tau
.}
\\
Then
\bigmath{
  S_{e'}
  =
  \domres\id{\Vall\setminus\DOM C}
  \nottight\circ
  \inparentheses{
    \displaystyle\bigcup_{y\in\RAN\varsigma}
      \revrelappsin\lhd{\reverserelation\varsigma\funarg y}
      \times
      \{y\}
    \ \cup\displaystyle\bigcup_{x\in\Vsome\setminus\RAN\varsigma}
       \revrelappsin\lhd x\times\{x\}
  }
.}
\\
Due to 
\bigmath{
 R'
 =
 \domres\id{\Vsome\setminus\RAN\varsigma}\tight\circ R
 \ \cup\displaystyle\bigcup_{y\in\RAN\varsigma}\{y\}\times
   \relappsin\unlhd{\reverserelation\varsigma\funarg y}
 \ \cup\ \Vsome\tighttimes\DOM C
,}
we get
\\
\bigmath{
  S_{e'}\circ R'
  \nottight{\nottight{\nottight\subseteq}}
  \domres\id{\Vall\setminus\DOM C}\nottight\circ
}
\\
\bigmath{
  \inparenthesestight{
    \displaystyle\bigcup_{y\in\RAN\varsigma}
      \revrelappsin\lhd{\reverserelation\varsigma\funarg y}
      \times
      \relappsin\unlhd{\reverserelation\varsigma\funarg y}
    \ \ \cup\displaystyle\bigcup_{x\in\Vsome\setminus\RAN\varsigma}
    \inparenthesestight{\revrelappsin\lhd x\tighttimes\{x\}}
    \circ R
    \ \ \cup\ \ \Vall\tighttimes\DOM C
  }
}
\\
\bigmath{
  \nottight{\nottight{\nottight\subseteq}}
  \domres\id{\Vall\setminus\DOM C}\nottight\circ\inparenthesesinline{
    \tight\lhd\ \cup\ \Vall\tighttimes\DOM C
  }
.}
Thus, 
\transclosureinline{\inparenthesesinlinetight{S_{e'}\tight\circ R'}} 
is a wellfounded ordering because \math\lhd\ is wellfounded by (1). 
This means that
\math{e'} is a strong existential \pair\algebra{R'}-valuation.
It now suffices to show that \math{G\varsigma} is \trip\tau{e'}\algebra-valid
for all \math{\tau\in\FUNSET\Vall\algebra}. 
Set \math{\pi:=\domres\tau{\Vall\setminus\DOM C}}.
We get the following equalities for the below reasons:
\\\LINEmath{\begin{array}{l l}
  \EVAL{\algebra\uplus\epsilon\funarg{e'}\funarg\tau\uplus\tau}
  \funarg{G\varsigma}
  &=\\
  \EVAL{
    \algebra
    \uplus
    \inparenthesesinlinetight{\inparenthesesinlinetight{
        \domres\id\Vsome
        \uplus
        \domres\id{\Vall\setminus\DOM\varsigma}
        \uplus
        \varsigma
      }
    \circ
    \EVAL{\algebra\uplus\epsilon\funarg{e'}\funarg\tau\uplus\tau}}
  }
  \funarg{G}
  &=\\
  \EVAL{
    \algebra
    \uplus
    \epsilon\funarg{e'}\funarg\tau
    \uplus
    \domres\tau{\Vall\setminus\DOM\varsigma}
    \uplus
    \inparenthesesinlinetight{
      \varsigma
      \tight\circ
      \inparenthesesinlinetight{
         \epsilon\funarg{e'}\funarg\tau
      }
    }
  }
  \funarg{G}
  &=\\
  \EVAL{
    \algebra
    \uplus
    \epsilon
    \funarg e
    \funarg{\pi\uplus\xi_\pi}
    \uplus
    \domres\tau{\Vall\setminus\DOM\varsigma}
    \uplus
    \domres{\xi_\pi}{\DOM\varsigma}
  }
  \funarg{G}
  &=\\
  \EVAL{
    \algebra
    \uplus
    \epsilon
    \funarg e
    \funarg{\pi\uplus\xi_\pi}
    \uplus
    \pi\uplus\xi_\pi
  }
  \funarg{G}
  &=\\
  \TRUEpp
\end{array}}\\
First: By the Substitution-Lemma.
Second: By distributing \math\circ\ over \math\cup.
Third: Since, for
\math{x\in\VARsome G}
we have \math{x\in\Vsome\tightsetminus\RAN\varsigma}
and thus
\bigmath{
  \epsilon
  \funarg{e'}
  \funarg\tau
  \funarg x
  =
  \epsilon
  \funarg{e}
  \funarg{\pi\uplus\xi_\pi}
  \funarg x
.}
Moreover, since, for \math{x\in\DOM\varsigma},
\bigmath{
  \epsilon
  \funarg{e'}
  \funarg\tau
  \funarg{\varsigma\funarg x}
  =
  \xi_\pi
  \funarg{\reverserelation\varsigma\funarg{\varsigma\funarg x}}
  =
  \xi_\pi
  \funarg x
,}
we get 
\bigmath{
  \varsigma
  \circ
  \inparenthesesinlinetight{\epsilon\funarg{e'}\funarg\tau}
  =
  \domres{\xi_\pi}{\DOM\varsigma}
.}
Fourth: By noting that
\bigmath{\DOM\varsigma=\VARall G\cap\DOM C.}
Fifth: Because \math{\pi\tight\uplus\xi_\pi} 
is \pair e\algebra-compatible with \math C
(by (2)) and \math G is \stronglyvalid e\algebra C.

\noindent
\underline{(3b):}
When \math G is \stronglyvalidinA R C in \nolinebreak\algebra,
then there is some strong existential \pair R\algebra-valuation 
\nolinebreak\math e such that \math G is \stronglyvalid e\algebra C.
By (3a), \math{G\varsigma} is 
\stronglyvalidinA{R'}\emptyset\ in \nolinebreak\algebra.
Since 
\bigmath{
  \domres R{\Vsome\setminus\RAN\varsigma}
  \tightsubseteq
  R''
  \tightsubseteq
  R'
,}
by \lemmref{lemma strong reduces to 5a},
\math{G\varsigma} is 
\stronglyvalidinA{\domres R{\Vsome\setminus\RAN\varsigma}}\emptyset\
and \stronglyvalidinA{R''}\emptyset\ in \nolinebreak\algebra.
\end{proofparsepqed}

\vfill\begin{proofparsepqed}{\lemmref{lemma strong reduces to}}
(1), (2), (3), and (4) are trivial.

\noindent
\underline{(5):}
Let \math{e'} be a strong existential \pair\algebra{R'}-valuation
and \math{\pi} be \pair{e'}\algebra-compatible with \math{C'}
such that \math{G_1} is \trip\pi{e'}\algebra-valid.
Let \math e be the strong existential \pair\algebra R-valuation
with \bigmath{\epsilon\funarg e=\epsilon\funarg{e'}} given
by \lemmref{lemma changing R}(1) due to \bigmath{R\tightsubseteq R'.}
Then \math\pi\ is \pair e\algebra-compatible with \math{C'},
and \math{G_1} \nolinebreak is \trip\pi e\algebra-valid.
Moreover, due to \bigmath{C\tightsubseteq C',}
\math\pi\ is \pair e\algebra-compatible with \math{C}.
Thus, since \math{G_0} strongly \pair R C-reduces to \math{G_1},
also \math{G_0} is \trip\pi e\algebra-valid. This also means that 
\math{G_0} \nolinebreak is \trip\pi{e'}\algebra-valid as was to be shown.

\noindent
\underline{(6a):}
Suppose that \math{G_0\sigma} is 
\stronglyvalidinA{R'}{C'} in \nolinebreak\algebra.
Then there is some strong existential \pair\algebra{R'}-valuation \math{e'}
such that \math{G_0\sigma} is \stronglyvalid{e'}\algebra{C'}.
Then, by \lemmref{lemma changing R}(2), 
there is some strong existential \pair\algebra{R}-valuation \math{e}
such that for all \math{\pi\in\FUNSET\Vall\algebra}:
\bigmath{
  \epsilon\funarg e\funarg\pi
  =
  \sigma\circ\EVAL{\algebra\uplus\epsilon\funarg{e'}\funarg\pi\uplus\pi}
.}
Moreover, for \math{y\in\Vall} we have:
\bigmath{
  \pi\funarg y
  =
  \EVAL{\algebra\uplus\epsilon\funarg{e'}\funarg\pi\uplus\pi}\funarg{y}
,}
\\  
\ie
\LINEmath{
      \epsilon
      \funarg e
      \funarg\pi
      \uplus
      \pi
  =
        \inparenthesesinlinetight{
          \sigma
          \uplus
          \domres\id\Vall
        }
        \circ
        \EVAL{\algebra\uplus\epsilon\funarg{e'}\funarg\pi\uplus\pi}
.}
\\
Thus, for any formula \math B, we have
\\\LINEmath{
  \begin{array}{l l}
    \EVAL{
      \algebra
      \uplus
      \epsilon
      \funarg e
      \funarg\pi
      \uplus
      \pi
    }
    \funarg B
  &=\\
    \EVAL{
      \algebra
      \uplus
      \inparenthesesinlinetight{\inparenthesesinlinetight{
          \sigma
          \uplus
          \domres\id\Vall
        }
        \circ
        \EVAL{\algebra\uplus\epsilon\funarg{e'}\funarg\pi\uplus\pi}
      }
    }
    \funarg B
  &=\\
    \EVAL{\algebra\uplus\epsilon\funarg{e'}\funarg\pi\uplus\pi}
    \funarg{B\sigma},
  \end{array}
}
\\
the latter step being due to the Substitution-Lemma.

\noindent
Thus, for any set of sequents \math{G'} 
and any \math{\pi\in\FUNSET\Vall\algebra}: 

\noindent
\LINEnomath{
  \trip\pi e\algebra-validity of
  \math{G'}
  is logically equivalent to 
  \trip\pi{e'}\algebra-validity of
  \math{G'\sigma}.
}(:\math{\S_1})

\noindent
Especially, for any \math{\pi\in\FUNSET\Vall\algebra}: 
\\\LINEnomath{
\math\pi\ is \pair e\algebra-compatible with \math C
\uiff\ \math\pi\ is \pair{e'}\algebra-compatible with \math{C'}.
}(:\math{\S_2})

\noindent
Thus,
for any set of sequents \math{G'}: 

\noindent
\LINEnomath{
  \math{G'} is \stronglyvalid e\algebra C
  \uiff\
  \math{G'\sigma} is \stronglyvalid{e'}\algebra{C'}.
}

\noindent
Especially, \math{G_0} is \stronglyvalid e\algebra C.
Thus, \math{G_0} is \stronglyvalidinA R C in \nolinebreak\algebra.

\noindent
\underline{(6b):}
Let \math{e'} be some strong existential \pair\algebra{R'}-valuation,
\math\pi\ be \pair{e'}\algebra-compatible with \nolinebreak\math{C'},
and suppose that \math{G_1\sigma} is \trip\pi{e'}\algebra-valid.
Let \math{e} be the existential \pair\algebra{R}-valuation 
given by \lemmref{lemma changing R}(2).
Then, by (\math{\S_2}) in the proof of (6a),
\math\pi\ is \pair{e}\algebra-compatible with \nolinebreak\math{C},
and, by (\math{\S_1}) in the proof of (6a),
\math{G_1} is \trip\pi e\algebra-valid.
By assumption, 
\math{G_0} \nolinebreak strongly \pair R C-reduces to \nolinebreak\math{G_1}.
Thus, \math{G_0} is \trip\pi e\algebra-valid, too.
By (\math{\S_1}) in the proof of (6a), this means that
\math{G_0\sigma} is \trip\pi{e'}\algebra-valid as was to be shown.
\end{proofparsepqed}\vfill\pagebreak

\begin{proofqed}{\theoref{theorem strong closed means valid}}\\
Since \Ax\ is \stronglyvalidinA{\Vsome\tighttimes\Vall}\emptyset, 
\math t is closed,
\bigmath{R\subseteq\Vsome\tighttimes\Vall,}
and 
\bigmath{\emptyset\subseteq C,}
by \lemmref{lemma strong reduces to 5a}, 
\Open t is \stronglyvalidinA R C.
Since 
\bigmath{(\Feins,t)\tightin\foresteins}
and 
\quar\foresteins C R < satisfies the invariant condition,
\math{\{\Feins\}} strongly \pair R C-reduces to \Open t.
Then, by \lemmref{lemma strong reduces to}(1),
\math{\Feins} is \stronglyvalidinA R C.
Finally, by \lemmref{lemma compatibility}(3b),
\math{\Feins\varsigma} is 
\stronglyvalidinA{R'}\emptyset\ and
\stronglyvalidinA{\domres R{\Vsome\setminus\RAN\varsigma}}\emptyset.
\end{proofqed}

\vfill
\begin{proofparsepqed}
{\theoref{theorem strong abstract deductive calculi}}
\quar\emptyset\emptyset\emptyset\emptyset\ trivially satisfies
the strong invariant condition. For the iteration
steps, let \math{\pair\Feinsprimeprime{t''}\in\forestzwei}.
Assuming the strong invariant condition for 
\quar\foresteins C R <, we have to show that
\math{C'} is a \pair{R'}{<'}-\cc\ and that
\math{\{\Feinsprimeprime\}} strongly \pair{R'}{C'}-reduces to \Open{t''}.

\noindent
\underline{Hypothesizing:}
Due to the assumed \bigmath{R\tight\circ\tight<\subseteq R}
and the required \bigmath{R''\tight\circ\tight<\subseteq R'=R\tightcup R'',}
we have 
\bigmath{R'\circ\tight<\nottight{\nottight=}\inparenthesesinlinetight
  {R\tightcup R''}\circ\tight<\nottight{\nottight\subseteq}R\tightcup R''
  \nottight{\nottight=}R'
.}
Thus, \math{C} is a \pair{R'}{<}-\cc.
Moreover, due to \bigmath{C'\tightequal C} and 
\bigmath{\tight<'\tightequal\tight<,}
\trip{C'}{R'}{\tight<'} is an extension of \trip C R <.
In case of \bigmath{\pair\Feinsprimeprime{t''}\tightin\foresteins,}
\math{\{\Feinsprimeprime\}} \pair R C-reduces to \Open{t''} by
assumption, and then, due to \lemmref{lemma strong reduces to}(5),
\math{\{\Feinsprimeprime\}} strongly \pair{R'}{C'}-reduces to \Open{t''}.
Otherwise we have 
\bigmath{\pair\Feinsprimeprime{t''}\tightequal\pair\Feins t
.}
Then
\bigmath{\{\Feinsprimeprime\}=\{\Feins\}=\Open t=\Open{t''}
.}
Thus, by \lemmref{lemma strong reduces to}(2),
\math{\{\Feinsprimeprime\}} strongly \pair{R'}{C'}-reduces to \Open{t''}.

\noindent
\underline{Expansion:}
In case of \bigmath{\pair\Feinsprimeprime{t''}\tightin\foresteins,}
\math{\{\Feinsprimeprime\}} \pair R C-reduces to \Open{t''} by
assumption, and then, due to \trip{C'}{R'}{\tight<'} being an extension of
\trip C R < and \lemmref{lemma strong reduces to}(5),
\math{\{\Feinsprimeprime\}} 
\nolinebreak strongly \pair{R'}{C'}-reduces to \Open{t''}.
Otherwise we have 
\bigmath{
  \pair\Feinsprimeprime{t''}
  \tightequal
  \pair\Feins{t'}
.}
Since \bigmath{\Open{t}\tightsetminus\{\Fzwei\}\subseteq\Open{t'},}
by \lemmref{lemma strong reduces to}(2),
\math{\Open{t}\tightsetminus\{\Fzwei\}} strongly 
\pair{R'}{C'}-reduces to \Open{t'}.
Thus, since by assumption
\math{\{\Fzwei\}} strongly \pair{R'}{C'}-reduces to a subset of \Open{t'},
by \lemmref{lemma strong reduces to}(4)
\Open{t} strongly \pair{R'}{C'}-reduces to \Open{t'}.
Moreover, due to 
\bigmath{
  \pair\Feins t\tightin\foresteins
,}
by assumption
\math{\{\Feins\}} \nolinebreak strongly \pair R C-reduces to \Open t.
Thus, by \lemmref{lemma strong reduces to}(5),
\math{\{\Feins\}} strongly \pair{R'}{C'}-reduces to \Open t.
Thus, since 
\Open{t} strongly \pair{R'}{C'}-reduces to \Open{t'},
by \lemmref{lemma strong reduces to}(3),
\math{\{\Feins\}} strongly \pair{R'}{C'}-reduces to \Open{t'},
\ie\
\math{\{\Feinsprimeprime\}} strongly \pair{R'}{C'}-reduces to \Open{t''}.

\noindent
\underline{Instantiation:}
By \lemmref{lemm ex str s up},
\math{C'} is a \pair{R'}{<'}-\cc.
There is some \math{\pair\Feins t\in\foresteins} such that
\bigmath{
  \pair\Feins t\sigma
  =
  \pair\Feinsprimeprime{t''}
.}
By assumption, 
\math{\{\Feins\}} strongly \pair R C-reduces to \Open t.
By \lemmref{lemma strong reduces to}(6b),
\math{\{\Feins\sigma\}} strongly \pair{R'}{C'}-reduces to 
\math{\Open t\sigma},
\ \ie\
\math{\{\Feinsprimeprime\}} strongly \pair{R'}{C'}-reduces to \Open{t''}.
\end{proofparsepqed}

\vfill
\begin{proofparsepqed}{\theoref{theorem strong sub-rules}}
Let \algebra\ be an arbitrary \math\Sigmaoffont-structure
(\math\Sigmaoffont-algebra).
We only prove the first example of each kind of rule to be a sub-rule
of the Expansion rule and leave the rest as an exercise.

\noindent
\underline{\math\alpha-rule:}
We have to show that 
\math{
  \{\Gamma~\inparenthesesinlinetight{A\tightoder B}~\Pi\}  
}
strongly \pair R C-reduces to 
\math{
  \{A~B~\Gamma~\Pi\}
}
in \nolinebreak\algebra.
This is trivial, however, because \trip\pi e\algebra-validity 
of the two sets is logically equivalent for each strong existential
\pair\algebra R-valuation \math e
and \math{\pi\in\FUNSET\Vall\algebra}.

\noindent
\underline{\math\beta-rule:}
We have to show that 
\math{
  \{\Gamma~\inparenthesesinlinetight{A\tightund B}~\Pi\}
}
strongly \pair R C-reduces to 
\math{
  \{A~\Gamma~\Pi,~~~~~~B~\Gamma~\Pi\}
}
in \nolinebreak\algebra.
This is trivial, however, because \trip\pi e\algebra-validity 
of the two sets is logically equivalent for each strong existential
\pair\algebra R-valuation \math e
and \math{\pi\in\FUNSET\Vall\algebra}.

\pagebreak

\yestop\noindent
\underline{\math\gamma-rule:}
We have to show that \math{\{\Gamma~\exists x\tight: A~\Pi\}}
strongly \pair R C-reduces to 
\math{
  \{A\{x\tight\mapsto\existsvari x{}\}~\Gamma~\exists x\tight: A~\Pi\}
}
\linebreak
in \nolinebreak\algebra.
This is the case, however, because \trip\pi e\algebra-validity 
of the two sets is logically equivalent for each strong existential
\pair\algebra R-valuation \math e
and \math{\pi\in\FUNSET\Vall\algebra}.
The direction from left to right is given because the former
sequent is a sub-sequent of the latter. The other direction,
which is the only one we actually have to show here, is also
clear because \trip\pi e\algebra-validity of 
\bigmath{
  A\{x\tight\mapsto\existsvari x{}\}
}
implies
\trip\pi e\algebra-validity of 
\bigmath{
  \exists x\tight:\, A
.}
Although this is clear, we should be a little more explicit here
because the standard semantic definition of \math\exists\
(\cf\ \eg\ \cite{wirthdiss}, \p~188) 
does not use \fev s 
and is somewhat more complicated than it could be
in terms of \fev s. Moreover, in the note above the theorem 
we remarked that 
the restriction on \existsvari x{} not occurring in the former
sequent is not really necessary. Thus, in order to be more explicit
here, assume that
the latter sequent is \trip\pi e\algebra-valid for some
strong existential \pair\algebra R-valuation \math e
and some \math\pi\ that is \pair e\algebra-compatible with \math C.
We have to show that the former sequent is \trip\pi e\algebra-valid.
If this is not the case,
\bigmath{
  A\{x\tight\mapsto\existsvari x{}\}
}
must be \trip\pi e\algebra-valid.
Let \math{\forallvari y{}\in\Vall\tightsetminus\VARall A}.
Then,
since
\bigmath{
  A
  \{x\tight\mapsto\forallvari y{}\}
  \{\forallvari y{}\tight\mapsto\existsvari x{}\}
}
is equal to 
\bigmath{
  A\{x\tight\mapsto\existsvari x{}\}
,}
we know that
\bigmath{
  A
  \{x\tight\mapsto\forallvari y{}\}
  \{\forallvari y{}\tight\mapsto\existsvari x{}\}
}
is valid in
\bigmath{\algebra\uplus\epsilon\funarg e\funarg\pi\uplus\pi.}
Then, by the Substitution-Lemma,
\bigmath{
  A
  \{x\tight\mapsto\forallvari y{}\}
}
is valid in
\bigmath{\algebra\uplus\epsilon\funarg e\funarg\pi\uplus\pi'}
for \math{\pi'\in\FUNSET\Vall\algebra} given by
\bigmath{
  \domres{\pi'}{\Vall\setminus\{\forallvari y{}\}}
  :=
  \domres\pi{\Vall\setminus\{\forallvari y{}\}}
}
and
\bigmath{
  \pi'\funarg{\forallvari y{}}
  :=
  \epsilon\funarg e\funarg\pi\funarg{\existsvari x{}}
.}
By the standard semantic definition of \math\exists\
and since quantification on \math x cannot occur
in \math A (as \math{\exists x\tight:\,A} is a formula
in our restricted sense, \cf\ \sectref{section quantification restriction}),
this means that 
\math{
  \exists x\tight:
  \inparenthesesinlinetight{ 
  A
  \{x\tight\mapsto\forallvari y{}\}
  \{\forallvari y{}\tight\mapsto x\}
  }
}
is valid in \math{\algebra\uplus\epsilon\funarg e\funarg\pi\uplus\pi}.
Since \forallvari y{} does not occur in \math A,
this formula is equal to \math{\exists x\tight: A},
which means that the former sequent is \trip\pi e\algebra-valid
as was to be shown.

\yestop\noindent
\underline{\math\delta-rule:}
Firstly, we have to show that 
\math{C'} is a \pair{R'}{<'}-\cc.
Since 
\bigmath{
  \forallvari x{}
  \notin
  \VARall A
  \cup
  \DOM<
} 
and \math< is a wellfounded
ordering,
\bigmath{
 \tight{<'}
 \nottight{\nottight{:=}}
 \tight<
 \nottight{\nottight{\cup}}
 \revrelapp{\tight\leq}{\VARall A}
 \times
 \{\forallvari x{}\}
}
is a wellfounded ordering with 
\bigmath{
  \forallvari x{}
  \notin
  \DOM{<'}
,} 
too.
Therefore,
\bigmath{
  R''\circ\tight{<'}
  =
  \emptyset
,}
and then 
\bigmath{
  R'\circ\tight{<'}
  =
  \inparenthesesinlinetight{R\cup R''}\circ\tight{<'}
  =
  R\circ\tight{<'}
  =
  R\circ\inparenthesesinlinetight{\tight<\cup\tight{<''}}
  =
  \inparenthesesinlinetight{R\circ\tight<}
  \cup
  \inparenthesesinlinetight{R\circ\tight{<''}}
  \subseteq
  R\cup R'' 
  =
  R'  
;}
where the inclusion is due to the following:
first, we have \bigmath{R\circ\tight<\subseteq R} because
\math C is a \pair R <-\cc;
second, in case of \bigmath{z_0\nottight R z_1<''z_2}
we have \bigmath{z_2=\forallvari x{}} and there is some
\math{z'\in\VARall A} with
\bigmath{z_1\leq z';} then, again by \bigmath{R\circ\tight<\subseteq R,} 
we get \bigmath{z_0\nottight R z',} \ie\ 
\bigmath{z_0\nottight{R''}\forallvari x{}=z_2.}
Since \bigmath{\tight<\subseteq\tight{<'},}
\bigmath{R\subseteq R',} 
\bigmath{
  C'=C\cup\{\pair{\forallvari x{}}{A\{x\tight\mapsto\forallvari x{}\}}\}
,}
\bigmath{
  \VARall{C'\funarg{\forallvari x{}}}\tightsetminus\{\forallvari x{}\}
  =
  \VARall{A\{x\tight\mapsto\forallvari x{}\}}\tightsetminus\{\forallvari x{}\}
  =
  \VARall A\tightsetminus\{\forallvari x{}\}
  =
  \VARall A
  \subseteq
  \revrelappsin{\tight{<'}}{\forallvari x{}}
,}
and 
\bigmath{
  \VARsome{C'\funarg{\forallvari x{}}}
  =
  \VARsome{A\{x\tight\mapsto\forallvari x{}\}}
  =
  \VARsome A
  \subseteq
  \revrelappsin{R'}{\forallvari x{}}
,}
the remaining requirements for \math{C'} to be a \pair{R'}{<'}-\cc\
are easily checked.

Secondly, we have to show that 
\math{
  \{\Gamma~\forall x\tight: A~\Pi\}
}
strongly \pair{R'}{C'}-reduces to 
\math{
  \{A\{x\tight\mapsto\forallvari x{}\}~\Gamma~\Pi\}
}
in \nolinebreak\algebra.
Assume that the latter sequent is \trip\pi e\algebra-valid for some
strong existential \math{R'}-valuation \math e
and some \math\pi\ that is \pair e\algebra-compatible with \math{C'}.
We have to show that the former sequent is \trip\pi e\algebra-valid.
If some formula in \math{\Gamma\Pi} is 
\trip\pi e\algebra-valid, 
then the former sequent is 
\trip\pi e\algebra-valid, too.
Otherwise, 
this means that \math{A\{x\tight\mapsto\forallvari x{}\}}
is \trip\pi e\algebra-valid.
Since \math\pi\ is \pair e\algebra-compatible with \math{C'},
\math{A\{x\tight\mapsto\forallvari x{}\}}
is \trip{\pi'}e\algebra-valid
for all \math{\pi'\in\FUNSET\Vall\algebra} with 
\bigmath{
  \domres{\pi'}{\Vall\setminus\{\forallvari x{}\}}
  =
  \domres\pi{\Vall\setminus\{\forallvari x{}\}}
.}
Since 
\bigmath{
  {\VARsome{A\{x\tight\mapsto\forallvari x{}\}}\times\{\forallvari x{}\}}
  =
  {\VARsome A\times\{\forallvari x{}\}}
  \subseteq
  R'
,}
we know that 
\math{A\{x\tight\mapsto\forallvari x{}\}}
is even valid in
\bigmath{
  \algebra\uplus\epsilon\funarg e\funarg\pi\uplus\pi'
}
for all \math{\pi'\in\FUNSET\Vall\algebra} with 
\bigmath{
  \domres{\pi'}{\Vall\setminus\{\forallvari x{}\}}
  =
  \domres\pi{\Vall\setminus\{\forallvari x{}\}}
.}
By the standard semantic definition of \math\forall\
(\cf\ \eg\ \cite{wirthdiss}, \p~188) 
and since quantification on \math x cannot occur
in \math A (as \math{\forall x\tight:A} is a formula
in our restricted sense, \cf\ \sectref{section quantification restriction}),
this means that 
\math{
  \forall x\tight:
  \inparenthesesinlinetight{ 
  A
  \{x\tight\mapsto\forallvari x{}\}
  \{\forallvari x{}\tight\mapsto x\}
  }
}
is valid in \math{\algebra\uplus\epsilon\funarg e\funarg\pi\uplus\pi}.
Since \forallvari x{} does not occur in \math A,
this formula is equal to \math{\forall x\tight: A},
which means that the former sequent is \trip\pi e\algebra-valid
as was to be shown.
\end{proofparsepqed}
\vfill\pagebreak
\def\refname{References and Notes}%

\newcommand\textendfootnote[1]{\footnotemark#1\par}
\setcounter{footnote}{0}

\yestop
\noindent
\textendfootnote
{For Skolemization in constrained logics \cf\ \cite{skolemconstraint},
 where, however, only the existence of solutions of constraints
 and not the form of the solutions itself is preserved.}
\textendfootnote
{While this paradigm of inductive theorem proving was already
 used by the Greeks, \fermatname\ (1601-1665) rediscovered it
 under the name ``descente infinie'', and in our time it is
 sometimes called ``implicit induction'', \cf\ \cite{wirthbecker}, 
 \cite{wirthdiss}.}
\textendfootnote
{The notation \relapp R A is in the tradition of 
 \cite{bourbaki}, Chapitre~II, \S~3, D\'efinition~3,
 where \math{R\langle A\rangle} is written 
 in order to
 clearly distinguish relation application \relapp R A from function
 application \math{R\funarg A}. In \cite{wirthdiss} we still used to 
 write \math{R[A]} instead of \relapp R A. In this \paper, however,
 this notation would lead to confusion with our use of optional
 brackets.}
\textendfootnote
{We do not need the more complicated definitions of 
 a sequent as a pair of lists of formulas or 
 as a T/F-tagged list of formulas
 because we do not consider calculi 
 where the separation of a sequent into
 antecedent and succedent is important,
 like LJ in \cite{gentzen} or 
 the ``symmetric \gentzen\ systems'' in \cite{smullyan}.}
\textendfootnote
{Note that \reverserelation R is an\emph{inverse}
 (in the sense that
 \bigmath{
   R\tight\circ\reverserelation R
   \tightequal
   \domres\id{\DOM R}
 }
 and 
 \bigmath{
   \reverserelation R\tight\circ R
   \tightequal
   \domres\id{\RAN R}
 }
 holds)
 \uiff\ \math R is an injective function.}
\textendfootnote
{For the notion of a\emph{tree} \cf\ \cite{knutheins}.
 As a special feature we would like an explicit representation
 of leaves, such that, when we add the elements of a set \math G
 as children to a leaf node \math l, this \math l is not a leaf anymore,
 even if \math G is empty.}
\textendfootnote
{Note that this restriction on \existsvari x{} is not required for
 soundness (\cf\ \theoref{theorem sub-rules}) but for efficiency only.}
\textendfootnote
{Note that this restriction on \existsvari x{} is not required for
 soundness (\cf\ \theoref{theorem strong sub-rules}) but for efficiency only.}
\textendfootnote
{Actually, when \math{E_\sigma} is efficiently added to the graph
 representing \math R and \math{U_\sigma} in order to represent 
 \math{R'} \bigmath{:=}
 \math{E_\sigma\circ R\circ\refltransclosureinline{\inparenthesesinlinetight{
 U_\sigma\tight\circ R}}}, \ 
 an element \math{\pair{\existsvari u{}}{\existsvari x{}}\in E_\sigma}
 is simply implemented by drawing a new edge from the 
 (possibly new) node for 
 \existsvari u{} to the old node for \existsvari x{}.
 (\existsvari u{} gets a new node \uiff\ 
  \ \math{
   \pair{\existsvari u{}}{\existsvari u{}}\notin E_\sigma
  }.)
 Although this graph is not really bipartite in \Vsome- and \Vall-nodes, 
 when checking for acyclicity of 
 \bigmath{
   U_{\sigma'}\tight\circ R'
 ,}
 when finding a new \Vsome-node to be already on the active path,
 we can detect a cycle of \Vsome-nodes simply by asking whether
 we are coming from a \Vsome-node, in which case we skip the new
 \Vsome-node and do not signal a cycle of 
 \bigmath{
   U_{\sigma'}\tight\circ R'
 .}}
\textendfootnote
{We have very recently presented these calculi at the 
 \nth 2 \Int\ Workshop on First-Order Theorem Proving (FTP) in 
 \Nov\ 1998 in Vienna (\cf\ \cite{wirthftp}), where nobody in the
 audience was able to point out other work in this direction.}

\yestop
\yestop
\noindent
{{\bf Acknowledgements:}
I would like to thank \furbachname\ and his whole group
for all they taught me on tableau calculi.
Furthermore, 
I would like to thank \kohlhasename\ for his encouragement 
to drop Skolemization
and \padawitzname\ for the possibility to finish this work.
Finally, I am indebted to an anonymous referee for his careful reading
of a short version of this \paper\
and his most useful remarks.

}
\end{document}